\documentclass[11pt,draftcls,onecolumn]{IEEEtran}

\usepackage{cite}

\usepackage{colortbl}
\usepackage{color}
\usepackage{bm}
\usepackage{bbm}
\usepackage{amsmath}

\usepackage{cases}

\usepackage{amssymb}
\usepackage{amsthm}
\usepackage{graphicx}
\usepackage{mathrsfs}
\usepackage{empheq}	
\usepackage[mathcal]{euscript}
\usepackage[margin = 2cm]{geometry}
\usepackage{framed}
\usepackage{comment}

\usepackage{dsfont}

\DeclareFontFamily{U}{mathx}{\hyphenchar\font45}
\DeclareFontShape{U}{mathx}{m}{n}{
      <5> <6> <7> <8> <9> <10>
      <10.95> <12> <14.4> <17.28> <20.74> <24.88>
      mathx10
      }{}
\DeclareSymbolFont{mathx}{U}{mathx}{m}{n}
\DeclareFontSubstitution{U}{mathx}{m}{n}
\DeclareMathAccent{\widecheck}{0}{mathx}{"71}

\newtheorem{theorem}{Theorem}
\newtheorem{lemma}{Lemma}

\newtheorem{assumption}{Assumption}

\newtheorem{property}{Property}
\newtheorem{definition}{Definition}

\newcommand{\E}{\mathbb{E}}

\newcommand{\beq}{\begin{equation}}
\newcommand{\eeq}{\end{equation}}
\newcommand{\beqa}{\begin{IEEEeqnarray}{rCl}}
\newcommand{\eeqa}{\end{IEEEeqnarray}}

\newcommand{\dfz}{\triangleq}

\newcommand{\matpow}{\bar{{\sf P}}}

\title{Distributed Adaptive Learning Under Communication Constraints}
\author{Marco Carpentiero, Vincenzo Matta, and Ali H. Sayed
\thanks{
Marco Carpentiero and Vincenzo Matta are with the Department of Information and Electrical Engineering and Applied Mathematics (DIEM), University of Salerno, via Giovanni Paolo II, I-84084, Fisciano (SA), Italy, and also with the National Inter-University Consortium for Telecommunications (CNIT), Italy (e-mails: \{mcarpentiero, vmatta\}@unisa.it).

Ali H. Sayed is with the \'Ecole Polytechnique F\'ed\'erale de Lausanne EPFL, School of Engineering, CH-1015 Lausanne, Switzerland (e-mail: ali.sayed@epfl.ch).
}
}

\begin{document}
	\maketitle

\begin{abstract}
This work examines adaptive distributed learning strategies designed to operate under communication constraints. 
We consider a network of agents that must solve an online optimization problem from continual observation of {\em streaming} data. To this end, the agents implement a distributed {\em cooperative} strategy where each agent is allowed to perform {\em local} exchange of information with its neighbors. In order to cope with communication constraints, the exchanged information must be unavoidably compressed to some extent. 
We propose a distributed diffusion strategy nicknamed as ACTC (Adapt-Compress-Then-Combine), which relies on the following main steps: $i)$ an adaptation step where each agent performs an individual stochastic-gradient update with {\em constant step-size}; $ii)$ a compression step that leverages a recently introduced class of {\em stochastic compression operators}; and $iii)$ a combination step where each agent combines the {\em compressed} updates received from its neighbors.
The distinguishing elements of novelty of this work are as follows. 
First, we focus on {\em adaptive} strategies, where constant (as opposed to diminishing) step-sizes are critical to infuse the agents with the ability of responding in real time to nonstationary variations in the observed model.
Second, our study considers the general class of {\em directed} (i.e., non-symmetric) and {\em left-stochastic} combination policies, which allow us to enhance the role played by the network topology in the learning performance.
Third, in contrast with related works that assume strong convexity for all individual agents' cost functions, we require only a {\em global strong convexity at a network level}. For global strong convexity, it suffices that a single agent has a strongly-convex cost, while the remaining agents might feature even a non-convex cost. 
Fourth, we focus on a {\em diffusion} (as opposed to consensus) strategy, which will be shown to entail some gains in terms of learning performance.
Under the demanding setting of {\em global} strong convexity and {\em compressed information}, we are able to prove that the iterates of the ACTC strategy fluctuate around the right global optimizer (with a mean-square-deviation in the order of the step-size), achieving remarkable savings in terms of bits exchanged between neighboring agents. 
Illustrative examples are provided to validate the theoretical results, and to compare the ACTC strategy against up-to-date stochastic-gradient solutions with compressed data, highlighting the benefits of the proposed solution.

\end{abstract}


\section{Introduction}
\IEEEPARstart{I}{n} the last decades, the steady progress of statistical learning and network science led to great interest in {\em distributed optimization} strategies implemented by multi-agent networks~\cite{ProcIEEEspecialissue2020,TsitsiklisBertsekasAthansTAC1986,NedicBertsekasSIAM2001,NedicOzdaglar2010,BoydFoundTrends,NedicJSTSP2013,KhanTAC2016,KhanTAC2018,RabbatRibeiroCoopGraphSP2018,BajwaTSIPN,SayedTuChenZhaoTowficSPmag2013,SayedProcIEEE2014,ChenSayedTIT2015part1,ChenSayedTIT2015part2,  DeGroot,XiaoBoydSCL2004,BoydGhoshPrabhakarShahTIT2006,DimakisKarMouraRabbatScaglioneProcIEEE2010}.
Primary advantages of distributed strategies include scalability, possibility of working with reduced-size and spatially dispersed datasets, and robustness against failures. 
Even more remarkably, distributed optimization allows each network agent to overcome its individual limitations through cooperation with its neighbors. 
For example, agents can have limited view on their own and their {\em individual} cost functions may lead to poor and/or multiple minimizers, i.e., poor solutions to a certain learning task ({\em local unidentifiability}). 
This issue can be remediated through a distributed strategy, which might enable minimization of a global cost function that attains a more satisfying and unique solution ({\em global identifiability}).
Moreover, in many contexts the distributed cooperative strategies can deliver superior performance w.r.t. single-agent strategies, attaining the same performance as a centralized solution~\cite{Sayed,MattaSayedCoopGraphSP2018}.

Several useful distributed implementations have been proposed and examined in great detail in previous works. 
These implementations can be categorized in terms of different attributes. For example, we have distributed implementations for gradient-descent and stochastic-gradient descent algorithms; constant as opposed to diminishing step-size implementations; consensus and diffusion strategies~\cite{Sayed}.
In all these cases, distributed optimization and learning algorithms require exchange of information among spatially dispersed agents. Accordingly, the exchanged data must be unavoidably compressed to some extent. 
For this reason, {\em data compression} lies at the core of any distributed implementation. 
Unfortunately, a thorough characterization of distributed strategies with compressed data is challenging, especially due to the technical difficulties arising from the nonlinear character of the compression operator (e.g., a quantizer). 
The main purpose of this work is to carry out a detailed analysis of the learning behavior of distributed strategies operating under communication constraints.

\subsection{Distributed Quantization}
\label{sec:DistQua}
The earlier works on quantization for distributed inference considered: $i)$ network architectures with a single fusion center; $ii)$ a static setting where inference is performed in a single step (e.g., no gradient descent or other iterative algorithms are used); and $iii)$ known parametric models for the underlying distributions of the data. Under this setting, there are several quantization strategies available in the literature, with reference to different inferential and learning tasks, such as decentalized estimation~\cite{LamReibmanTC1993,GubnerTIT1993,AsymptoticDesignQuantizers2007,TongARE}, decentralized detection and classification~\cite{LongoLookabaughGrayTIT1990,Tsitsiklis1993,ViswanathanVarshneyProcIEEE1997,BlumKassamPoorProcIEEE1997,ChamberlandVeeravalliTSP2003, SaligramaTSP2006,HanAmariTIT1998,CEOproblem}, also with many useful variations including cross-layer~\cite{MergenNawareTongTSP2007,LBMA} or censoring approaches~\cite{RagoWillettBarShalomTAES1996}.

One critical difficulty in data compression for inferential problems is the lack of knowledge about the underlying data distributions, which depend on the same {\em unknown} parameters that the agents are trying to learn.
This lack of knowledge complicates severely the tuning of the quantizers' thresholds, since these thresholds would depend on the unknown parameters to be estimated.
A breakthrough in the solution of this problem was given by the introduction of {\em stochastic quantizers} in the distributed learning strategies. In the context of data compression (with single-agent and not for inference applications) the usage of stochastic quantizers  can be dated back to the seminal works on probabilistic analysis of quantizers~\cite{Widrow1956} and dithering~\cite{GrayStockhamTIT1993}. 
In comparison, one of the earliest appearances of stochastic quantizers in distributed estimation was provided in~\cite{LuoTIT2005}, which inspired subsequent applications to distributed learning~\cite{PreddKulkarniPoorSPM2006,MaranoMattaWillettTSP2013}. 
These works rely on a universal, fully data-driven approach, where the thresholds of the stochastic quantizers do not require knowledge of the underlying distribution. In fact, these thresholds are randomly adapted to the measurements observed locally by the agents, in such a way that the resulting quantizer is {\em unbiased on the average}, despite the lack of knowledge about the underlying distribution. This property is critical to make the inferential strategies capable of learning well.

As said, the aforementioned works refer to static strategies where inference is performed in a single step. 
In optimization and learning problems, there is often the necessity of employing iterative strategies to converge to the desired solution. In these settings, the data compression issue becomes even more critical, since the quantization error can accumulate over successive iterations, destroying the convergence of the pertinent optimization algorithms. 

This issue motivated the recent introduction of randomized quantizers for distributed optimization algorithms such as gradient descent or stochastic gradient descent algorithms. 
One recent solution that attracted attention in this context is the general class of randomized compression strategies introduced in~\cite{AlistarhNIPS2017}. This class relies on two properties that are critical to guarantee proper learning, namely, {\em unbiasedness} and {\em non blow-up}. We will examine in detail these properties in Sec.~\ref{sec:Alistarh}. 

More specifically, in~\cite{AlistarhNIPS2017} it was assumed that each agent broadcasts compressed versions of its locally observed gradients to all other agents in the network.
However, calling upon the theory of predictive quantization (e.g., the sigma-delta modulation adopted in PCM~\cite{GershoGrayBook}), we see that the impact of quantization errors on convergence can be reduced by properly leveraging the inherent memory arising in recursive implementations such as gradient descent implementations. Two canonical paradigms to achieve this goal are {\em error-feedback management} and {\em differential quantization}, which, perhaps surprisingly, have been applied to distributed optimization quite recently. 
 
Under {\em error-feedback management}, at each time step: first, the current gradient plus an error stored from the previous iteration is compressed; and then, the compression error is compensated in the subsequent iteration. 
One of the earliest works showing the benefits of error compensation applied to stochastic gradient algorithms is~\cite{SeideFuDroppoLiYuINTERSPEECH2014}, where the effectiveness of extreme $1$-bit quantization with error feedback is shown in the context of distributed training of deep neural networks.
Another example of error compensation is provided in~\cite{WuHuangHuangZhangICMLR2018}, where the agents' gradient updates are compressed using the random quantizers proposed in~\cite{AlistarhNIPS2017}. 
Before feeding the quantizer, the local gradients are corrected with a compensation term that accounts for the quantization error from previous iterations. 
Convergence guarantees of stochastic gradient algorithms with compressed gradient updates and error compensation were provided in~\cite{StichCordonnierJaggiNIPS2018} and~\cite{KarimireddyRebjockStichJaggiICML2019}.

Let us switch to the {\em differential quantization} approach. This is a more direct way to leverage the memory present in the iterative algorithm, which aims at reducing the error variance by compressing only the difference between subsequent iterates.
For a fixed budget of quantization bits, it is indeed more convenient to compress the difference (i.e., the innovation) between consecutive samples, rather than the samples themselves. This is because: $i)$ the innovation typically exhibits a reduced range as compared to the entire sample; and $ii)$ owing to the correlation between consecutive samples, quantizing the entire sample will waste resources by transmitting redundant information. 
The information-theoretic fundamental limits of (non-stochastic) gradient descent under differential quantization have been recently established in~\cite{LinKostinaHassibiISIT2021}.

However, the aforementioned works on error compensation and differential quantization referred either to {\em fully-connected} networks (i.e., there exists a direct communication link between any two agents) or to agents communicating with a {\em single} fusion center tasked to perform a centralized gradient update.
In this work, we focus instead on the more challenging setting where optimization must be {\em fully decentralized}. 
Under this scenario, each agent is responsible for its own inference, which is obtained by successive steps of local interaction with its neighbors. 
When moving to the fully distributed setting, new challenges related to the topology and the interplay between agents need to be considered, adding significant complexity to the analysis of compression for distributed optimization. 
The first important difference is that, since the agents are generally not fully connected, even without communication constraints they cannot compute a common gradient function, implying that exchanging only gradient updates would impair convergence~\cite{NedicOlshevskyRabbatIEEEPROC2018}. 
Therefore, the fully distributed setting requires agents to exchange their iterates rather than plain gradient updates.

Typical strategies for fully-distributed optimization {\em without} compression constraints are consensus or diffusion strategies~\cite{Sayed}. However, applying these strategies with compressed data is nontrivial, since without proper design of the quantizers, significant bias is introduced in the learning algorithm, which prevents plain consensus or diffusion implementations from converging to the right minimizer. 
One early characterization of adaptive diffusion with compressed data was provided in~\cite{ZhaoTuSayedSP2012}, where the compression errors were modeled as noise over the communication channel.
In comparison, there are other works that address the quantization issue in fully-distributed strategies by focusing on the explicit encoder structure.
Some useful results are available for the case of {\em exact, i.e., non-stochastic gradient-type algorithms}. 
In~\cite{NedicOlshevskyOzdaglarTsitsiklisCDC2008}, uniform quantization of the iterates is considered for a distributed implementation of the subgradient descent algorithm.
In this scheme, the agents update locally their state variables by averaging the state variables received from their neighbors, and then follow the subgradient descent direction. More recently, additional convergence results were presented in~\cite{DoanMaguluriRombergIEEEAUTCONT12020}, where random (dithered) quantization is applied, along with a weighting scheme to give more or less importance to the analog local state and the quantized averaged state of the neighbors. 
In~\cite{ReisidazehMokhtariHassaniPedarsaniIEEESP2019}, the randomized quantizers proposed in~\cite{AlistarhNIPS2017} are considered for a distributed gradient descent implementation using consensus with compressed iterates and an update rule similar to the one adopted in~\cite{DoanMaguluriRombergIEEEAUTCONT12020}. 

All the aforementioned works on fully-distributed schemes under communication constraints differ significantly from our proposal, as they rely on availability of the {\em exact} gradient. In the present work, we focus instead on the {\em adaptive} setting where the agents collect {\em noisy} streaming data to evaluate a {\em stochastic instantaneous approximation} of the actual gradient, and must be endowed with online algorithms capable to respond in real time to drifts in the underlying conditions. 
Useful communication-constrained and fully-decentralized implementations that can be applied to this setting were recently proposed in~\cite{KoloskovaStichJaggiICML2019,KoloskovaLinStichJaggiICLR2020,BitsForFree}.  

We are now ready to summarize the main novel and distinguishing contributions offered in this article, in comparison to the pertinent previous works.

\subsection{Main Contributions}

\vspace*{5pt}
\noindent
{\em --- Diffusive Adaptation.} 
The Adapt-Compress-Then-Combine (ACTC) strategy proposed in this work belongs to the family of {\em diffusion} strategies~\cite{Sayed}, while the available works on distributed optimization under communication constraints focus on consensus strategies~\cite{KoloskovaStichJaggiICML2019,KoloskovaLinStichJaggiICLR2020,BitsForFree}.  
Along with many commonalities, one fundamental difference between diffusion and consensus resides in the asynchrony of the latter strategy in the combination step (where the updated state of an agent is combined with the previous states of its neighbors)~\cite{Sayed}. 
This asynchrony has an effect both in terms of stability and learning performance. In fact, it has been shown that consensus algorithms can feature smaller range of stability as compared to diffusion strategies and a slightly worse learning performance~\cite{Sayed}. 
For these reasons, in this work we opt for a diffusion scheme. Starting from the traditional (uncompressed) Adapt-Then-Combine (ATC) diffusion strategy detailed in~\cite{Sayed}, we allow for local exchange of {\em compressed} variables by means of {\em stochastic quantizers}. 
We will see that, thanks to the diffusion mechanism, the ACTC scheme will be able to adapt and learn well, and in particular it will outperform previous distributed quantized strategies based on consensus.

We focus on a dynamic setting where the agents are called to learn by continually collecting {\em streaming data} from the environment. Under an adaptive setting, once the distributed learning algorithm starts, we want the agents to learn virtually forever, by automatically adapting their behavior in face of nonstationary drifts in the streaming data. 
To this end, {\em stochastic} gradient algorithms with {\em constant} step-size are necessary. 
These algorithms have been shown to tradeoff well learning and adaptation.
On the learning side, each agent resorts to some instantaneous approximation of the cost function (which is not perfectly known in practice) and tries to learn with increasing precision by leveraging the increasing information coming from the streaming data.
On the adaptation side, the constant step-size leaves a persistent amount of ``noise'' in the algorithm (the ``gradient noise'')  which automatically infuses the algorithm with the ability of promptly reacting to drifts. 
In contrast, over diminishing step-size implementations, the gradient noise is progressively annihilated over time. 
As a consequence, diminishing step-size algorithms learn infinitely better as time progresses under stationary conditions. 
At the same time, if the minimizer changes (e.g., because of drifts in the underlying distribution) diminishing step-size algorithms get stuck on the previously computed minimizer, exhibiting a sort of ``elephant's memory'', i.e., requiring a time to get out from a local minimizer that is at least proportional to the time the algorithm needed to approach that minimizer.

The existing results on distributed stochastic gradient descent with compressed data focus mainly on non-adaptive implementations with diminishing step-size. Some results for constant step-sizes are available in~\cite{BitsForFree}, under a setting that differs from our setting in terms of the critical features described in the next two items, namely, type of combination policy and assumptions on the local cost functions. 

\vspace*{5pt}
\noindent
{\em --- Left-Stochastic combination policies.} 
The existing works on distributed optimization under communication constraints focus on {\em symmetric and doubly-stochastic} combination policies. This is not necessarily the case in distributed optimization algorithms. In particular, communication between pair of nodes can be asymmetric, meaning that node $k$ can scale the data received from a neighbor $\ell$ with a weight $a_{\ell k}$ that differs from the weight $a_{k\ell}$ used by $k$ to scale the data received from $\ell$. In particular, we can have {\em directed} graphs where, e.g., $\ell$ and $k$ are communicating only in one direction (e.g., $a_{\ell k}>0$ while $a_{k\ell}=0$). 
For this reason, in this work we consider the more general setting of {\em left-stochastic} combination policies. 
Left-stochastic matrices allow us to represent a significantly richer variety of distributed interactions, where the network topology plays a fundamental role. For example, the limiting Perron eigenvector of left-stochastic matrices is not uniform, a property that can be exploited to compensate for non-uniform agents' behavior~\cite{Sayed}.
Moreover, by acting on the topology and/or on the left-stochastic combination matrix, one can tune the Perron eigenvector so as to  explore different Pareto-optimal solutions~\cite{Sayed}. 
Last but not least, differently from doubly-stochastic matrices, left-stochastic matrices can be constructed in practice without requiring any coordination across the agents.
From a technical viewpoint, the fact that our combination matrices are not required to be neither symmetric nor doubly stochastic, introduces significant additional complexity in the technical analysis. 

\vspace*{5pt}
\noindent
{\em --- Global strong convexity.} 
Convergence of the stochastic gradient {\em iterates} is typically examined under the assumption that the gradients are Lipschitz and the cost functions are strongly convex. 
In the distributed setting, the latter property is usually translated into assuming that all the local cost functions pertaining to the individual agents are strongly convex~\cite{KoloskovaStichJaggiICML2019,BitsForFree}.
Sometimes the additional assumption of uniform gradient boundedness is adopted (e.g., in~\cite{KoloskovaStichJaggiICML2019}), which can however hold only approximately in the Lipschitz and strongly-convex setting. 
For Lipschitz gradient and strongly-convex local function, without the uniform boundedness approximation, convergence results were recently obtained for distributed primal-dual algorithms with compressed communication~\cite{BitsForFree}. 

Notably, in the present work we relax the aforementioned assumptions, since we do not rely on any uniform boundedness approximation, and require only a {\em global} cost function (i.e., a linear combination of the local cost functions) to be strongly convex.
This extension allows us to cover relevant cases where, e.g., a {\em single} agent has a strongly-convex cost, with the remaining agents having possibly non-convex costs. 
Moreover, global, as opposed to local strong convexity, can be exploited to implement distributed regularization procedures where the limitations of the individual agents can be overcome by cooperation. For example, $N-1$ agents might be unable to learn the true minimizer (local unidentifiability), but they can nevertheless compensate their limited view by cooperating with a single farsighted agent. 
One example of this type will be considered in Sec.~\ref{sec:experiments}. 

It is worth mentioning that there are recent works for distributed optimization under communication constraints for non-convex problems~\cite{KoloskovaLinStichJaggiICLR2020}. 
However, owing to the lack of convexity, these studies do not focus on the convergence of the iterates, which is instead our focus in this work. 
Convergence of the iterates is especially relevant in the inferential setting where knowing the minimizer is critical to provide interpretation of the model, performing influence factor analysis, variable selection, and so on.
Moreover, convergence of the iterates is also relevant for prediction purposes, since the minimizers (i.e., the estimated model parameters) are the input of the prediction stage and, hence, faithful estimation becomes critical to faithful prediction.

{\em -- Summary of main results}. 
The main achievement of this work is the characterization of the learning behavior of the ACTC strategy, namely, of an adaptive diffusion strategy for learning under communication constraints. 
The pertinent technical proof, articulated in several intermediate results reported in the appendices, is demanding, especially due to the relaxation of the traditional assumption of strongly-convex local functions, and, even more remarkably, due to the further nonlinear behavior induced by the compression operator. 
An essential building block to prove the result is given by the unifying description and mathematical tools developed in~\cite{ChenSayedTIT2015part1,Sayed}, which allow us to decouple the learning dynamics over two main components by means of a suitable network coordinate transformation. These two components are: $i)$ a {\em centralized} stochastic-gradient component that converges to a unique solution common to all agents; $ii)$ a deviation component that dies out as time elapses, and which takes into account the initial discrepancy between the trajectories of the individual agents. 

Finally, we compare our algorithm against two up-to-date solutions, namely, the algorithms CHOCO-SGD~\cite{KoloskovaStichJaggiICML2019} and DUAL-SGD~\cite{BitsForFree}, showing that both algorithms are outperformed by our ACTC strategy. As we will carefully explain in Sec.~\ref{sec:experiments}: $i)$ the improvement on DUAL-SGD arise mainly from the advantage of primal-domain strategies (like the proposed ACTC) over primal-dual distributed strategies (like DUAL-SGD); $ii)$ the improvement on CHOCO-SGD arise primarily from using a diffusion-type strategy as opposed to the consensus strategy employed by CHOCO-SGD. 

\vspace*{10pt}
{\bf Notation}. We use boldface letters to denote random variables, and normal font letters for their realizations. 
Capital letters refer to matrices, small letters to both vectors and scalars. 
Sometimes we violate the latter convention, for instance, we denote the total number of network agents by $N$. 
All vectors are column vectors. In particular, the symbol $\mathds{1}_L$ denotes an $L\times 1$ vector whose entries are identically equal to $1$. Likewise, the identity matrix of size $L$ is denoted by $I_L$.  
For two square matrices $X$ and $Y$, the notation $X\geq Y$ signifies that $X-Y$ is positive semi-definite.
In comparison, for two rectangular matrices $X$ and $Y$, the notation $X\succeq Y$ signifies that the individual entries of $X-Y$ are nonnegative.
For a vector $x$, the symbol $\|x\|$ denotes the $\ell_2$ norm of $x$. 
For a matrix $X$, the  $\ell_2$ induced matrix norm is accordingly $\|X\|$. 
Other norms will be characterized by adding the pertinent subscript. 
For example $\|x\|_1$ will denote the $\ell_1$ norm of $x$, and $\|X\|_1$  the $\ell_1$ induced matrix norm (maximum absolute column sum of $X$).
The symbol $\otimes$ denotes the Kronecker product.
The symbol $*$ denotes complex conjugation. 
$X^{\top}$ is the transpose of matrix $X$, whereas $X^{\mathsf{H}}$ is the Hermitian (i.e., conjugate) transpose of a complex matrix $X$.
The symbol $\E$ denotes the expectation operator.
For a nonnegative function $f(\mu)$, the notation $f(\mu)=O(\mu)$ signifies that there exists a constant $C>0$ and a value $\mu_0$ such that $f(\mu)\leq C \mu$ for all $\mu\leq\mu_0$. 

\section{Background}
We consider a network of $N$ agents solving a distributed optimization problem. 
Each individual agent $k=1,2,\ldots,N$ is assigned a {\em local} cost or risk function:
\beq
J_k(w): \mathbb{R}^M\rightarrow \mathbb{R}.
\eeq
The local cost functions are assumed to satisfy the following regularity condition.
\begin{assumption}[Individual cost function smoothness]
\label{Individual cost function smoothness}
For all $w\in\mathbb{R}^M$, each cost function $J_k(w)$ is twice-differentiable and its Hessian matrix satisfies the following Lipschitz condition, for some positive constants $\{\eta_k\}$:
\beq
\nabla^2 J_k(w) \leq \eta_k \, I_M.
\label{nablaLip}
\eeq
~\hfill$\square$    
\end{assumption}

In practice, it is seldom the case that the cost functions are perfectly known to the agents. 
In contrast, each agent usually has access to a {\em stochastic} approximation of the true cost function. 
For example, in the adaptation and learning theory the cost functions are often modeled as the expected value of a loss function $L_k(w;\bm{x}_k)$, namely,
\beq
J_k(w)=\E[L_k(w;\bm{x}_k)],
\label{eq:costasexpec}
\eeq
where the expectation is taken w.r.t. a random variable $\bm{x}_k$ that can represent, e.g., some training data observed by agent $k$. 
In many scenarios of interest, the statistical characterization of $\bm{x}_k$ is not available to the agents and, hence, $J_k(w)$ is not known and is rather approximated by the stochastic quantity $L_k(w;\bm{x}_k)$. 
Moreover, if different data $\bm{x}_{k,i}$ are collected over time by agent $k$, the stochastic approximation takes the form of an {\em instantaneous} approximation $L_k(w;\bm{x}_{k,i})$ depending on time index $i$. 

More generally, whether or not the cost function is defined through \eqref{eq:costasexpec}, in the following treatment we assume that agent $k$ at time $i$ is able to approximate the true gradient $\nabla J_k(w)$ through a {\em stochastic instantaneous approximation} $\bm{g}_{k,i}(w)$ which, without loss of generality, can be written as the true gradient plus a {\em gradient noise} term $\bm{n}_{k,i}(w)$, namely,
\beq
\bm{g}_{k,i}(w)=\nabla J_k(w) + \bm{n}_{k,i}(w).
\label{gradNoiseDef}
\eeq

\subsection{Classical ATC Diffusion Strategy}
The Adapt-Then-Combine (ATC) diffusion strategy is a popular distributed mechanism that consists of iterated application of the following two steps, for $i=1,2,\ldots$
\begin{equation}\label{ATC}
    \begin{cases}
      \bm{\psi}_{k,i} = \bm{w}_{k,i-1} - \mu_k \bm{g}_{k,i}(\bm{w}_{k,i-1})
~~&\textnormal{[Adapt]}\\
\\
      \bm{w}_{k,i} = \displaystyle{\sum_{\ell=1}^N} a_{\ell k}\bm{\psi}_{\ell,i} &\textnormal{[Combine]}
    \end{cases}
\end{equation} 
In \eqref{ATC}, agents $k=1,2,\ldots, N$ evolve over time $i$ by producing a sequence of iterates $\bm{w}_{k,i}\in\mathbb{R}^M$.
The adaptation step is a {\em self-learning} step, where each agent $k$ at time $i$ computes its own instantaneous stochastic approximation $\bm{g}_{k,i}(\cdot)$ of the local cost function $J_k(\cdot)$, evaluated at the previous iterate $\bm{w}_{k,i-1}$. Such an approximation is scaled by a small step-size $\mu_k>0$ and used to update the previous iterate $\bm{w}_{k,i-1}$ following the (stochastic) gradient descent. 
The maximum step-size across the agents will be denoted by:
\beq
\mu\triangleq \max_{k=1,2,\ldots,N} \mu_k,
\eeq
giving rise to the {\em scaled} step-sizes:
\beq
\alpha_k\dfz\frac{\mu_k}{\mu}.
\label{eq:alphascalestepdef0}
\eeq
The combination step is a {\em social learning} step, where agent $k$ aims at realigning its descent direction with the rest of the network by combining its local update $\bm{\psi}_{k,i}$ with the other agents' updates scaled by some nonnegative scalars $\{a_{\ell k}\}$, which are referred to as {\em combination weights}. 
The support graph of the combination matrix $A=[a_{\ell k}]$ describes the connections between agents, i.e., the topology of a network whose vertices correspond to the agents, and whose edges represent directional links between agents. 
According to this model, when no communication link exists between agents $\ell$ and $k$, the combination weights $a_{\ell k}$ and $a_{k \ell}$ must be equal to zero. Likewise, when information can flow only from $\ell$ to $k$, we will have $a_{\ell k}>0$ and $a_{k\ell}=0$. In summary, the combination process is a local process where only neighboring agents interact. 

It is useful to introduce the neighborhood of agent $k$:
\beq
\mathcal{N}_k\triangleq\{\ell=1,2,\ldots,N: a_{\ell k}>0\},
\eeq
which is a {\em directed} neighborhood that accounts for the incoming flow of information from $\ell$ to $k$ (possibly including the self-loop $\ell=k$).

We will work under the following standard regularity conditions on the network. 

\begin{assumption}[Strongly-Connected Network]
\label{Strong Connectivity}
The network is strongly-connected, which means that, given any pair of nodes $(\ell,k)$, a path with nonzero weights exists in both directions (i.e., from $\ell$ to $k$ and vice versa), and that at least one agent $k$ in the entire network has a self-loop ($a_{kk}>0$).~\hfill$\square$ 
\end{assumption}

\begin{assumption}[Stochastic combination matrix]
\label{Stochastic combination matrix}
For each agent $k=1,...,N$ the following conditions hold:
\begin{equation}\label{combWeights}
a_{\ell k} \geq 0, \quad \sum_{\ell\in\mathcal{N}_k}^N a_{\ell k} = 1,\quad a_{\ell k}=0 \textnormal{ for }\ell\notin\mathcal{N}_k,
\end{equation}
which imply that the combination matrix $A=[a_{\ell k}]$ is a left-stochastic matrix.~\hfill$\square$
\end{assumption}

Under Assumptions~\ref{Strong Connectivity} and~\ref{Stochastic combination matrix}, the combination matrix $A$ is a primitive matrix, and thus satisfies the Perron-Frobenius theorem, which in particular implies the existence of the {Perron vector} $\pi=[\pi_1,\pi_2,\ldots,\pi_N]^{\top}$, a vector with all strictly positive entries satisfying the following relationship:
\beq
A \pi =\pi,\qquad \mathds{1}_N^{\top} \, \pi =1.
\label{eq:Perronvecfirstdef}
\eeq
For later use, it is convenient to introduce the vector $p$ that mixes the topological information encoded in the Perron eigenvector with the scaled step-sizes $\{\alpha_k\}$, namely,
\beq
p=[\alpha_1 \pi_1 ,\alpha_2 \pi_2,\ldots,\alpha_N \pi_N]^{\top}.
\label{eq:scalePerron}
\eeq

We are now ready to introduce the global strong convexity assumption that will be required for our results to hold.
\begin{assumption}[Global Strong Convexity]
\label{Global Strong}
Let $p_k=\alpha_k \pi_k$ be the $k$-th entry of the scaled Perron eigenvector in \eqref{eq:scalePerron}. 
The (twice differentiable) aggregate cost function:
\beq
J(w)=\sum_{k=1}^N p_k J_k(w)
\label{eq:globalCost}
\eeq
is $\nu$-strongly convex, namely, a positive constant $\nu$ exists such that, for all $w\in\mathbb{R}^M$ we have:
\beq
\sum_{k=1}^N p_k \nabla^2 J_k(w) \geq \nu \,I_M.
\label{sqNablaStrConvex}
\eeq
\hfill$\square$    
\end{assumption}
We remark that, for Assumption~\ref{Global Strong} to hold, it is necessary that strong convexity holds for {\em only one} local cost function, with the other cost functions being allowed to be non-convex, provided that \eqref{sqNablaStrConvex} is satisfied!\footnote{In particular, strong convexity of a {\em single} cost function is sufficient when the remaining cost functions are convex.} 
In contrast, most works consider the more restrictive setting where {\em all individual cost functions are strongly convex}. In this work, we depart from this restrictive assumption, and adhere instead to the more general setting that was considered in~\cite{ChenSayedTIT2015part1,ChenSayedTIT2015part2,Sayed} for the uncompressed-data setting. 
Therefore, Assumption~\ref{Global Strong} turns out to be a useful generalization that will allow us to cover important practical cases, such as the case where all but a single agent work with {\em locally not identifiable} models. Under these scenarios, the agents with unidentifiable models would not be in the position of converging to a meaningful minimizer, whereas the presence of only a single good agent will enable successful collective learning for the entire network. This particular setting will be illustrated in more detail in Sec.~\ref{sec:experiments}. 
On the other hand, as we will see, working under the assumption that the local functions are not all strongly convex introduces significant complexity in the analysis and makes the derivations more demanding.

For later use, it is useful to notice that Assumptions~\ref{Individual cost function smoothness} and~\ref{Global Strong} entail a relationship between the Lipschitz constants $\{\eta_k\}$ and the strong convexity constant $\nu$. 
Specifically, using \eqref{nablaLip} and \eqref{sqNablaStrConvex} and applying the triangle inequality we can write:
\beq
\nu \leq 
\sum_{k=1}^N p_k \|\nabla^2 J_k(w)\|
\leq 
\sum_{k=1}^N p_k \eta_k\dfz \eta.
\label{eq:LipCvxConst}
\eeq
The adaptation and learning performance of the ATC strategy has been examined in great detail in previous works~\cite{ChenSayedTIT2015part1,ChenSayedTIT2015part2, Sayed}. 
The major conclusion stemming from these works is that, under Assumptions~\ref{Individual cost function smoothness}--~\ref{Stochastic combination matrix} (plus some classical assumptions on the gradient noise --- see Assumption~\ref{Gradient noise process} further ahead), the ATC strategy is able to drive each agent toward a close neighborhood of the minimizer $w^{\star}$ of the {\em global} cost function \eqref{eq:globalCost}.
We remark that in the considered framework each cost function can be different, having a specific minimizer $w_k^0$ (or even multiple minimizers), which may not coincide with the {\em unique} global network minimizer $w^{\star}$ of the aggregate cost function $J(w)$. 

We notice also that the structure of \eqref{eq:globalCost} allows us to solve different optimization problems where the objective function can be expressed as the linear combination of local cost functions, including the special case where the weights $p_k$ are all uniform, which can be obtained when the step-sizes are all equal (i.e., $\alpha_k=1$ for all $k$) and the combination matrix is doubly-stochastic (since the Perron eigenvector of a doubly-stochastic matrix has entries $p_{\ell}=1/N$ for all $\ell=1,2,\ldots,N$). In addition, and remarkably, it was shown in~\cite{Sayed,ChenSayedPareto} that the minimizer of \eqref{eq:globalCost} corresponds to a Pareto solution of the multi-objective optimization problem:
\beq
\min_{w}\{J_1(w),J_2(w),\ldots,J_N(w)\},
\eeq
where the choice of the weights $\{p_k\}$ drives convergence to a particular Pareto solution.

\section{ATC with compressed communication}
The ATC diffusion scheme \eqref{ATC} is designed under the assumption that agents exchange over the communication channel their intermediate updates $\bm{\psi}_{k,i}$. However, in a realistic environment the information shared by the agents must be necessarily  {\em compressed}. This necessity gives rise to at least two fundamental questions.
First, is it possible to design diffusion strategies that preserve the adaptation and learning capabilities of the ATC strategy despite the presence of data compression? 
Assume the answer to the first question is in the affirmative. Then it is natural to ask whether there is a limit on the amount of compression, since it is obviously desirable for the agents to save as much bandwidth and energy as possible.
Our analysis will give precise elements to address these important questions. To start with, we introduce a compressed version of the ATC strategy.

There exist obviously several possibilities to perform data compression. In order to select one particular strategy, we need to consider the fundamental limitations of our setting, in particular: $i)$ lack of knowledge of the underlying statistical model; $ii$) correlation across subsequent iterates. 

Limitation $i)$ is classically encountered in quantization for inference~\cite{LuoTIT2005}, where the usage of standard techniques for quantizer design is not viable. This is because, when the true statistical models are unavailable, such techniques lead to severe estimation bias that eventually impairs the algorithm's convergence. 
An excellent tool to overcome this issue is provided by {\em stochastic} quantization, where suitable introduction of randomness allows to compensate for the bias (on average, over time). 

Limitation $ii)$ is classically encountered in the quantization of random processes. An excellent tool to solve this problem is provided by {\em differential quantization}, which leverages similarity (i.e., correlation) between subsequent samples by quantizing only their difference (i.e., innovation). 

Motivated by the above two observations, in the following we will rely on {\em stochastic differential quantization}, which will be plugged in the ATC recursion giving rise to the Adapt-Compress-Then-Combine (ACTC) diffusion strategy, which can be described as follows. 

The time-varying variables characterizing the ACTC recursion are: an intermediate update $\bm{\psi}_{k,i}$, a differentially-quantized update $\bm{q}_{k,i}$, and the current minimizer $\bm{w}_{k,i}$.
At time $i=0$ each agent $k$ is initialized with an arbitrary state value $\bm{q}_{k,0}$ (with finite second moment). 
Then, agent $k$ receives the initial states $\bm{q}_{\ell,0}$ from its neighbors $\ell\in\mathcal{N}_k$ (such initial sharing is performed with infinite precision, which is immaterial to our analysis since it happens only once) and computes an initial minimizer $\bm{w}_{k,0} = \sum_{\ell \in \mathcal{N}_k} a_{\ell k}\bm{q}_{\ell,0}$.

Then, for every $i>0$, the agents perform the following four operations.
First, each agent $k$ performs {\em locally} the same adaptation step as in the ATC strategy:
\beq
\bm{\psi}_{k,i} = \bm{w}_{k,i-1} - \mu_k\bm{g}_{k,i}(\bm{w}_{k,i-1})~~~~\textnormal{[Adapt]}
\eeq
Second, each agent $k$ compresses the {\em difference} between the update $\bm{\psi}_{k,i}$ and the previous quantized update $\bm{q}_{k,i-1}$, through a compression function $\bm{Q}_k: \mathbb{R}^M\rightarrow \mathbb{R}^M$:
\beq
\label{eq:compressACTC}
\bm{Q}_k(\bm{\psi}_{k,i}-\bm{q}_{k,i-1})~~~~\textnormal{[Compress]}
\eeq
The bold notation for the compression function highlights that {\em randomized} functions are permitted, as we will explain more carefully in Sec.~\ref{sec:Alistarh}. 
Then, agent $k$ receives from its neighbors $\ell\in\mathcal{N}_k$ the {\em compressed} values $\bm{Q}_{\ell}(\bm{\psi}_{\ell,i}-\bm{q}_{\ell,i-1})$. 
Since the quantization operation acts on differences, the quantized states $\bm{q}_{\ell,i}$ must be updated by adding the quantized difference to the previous value $\bm{q}_{\ell,i-1}$. 
Specifically, agent $k$ updates {\em the quantized values corresponding to all} $\ell\in\mathcal{N}_k$:
\begin{equation}
    \bm{q}_{\ell,i} = \bm{q}_{\ell,i-1} + \zeta\,\bm{Q}_{\ell}(\bm{\psi}_{\ell,i} - \bm{q}_{\ell,i-1}),
\label{eq:qupdate}
\end{equation}
where $\zeta\in (0,1)$ is a design parameter that will be useful to tune the stability of the algorithm, as we will carefully explain in due time.
Finally, agent $k$ combines the updated states corresponding to its neighbors as usual:
\beq
\bm{w}_{k,i} = \displaystyle{\sum_{\ell \in \mathcal{N}_k}} a_{\ell k}\bm{q}_{\ell,i}~~~~\textnormal{[Combine]}
\label{eq:combineACTC}
\eeq
It is important to remark that, in order to perform the update step in \eqref{eq:qupdate}, agent $k$ must possess the variables $\bm{q}_{\ell,i-1}$ from its neighbors $\ell\in\mathcal{N}_k$. This might appear problematic at first glance, since we have just seen that only the differences $\bm{Q}_\ell(\bm{\psi}_{\ell,i}-\bm{q}_{\ell,i-1})$ are actually received by $k$ from a neighboring agent $\ell$.
On the other hand, since at $i=0$ agent $k$ knows the initial quantized states $\{\bm{q}_{\ell,0}\}_{\ell\in\mathcal{N}_k}$, and since the quantized update in \eqref{eq:qupdate} depends on $\{\bm{q}_{\ell,i-1}\}_{\ell\in\mathcal{N}_k}$ and the quantized innovation $\{\bm{Q}_\ell(\bm{\psi}_{\ell,i}-\bm{q}_{\ell,i-1})\}_{\ell\in\mathcal{N}_k}$, we conclude that sharing of the quantized differences along with the initial states $\{\bm{q}_{\ell,0}\}_{\ell\in\mathcal{N}_k}$ is enough for agent $k$ to implement \eqref{eq:qupdate} at every instant $i$, by keeping memory only of the last neighboring variables $\{\bm{q}_{\ell,i}\}_{\ell\in\mathcal{N}_k}$.
In summary, the ACTC scheme can be compactly described as follows:

\begin{figure*}[t]
    \centering
    \includegraphics[width=\linewidth]{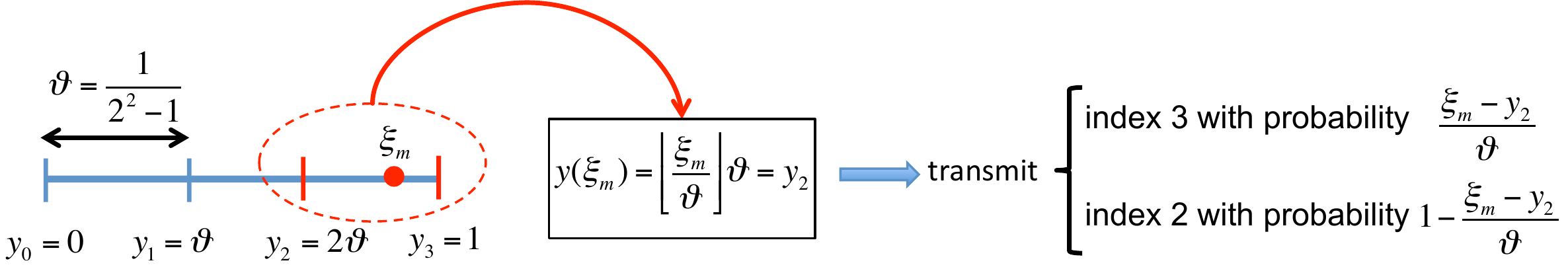}
    \caption{Sketch of the randomized quantizer described in Sec.~\ref{sec:AlistarhQuant}, for the case where the bit-rate is $r=2$.}
    \label{fig:randq}
\end{figure*}

\begin{equation}\label{ACTC}
    \begin{cases}
      \bm{\psi}_{k,i} = \bm{w}_{k,i-1} - \mu_k\bm{g}_{k,i}(\bm{w}_{k,i-1})\\
      \\
      \bm{q}_{k,i} = \bm{q}_{k,i-1} + \zeta\,\bm{Q}_k(\bm{\psi}_{k,i} - \bm{q}_{k,i-1}) \\
      \\
      \bm{w}_{k,i} = \displaystyle{\sum_{\ell \in \mathcal{N}_k}} a_{\ell k}\bm{q}_{\ell,i}
    \end{cases}
\end{equation} 
Before concluding this section, we remark that the combination step of the ACTC strategy considers only quantized variables, even if, in principle, agent $k$ might also combine its {\em analog} state $\bm{\psi}_{k,i}$ in place of the quantized counterpart $\bm{q}_{k,i}$. 
As a general principle, the more we compress, the less we spend (even in terms of local processing). In this respect, showing that the ACTC strategy learns properly by combining only quantized variables has its own interest. 
Moreover, considering only quantized variables gives to the algorithm a symmetric structure that avoids adding further complexity to the mathematical formalization that is necessary to support the analysis.

\subsection{Compression operators}
\label{sec:Alistarh}
In order to implement the ACTC recursion, it is necessary to specify the compression operators $\bm{Q}_k(\cdot)$. 
We will consider the class of compression operators that fulfill the following regularity assumptions.
\begin{assumption}[Compression operators]
\label{Compression operator}
For a positive parameter $\omega$ and an input value $x \in \mathbb{R}^M$, the compression operator $\bm{Q}:\mathbb{R}^M\rightarrow\mathbb{R}^M$ is a randomized operator satisfying the following two properties:
\begin{align}
&\E\,\big[\bm{Q}(x)\big] = x~~~~~~~~~~~~~~~~~ \textnormal{[unbiasedness]}
\label{unbiasedComp}\\
&\E\,\|\bm{Q}(x)-x\|^2 \leq \omega\,\|x\|^2\,~~ \textnormal{[non blow-up property]}
\label{boundVarianceComp}
\end{align}
where expectations are evaluated w.r.t. the randomness of the operator only. When $\bm{x}$ is random, the compression operator is statistically independent of $\bm{x}$. 
\hfill$\square$
\end{assumption}

By ``randomized operator'' we mean that, given a {\em deterministic} input $x$, the output $\bm{Q}(x)$ is {\em randomly} chosen (one meaningful way to perform such random choice will be illustrated in Sec.~\ref{sec:AlistarhQuant}). 
Accordingly, the expectations appearing in \eqref{unbiasedComp} and \eqref{boundVarianceComp} are computed w.r.t. the randomness inherent to operator $\bm{Q}(\cdot)$. 
As stated, whenever we apply the operator to a {\em random} input $\bm{x}$, we assume that the random mechanisms governing $\bm{Q}(\cdot)$ and $\bm{x}$ are statistically independent.

Two main observations are important to capture the meaning of Assumption~\ref{Compression operator}. 
The first one concerns the role of parameter $\omega$, which quantifies the amount of compression. Small values of $\omega$ correspond to small amount of compression, i.e., finely quantized data. Large values of $\omega$ are instead representative of severe compression.
The second observation concerns properties \eqref{unbiasedComp} and \eqref{boundVarianceComp}. 
It will emerge from the technical analysis that property \eqref{unbiasedComp} enables the possibility that the quantization errors arising during the ACTC evolution average to zero as time elapses. 
Property \eqref{boundVarianceComp} will be critical to guarantee that the variance of the quantization errors does not blow-up over time. 

Remarkably, the class of randomized quantizers introduced in Assumption~\ref{Compression operator} is fairly general and flexible, including a broad variety of compression paradigms. 
Some of these paradigms are particularly tailored to our setting, where we have to compress vectors with possibly large dimensionality $M$. 
For example, one useful paradigm is the {\em sparse compression paradigm}, where a small subset of the $M$ components of the input vector $x$ is sent with arbitrarily large precision, while the remaining entries are set to zero~\cite{BitsForFree}.

Another compression scheme that meets Assumption~\ref{Compression operator} was recently popularized in~\cite{AlistarhNIPS2017}, and will be illustrated in the next section. 

\subsection{Randomized Quantizers in~\cite{AlistarhNIPS2017}}
\label{sec:AlistarhQuant}
\begin{itemize}
\item 
The Euclidean norm $\|x\|$ of the input vector $x$ is represented with high resolution $h$, e.g., with machine precision. 
Then, each entry $x_m$ of $x$ is separately quantized.
\item
One bit is used to encode the sign of $x_m$.
\item
Then, we encode the absolute value of the $m$-th entry $x_m$. Since $\|x\|$ is transmitted with high precision, we can focus on the scaled value:
\beq
\xi_m\dfz \frac{|x_m|}{\|x\|}\in [0,1].
\eeq
The interval $[0,1]$ is partitioned into $L$ equal-size intervals --- see the illustrative example in Fig.~\ref{fig:randq}. 
The size of each interval is:
\beq
\vartheta\dfz \frac{1}{L},
\eeq
such that the intervals' endpoints can be accordingly represented as:
\beq
y_0=0, ~~~ y_1=\vartheta,~~~y_2=2\vartheta,~~~\ldots,~~~y_{L}=1.
\label{eq:endpoints}
\eeq
In order to avoid confusion, we stress that the quantization scheme will require to transmit one of the $L+1$ indices corresponding to the endpoints. This differs from classical quantization schemes where the index of the interval (instead of the endpoint) is transmitted. 
Accordingly, the bit-rate $r$ is equal to:
\beq
r=\log_2(L+1)\Leftrightarrow L=2^r - 1.
\eeq
\item
In view of \eqref{eq:endpoints}, the index of the (lower) endpoint of the interval the scaled entry $\xi_m$ belongs to, is computed as:
\beq
j(\xi_m)=\left\lfloor \frac{\xi_m}{\vartheta} \right\rfloor,
\eeq
and the corresponding endpoint is:
\beq
y(\xi_m)=j(\xi_m)\,\vartheta.
\eeq
Then, we {\em randomize} the quantization operation since we choose randomly to transmit the lower endpoint index $j(\xi_m)$ or the upper endpoint index $j(\xi_m)+1$. 
Specifically, the probability of transmitting one endpoint index is proportional to the distance of $\xi_m$ from that endpoint. In other words, the closer we are to one endpoint, the higher the probability of transmitting that endpoint will be. 
Formally, the random transmitted index $\bm{j}_{\rm{tx}}(\xi_m)$ is:
\beq
\bm{j}_{\rm{tx}}(\xi_m)=
\begin{cases}
j(\xi_m)+1,~~~\textnormal{with prob.} &\displaystyle{\frac{\xi_m - y(\xi_m)}{\vartheta}}
\\
j(\xi_m),~~~~~~~~\textnormal{otherwise}.
\end{cases}
\label{eq:Alistarh2}
\eeq
\item
Once the index $\bm{j}_{\rm{tx}}(\xi_m)$ is received, the unquantized value $\xi_m$ is rounded to the lower or upper endpoint depending on the realization of the transmitted index $\bm{j}_{\rm{tx}}(\xi_m)$, and then the information about the norm $\|x\|$ and the sign of $x_m$ is recovered, finally yielding the $m$-th component of the quantized vector $\bm{Q}(x)$:
\beq
[\bm{Q}(x)]_m=\|x\|\,{\sf sign}(x_m)\,\bm{j}(\xi_m)\,\vartheta,~~m=1,2,\ldots,M.
\label{eq:Alistarh1}
\eeq
\item
Accounting for the $h$ bits spent for representing the norm $\|x\|$ and the single bit for representing the sign of each of the $M$ entries of $x$, the total bit-rate is:
\beq
h + M\,(r+1).
\label{eq:overallbitbudget}
\eeq
\end{itemize}
It was shown in~\cite{AlistarhNIPS2017} that the value of the compression factor $\omega$ can be computed as:
\beq
\omega=\min\left\{
\frac{M}{L^2} , \frac{\sqrt{M}}{L}
\right\}.
\label{eq:omegainsight}
\eeq
Equation \eqref{eq:omegainsight} provides useful insight on the practical meaning of $\omega$ for the considered type of quantizers.
We see that, for fixed dimensionality $M$, the parameter $\omega$ decays exponentially fast with the number of bits ($\approx 2^{-2 r}$), whereas for fixed number of bits it grows as $\sqrt{M}$. 

In order to avoid misunderstanding, we stress that in the forthcoming treatment we will use interchangeably the terminology compression or quantization to indicate a general compression operator fulfilling Assumption~\ref{Compression operator}. 
Reference to a specific class of compression operators, such as the randomized quantizers in~\cite{AlistarhNIPS2017} that we have illustrated in this section, will be made when needed (e.g., in Sec.~\ref{sec:experiments}).

Before going ahead, it is necessary to make the following notational remark.
Since in our setting each individual agent $k$ will be allowed to employ a different compression operator $\bm{Q}_k(x)$, we will have possibly different compression parameters $\omega_k$, with their maximum value being denoted by:
\beq
\Omega\triangleq \max_{k=1,2,\ldots,N} \omega_k.
\label{eq:omegamax}
\eeq
We will show in the remainder of the article that the mean-square-error approaches $O(\mu)$ for small step-sizes, i.e., the algorithm is mean-square-error stable even in the presence of quantization errors and gradient noise. The derivations are demanding and challenging due to the nonlinear and coupled nature of the network dynamics, as is clear from the arguments in the appendices. Nevertheless, when all is said and done, we arrive at the reassuring conclusion that the diffusion strategy is able to learn well in quantized/compressed environments.

\section{Network Error Dynamics}
In this section we illustrate the main formalism that will be exploited to conduct our analysis. 
Since we are interested in computing the deviation of the ACTC iterates from the global minimizer $w^{\star}$, it is expedient to introduce the following centered variables:
\beq
\begin{cases}
\widetilde{\bm{w}}_{k,i} 
&\!\!\!\!\triangleq \bm{w}_{k,i} - w^{\star}\\
\widetilde{\bm{\psi}}_{k,i} 
&\!\!\!\!\triangleq \bm{\psi}_{k,i} - w^{\star}\\
\widetilde{\bm{q}}_{k,i} 
&\!\!\!\!\triangleq \bm{q}_{k,i} - w^{\star}\\
\end{cases}
\label{eq:centeredvar}
\eeq
It is also convenient to rewrite the adaptation step in order to make explicit the role of the {\em true} cost functions $J_k(w)$. 
To this end, we must exploit the {\em gradient noise} introduced in \eqref{gradNoiseDef}, which quantifies the discrepancy between the approximate and true gradients. 
Exploiting \eqref{gradNoiseDef}, the first line in \eqref{ACTC} can be rewritten as:
\beq
\bm{\psi}_{k,i} = \bm{w}_{k,i-1} - \mu_k\nabla J_k(w) - \mu_k \bm{n}_{k,i}(\bm{w}_{k,i-1}).
\label{eq:firstlineACTC2}
\eeq
Examining \eqref{eq:firstlineACTC2}, we see that the gradient noise contains an additional source of randomness given by its argument $\bm{w}_{k,i-1}=\sum_{\ell\in\mathcal{N}_k}a_{\ell k}\bm{q}_{\ell,i-1}$, whose randomness comes accordingly from the previous-step quantized iterates $\{\bm{q}_{\ell,i-1}\}_{\ell=1}^{N}$. 
We assume the following standard regularity properties for the gradient noise process. 
\begin{assumption}[Gradient Noise]
\label{Gradient noise process}
For all $k=1,2,\ldots,N$ and all $i>0$, the gradient noise meets the following conditions: 
\begin{align}
&\E\big[\bm{n}_{k,i}(\bm{w}_{k,i-1})\,
\big| \{\bm{q}_{\ell,i-1}\}_{\ell=1}^{N}
\big] = 0,
\label{eq:gradNoiseUnbiased} \\
&\E\big[\|\bm{n}_{k,i}(\bm{w}_{k,i-1})\|^2\,
\big| \{\bm{q}_{\ell,i-1}\}_{\ell=1}^{N}\big] \leq \beta^2_k\|\widetilde{\bm{w}}_{k,i-1}\|^2 + \sigma^2_{k},
\label{eq:gradNoiseBound}
\end{align}
for some  constants $\beta_k$ and $\sigma_k$.~\hfill$\square$ 
\end{assumption}

With reference to the actual gradient in (\ref{gradNoiseDef}), from the mean-value theorem one has (we recall that $\widetilde{\bm{w}}_{k,i-1}=\bm{w}_{k,i-1} - w^{\star}$)~\cite{Sayed}:
\begin{align}
\nabla J_k(\bm{w}_{k,i-1}) &= \nabla J_k(w^{\star}) \nonumber\\
&+  \left[\int_{0}^{1} \nabla^2 J_k(w^{\star} - t\widetilde{\bm{w}}_{k,i-1})dt\right]\,
\widetilde{\bm{w}}_{k,i-1},
\label{eq:meanValTheorem0}
\end{align}
where the integral of a matrix is intended to operate entrywise.
Introducing the bias of agent $k$,
\beq
b_k \triangleq \alpha_k\nabla J_k(w^{\star}),
\label{eq:biasdef}
\eeq
and the {\em Hessian matrix} of agent $k$:
\begin{equation}
    \bm{H}_{k,i-1} \triangleq \alpha_k\,\int_{0}^{1} \nabla^2 J_k(w^{\star} - t\widetilde{\bm{w}}_{k,i-1})dt,
\label{eq:Hessian}
\end{equation}
Eq. \eqref{eq:meanValTheorem0} yields:
\begin{equation}
\mu_k \nabla J_k(\bm{w}_{k,i-1}) = \mu \left( b_k +  \bm{H}_{k,i-1}\widetilde{\bm{w}}_{k,i-1}\right).
\label{eq:meanValTheorem}
\end{equation}
For notational convenience, the scaled step-size $\alpha_k$ has been embodied in the definitions of the bias and the Hessian matrix. Likewise, it is useful to introduce the {\em scaled} gradient noise vector:
\beq
\bm{s}_{k,i}\triangleq \alpha_k \, \bm{n}_{k,i}(\bm{w}_{k,i-1}).
\label{eq:gradnoisedeffirstappear}
\eeq
Using now \eqref{eq:centeredvar}, \eqref{eq:firstlineACTC2} and \eqref{eq:meanValTheorem} in \eqref{ACTC}, the ACTC recursion can be recast in the form:
\begin{equation}\label{centeredACTC}
    \begin{cases}
      \widetilde{\bm{\psi}}_{k,i} = (I_M - \mu \bm{H}_{k,i-1})\widetilde{\bm{w}}_{k,i-1} - \mu \,\bm{s}_{k,i} - \mu \,b_k\\
      \\
      \widetilde{\bm{q}}_{k,i} = \widetilde{\bm{q}}_{k,i-1} + \zeta \,\bm{Q}_k(\widetilde{\bm{\psi}}_{k,i} - \widetilde{\bm{q}}_{k,i-1}) \\
      \\
      \widetilde{\bm{w}}_{k,i} = \displaystyle{\sum_{\ell \in \mathcal{N}_k}} a_{\ell k}\widetilde{\bm{q}}_{\ell,i}
    \end{cases}
\end{equation} 
We remark that in the second equation of \eqref{centeredACTC}, the argument of the compression function is expressed in terms of centered variables by adding and subtracting $w^\star$.

We now manage to reduce \eqref{centeredACTC} to a simpler form.
First of all, the ACTC iterates $\widetilde{\bm{w}}_{k,i}$ are convex combinations of the quantized iterates $\{\widetilde{\bm{q}}_{k,i}\}$, implying that the characterization of the mean-square behavior of $\widetilde{\bm{q}}_{k,i}$ will enable immediate characterization of the mean-square behavior of $\widetilde{\bm{w}}_{k,i}$. It is therefore convenient to incorporate the third step of the ACTC strategy into the first step, and focus on the behavior of $\widetilde{\bm{q}}_{k,i}$, obtaining:
\begin{numcases}{\!\!\!\!\!}
\widetilde{\bm{\psi}}_{k,i} = (I_M - \mu \bm{H}_{k,i-1})
\displaystyle{\sum_{\ell \in \mathcal{N}_k}} a_{\ell k}\widetilde{\bm{q}}_{\ell,i}
 \!-\! \mu \,\bm{s}_{k,i} \!-\! \mu \,b_k
\nonumber\\
\label{eq:prefinalACTC}
\\
\widetilde{\bm{q}}_{k,i} = \widetilde{\bm{q}}_{k,i-1} + \zeta \, \bm{Q}_k(\widetilde{\bm{\psi}}_{k,i} - \widetilde{\bm{q}}_{k,i-1} )
\nonumber
\end{numcases}
Moreover, it is convenient to introduce the difference variable:
\beq
\bm{\delta}_{k,i}\triangleq \widetilde{\bm{\psi}}_{k,i}-\widetilde{\bm{q}}_{k,i-1}=\bm{\psi}_{k,i}-\bm{q}_{k,i-1}.
\eeq
Subtracting $\widetilde{\bm{q}}_{k,i-1}$ from the first equation in \eqref{eq:prefinalACTC}, we finally obtain:
\begin{numcases}{\!\!\!}
\!\!\bm{\delta}_{k,i} = (I_M - \mu\bm{H}_{k,i-1})
\displaystyle{\sum_{\ell \in \mathcal{N}_k}} a_{\ell k}\widetilde{\bm{q}}_{\ell,i}
- \widetilde{\bm{q}}_{k,i-1} 
\!-\! \mu\bm{s}_{k,i} \!-\! \mu\, b_k
\nonumber\\
\label{eq:finalACTC}\\
\!\!\widetilde{\bm{q}}_{k,i}= \widetilde{\bm{q}}_{k,i-1} + \zeta\,\bm{Q}_k(\bm{\delta}_{k,i})
\nonumber
\end{numcases}

\subsection{Recursions in Extended Form}
Since we are interested in a network-oriented analysis, it is useful to introduce a notation where the $N$ agents' vectors of size $M\times 1$ are stacked into the following $M N\times 1$ vectors:
\begin{equation}\label{netStackVars}
    \bm{\delta}_{i} \triangleq
    \begin{bmatrix}
        \bm{\delta}_{1,i}\\
        \bm{\delta}_{2,i}\\
        \vdots \\
        \bm{\delta}_{N,i}
    \end{bmatrix},
    \widetilde{\bm{q}}_{i} \triangleq
    \begin{bmatrix}
        \widetilde{\bm{q}}_{1,i}\\
        \widetilde{\bm{q}}_{2,i}\\
        \vdots \\
        \widetilde{\bm{q}}_{N,i}
    \end{bmatrix},
    \bm{s}_{i} \triangleq
    \begin{bmatrix}
        \bm{s}_{1,i}\\
        \bm{s}_{2,i}\\
        \vdots \\
        \bm{s}_{N,i}
    \end{bmatrix},
    b \triangleq
    \begin{bmatrix}
        b_{1}\\
        b_{2}\\
        \vdots \\
        b_{N}
    \end{bmatrix}.
\end{equation}
Likewise, it is useful to consider the extended compression operator $\bm{\mathcal{Q}}(\cdot)$ that applies the compression operation to each $M \times 1$ block of its input, and stacks the results as follows:
\begin{equation}\label{netStackQuant}
    \bm{\mathcal{Q}}(\bm{\delta}_i) \triangleq
    \begin{bmatrix}
            \bm{Q}_1(\bm{\delta}_{1,i}) \\
            \bm{Q}_2(\bm{\delta}_{2,i}) \\
            \vdots \\
            \bm{Q}_N(\bm{\delta}_{N,i})
        \end{bmatrix}\,.
\end{equation}
Finally, in order to express the recursion \eqref{centeredACTC} in terms of the joint evolution of the extended vectors we need to introduce the extended matrices:
\beq
\mathcal{A} \triangleq A \otimes I_M,~~
\bm{\mathcal{H}}_{i-1} \triangleq {\sf diag}\{ \bm{H}_{1,i-1}, \bm{H}_{2,i-1}, ..., \bm{H}_{N,i-1}\},
\label{hCal}
\eeq
where $\otimes$ denotes the Kronecker product.
It is now possible to describe compactly the ACTC strategy of the individual agents in \eqref{centeredACTC} as:
\begin{numcases}{\!\!\!\!\!}
\!\!\bm{\delta}_{i} = [(I_{MN} - \mu\,\bm{\mathcal{H}}_{i-1})\,\mathcal{A}^{\top}  -  I_{MN}]\widetilde{\bm{q}}_{i-1} 
\!-\! \mu\bm{s}_{i} \!-\! \mu\, b
\nonumber\\
\label{eq:netACTC}\\
\!\!\widetilde{\bm{q}}_{i} = \widetilde{\bm{q}}_{i-1} + \zeta\,\bm{\mathcal{Q}}(\bm{\delta}_i)
\nonumber
\end{numcases}

\subsection{Network Coordinate Transformation}
\label{sec:NCT}
By means of a proper linear transformation of the network evolution it is possible to separate the two fundamental mechanisms that characterize the learning behavior of the ACTC strategy. 
The first mechanism characterizes the {\em coordinated evolution} enabled by the social learning phenomenon. This is a desired behavior that will be critical to let each individual agent agree and converge to a small neighborhood of the global minimizer $w^\star$.
In comparison, the second mechanism represents the departure of the agents' evolution from the coordinated evolution, which arises from the distributed nature of the system (agents need some time to reach agreement through successive local exchange of information). We will see later how the ACTC diffusion strategy blends these two mechanisms so as to achieve successful learning.

The enabling tool to develop the aforementioned network transformation is the Jordan canonical decomposition of the (transposed) combination matrix $A$~\cite{Johnson-Horn}:
\footnote{More often, the similarity transform in \eqref{aJordan} is written as $VJ_{\rm{tot}} V^{-1}$. We opt for the alternative form in \eqref{aJordan} since, as we will see soon, with this choice the matrix $V$ will correspond to the {\em direct} network coordinate transformation that will be relevant to our treatment (and its inverse $V^{-1}$ will be the corresponding {\em inverse} transformation).}
\beq
A^{\top} \triangleq V^{-1} J_{\rm{tot}} V.
\label{aJordan}
\eeq
The matrix $J_{\rm{tot}}$ is the Jordan matrix of $A^{\top}$ (equivalently, of $A$, since $A$ and $A^{\top}$ are similar matrices~\cite{Johnson-Horn}) arranged in canonical form, i.e., made up of the Jordan blocks corresponding to the eigenvalues of $A$ as detailed in Appendix~\ref{app:collection}~\cite{Johnson-Horn}. In particular, the single\footnote{The existence of a single (and maximum magnitude) eigenvalue equal to $1$ is guaranteed by Assumptions~\ref{Strong Connectivity} and~\ref{Stochastic combination matrix}.} eigenvalue equal to $1$ corresponds to a $1\times 1$ block, and the other Jordan blocks can be collected into the $(N-1)\times (N-1)$ reduced Jordan matrix $J$ --- see \eqref{eq:redJordan}:
\beq
J_{\rm{tot}} = 
\begin{bmatrix}
        1 & 0
        \\
        0 & J
\end{bmatrix}.
\label{eq:JordanMat}
\eeq
The columns of $V^{-1}$ collect the generalized right-eigenvectors of $A^{\top}$ (hence, generalized left-eigenvectors of $A$)~\cite{GolubWilkinsonSIAM1976,Bronson}. Likewise, the rows of $V$ collect the generalized left-eigenvectors of $A^{\top}$. 
Recalling that $\pi$ is a right-eigenvector of $A$, and $\mathds{1}_N$ a left-eigenvector of $A$, the matrices $V$ and $V^{-1}$ can be conveniently block-partitioned as follows:
\beq
        V =         
    \begin{bmatrix}
        \pi^{\top} \\
\\
        V_R
        \end{bmatrix}
      ,~~~~~~
        V^{-1} = 
        \begin{bmatrix}
        \mathds{1}_N & V_L 
        \end{bmatrix},
\label{jordan}
\eeq
where the subscript $R$ is associated with generalized right-eigenvectors of $A$. The same applies to subscript $L$ as regards generalized left-eigenvectors.
 
We are now ready to detail the network coordinate transformation relevant to our analysis. 
To this end, we introduce the extended matrix:
\beq
    \mathcal{V} \triangleq V \otimes I_M\,, 
\label{otherCal}
\eeq
and the {\em transformed} extended vector $\widehat{\bm{q}}_i$:
\beq
    \widehat{\bm{q}}_i\triangleq\mathcal{V}\,\widetilde{\bm{q}}_{i} =
        \begin{bmatrix}
        (\pi^\top \otimes I_M)\,\widetilde{\bm{q}}_{i} \\
        (V_R \otimes I_M)\,\widetilde{\bm{q}}_{i}
        \end{bmatrix}
=    
        \begin{bmatrix}
        \bar{\bm{q}}_{i} \\
        \widecheck{\bm{q}}_{i}
        \end{bmatrix}.
    \label{eq:hatq}
\eeq
As we see, the transformed vector $\widehat{\bm{q}}_i$ has been partitioned in two blocks: an $M\times 1$ vector $\bar{\bm{q}}_i$ and an $M(N-1)\times 1$ vector $\widecheck{\bm{q}}_i$. These two vectors admit a useful physical interpretation. 
Vector $\bar{\bm{q}}_i$ is a linear combination, through the Perron weights, of the $N$ vectors composing $\widetilde{\bm{q}}_i$, i.e., of the vectors corresponding to all agents. As we will see, such combination reflects a ``coordinated'' evolution that will apply to all agents. 
In contrast, vector $\widecheck{\bm{q}}_i$ is representative of the departure of the individual agent's behavior from the coordinated behavior. In the following, we will sometimes refer to $\bar{\bm{q}}_i$ as the {\em coordinated-evolution component}, and to $\widecheck{\bm{q}}_i$ as the {\em network-error component}. 
It is worth noticing that the coordinated-evolution component $\bar{\bm{q}}_i$ is real-valued, whereas the network-error component $\widecheck{\bm{q}}_i$ is in general {\em complex-valued}, since the matrix $V_R$ can contain complex-valued eigenvectors.

In order to highlight the role of the aforementioned two components on the individual agents, we can apply the inverse transformation $\mathcal{V}^{-1}$ to $\widehat{\bm{q}}_i$, obtaining:
\beq
\widetilde{\bm{q}}_{k,i}=
\bar{\bm{q}}_{i}+\underbrace{( [V_L]_k\otimes I_M)}_{\dfz \mathcal{T}_k} \widecheck{\bm{q}}_i,
\label{eq:individualvsnetwork}
\eeq
where $[V_L]_k$ denotes the $k$-th row of matrix $V_L$.
Equation \eqref{eq:individualvsnetwork} reveals that agent $k$ progresses over time by combining the coordinated evolution $\bar{\bm{q}}_i$ (which is equal for {\em all} agents) and a perturbation vector $\mathcal{T}_k\, \widecheck{\bm{q}}_{i}$, which quantifies the specific discrepancy of agent $k$ (since matrix $\mathcal{T}_k$ depends on $k$) from the coordinated behavior. 
From the distributed optimization perspective, the goal is to reach agreement among all agents and, hence, it is necessary that the perturbation term is washed out as time elapses, letting all agents converge to the same coordinated behavior. 
Establishing that this is the case will be the main focus of our analysis.

For later use, it is convenient to introduce also the transformed versions of $\bm{\delta}_i$, $\bm{s}_i$ and $b$:
\begin{align}
&\widehat{\bm{\delta}}_i\triangleq\mathcal{V}\bm{\delta}_{i} =
\begin{bmatrix}
(\pi^\top \otimes I_M)\bm{\delta}_{i} \\
(V_R \otimes I_M)\bm{\delta}_{i}
\end{bmatrix}
=
\begin{bmatrix}
\bar{\bm{\delta}}_{i} \\
\widecheck{\bm{\delta}}_{i}
\end{bmatrix},
\label{eq:psiTrans}
\\
&\widehat{\bm{s}}_i\triangleq\mathcal{V}\bm{s}_{i} =
\begin{bmatrix}
(\pi^\top \otimes I_M)\bm{s}_{i} \\
(V_R \otimes I_M)\bm{s}_{i}
\end{bmatrix}
=
\begin{bmatrix}
\bar{\bm{s}}_{i} \\
        \widecheck{\bm{s}}_{i}
\end{bmatrix},
\label{sTrans}
\\
&\widehat{b}\triangleq\mathcal{V} b =
\begin{bmatrix}        
(\pi^\top \otimes I_M) b \\
(V_R \otimes I_M) b
\end{bmatrix}
=
\begin{bmatrix}
0 \\
\widecheck{b}
\end{bmatrix}.
\label{bTrans}
\end{align} 
The zero entry in the transformed bias vector arises from the fact that, in view of \eqref{eq:biasdef}, we have:
\beq
(\pi^{\top} \otimes I_M) b=
\sum_{k=1}^N \pi_k b_k=
\sum_{k=1}^{N} p_k \, \nabla J_k(w^\star)=0,
\eeq
since the Perron-weighted sum of the gradients computed at the limit point $w^{\star}$ corresponds to the {\em exact} minimizer of the global cost function in \eqref{eq:globalCost}.

\section{Mean-Square Stability}
In order to assess the goodness of an individual agent's estimate $\bm{w}_{k,i}$ we will focus on the mean-square-deviation:
\beq
\E\|\widetilde{\bm{w}}_{k,i}\|^2=\E\|\bm{w}_{k,i} - w^{\star}\|^2.
\label{eq:MSEfirstdef}
\eeq
As we explained before, it is instrumental to work in terms of the quantized iterates $\bm{q}_{k,i}$ and, more precisely, in terms of the transformed vectors $\widehat{\bm{q}}_i=\mathcal{V}\,\widetilde{\bm{q}}_i$.
In order to characterize the mean-square evolution of the transformed vectors, it is particularly convenient to adopt the formalism of the energy operators introduced in~\cite{ChenSayedTIT2015part1}.

\begin{definition}[Average Energy Operator]
Given a random vector:
\beq
\bm{x}=
\begin{bmatrix}
\bm{x}_1\\
\vdots\\
\bm{x}_N
\end{bmatrix},
\eeq
where each $\bm{x}_k$, for $k=1,2,\ldots, N$ is an $M\times 1$ vector, we consider the operator: 
\beq
\mathscr{P}[\bm{x}]=\begin{bmatrix}
\E\|\bm{x}_1\|^2
\\
\vdots\\
\E\|\bm{x}_N\|^2
\end{bmatrix}.
\label{eq:avenopdef}
\eeq
\hfill$\square$
\end{definition}

In particular, when we apply the average energy operator in \eqref{eq:avenopdef} to one of our transformed vectors, e.g., to $\widehat{\bm{q}}_i$ in \eqref{eq:hatq}, we obtain the following block decomposition:
\beq
\mathscr{P}[\widehat{\bm{q}}_i]=
\begin{bmatrix}
\E\|\bar{\bm{q}}_i\|^2
\\
\\
\mathscr{P}[\widecheck{\bm{q}}_i]
\end{bmatrix},
\label{eq:avenopbarcheck}
\eeq
where $\mathscr{P}[\widecheck{\bm{q}}_i]$ is an $(N-1)\times 1$ vector. 
Examining the time evolution of the energy vectors in \eqref{eq:avenopbarcheck} is critical because the individual agents' errors $\E\|\bm{q}_{k,i}\|^2$ can be related to these energy vectors through the inverse network transformation $\mathcal{V}^{-1}$ in \eqref{eq:individualvsnetwork}. 
In particular, we will be able to show that the quantity $\E\|\bar{\bm{q}}_i\|^2$ plays a domineering role in determining the (common) individual agents' steady-state mean-square behavior, whereas the quantity $\mathscr{P}[\widecheck{\bm{q}}_i]$ plays the role of a network error quantity that dies out during the transient phase.

\begingroup
\renewcommand{\arraystretch}{2.5}
\begin{table*}[t]
  \begin{center}
    \begin{tabular}{| >{\columncolor[gray]{0.8}}l|l|} 
      \hline
{\bf Step-sizes} $\mu_k$ &
$\mu=\displaystyle{\max_{k=1,2,\ldots,N} \mu_k},$
~~~~~~~~~~~~~~~~~~~~~~~~~$\alpha_k=\mu_k/\mu,$~~~~~~~~~~~
$C_{\alpha}={\sf diag}(\alpha_1^2,\alpha_2^2,\ldots,\alpha_N^2)$~~~~~~~~~~~~~~~~~
\\
\hline
{\bf Combination matrix} $A$ &
$\pi$ is the Perron eigenvector$, A \pi=\pi,$
~~~~~~~~~~~~~~~~~~~~~~~~~~~~~~~
$p=[\alpha_1 \pi_1,\alpha_2 \pi_2,\ldots,\alpha_N \pi_N]^{\top}$
~~~~~~
\\ 
&Jordan form: $A^{\top}=V^{-1} J_{\rm{tot}} V,$
~~~~~~~~~
\begingroup
\renewcommand{\arraystretch}{1}
$V =\begin{bmatrix}\pi^{\top} \\ \\V_R\end{bmatrix},$
\endgroup
~~~~~~~~~
$\lambda_2$ is the $2$nd largest magnitude eigenvalue of $A$
\\
&$V_{2R}$ is the $(N-1)\times N$ matrix with $(\ell,k)$-entry equal to the squared magnitude of the corresponding entry in $V_R$
\\
\hline
{\bf Convexity/Lipschitz} &
$\nu$ is the global-strong-convexity constant in \eqref{sqNablaStrConvex}$,$
~~~~~~~~~~~~~~~~~~~$\sigma_{11}, \sigma_{12}, \sigma_{21},\sigma_{22}$ are the constants in Lemma~\ref{Characterization of the block matrix}
\\
\hline
{\bf Gradient noise} &
Bounding constants at agent $k$: $\{\beta_k\}$ and $\{\sigma^2_{k}\}$ --- see \eqref{eq:gradNoiseBound}\\
&$C_\beta={\sf diag}(\beta_1^2,\beta_2^2,\ldots,\beta_N^2),$ 
~~~~~~~~~~~~~~~~~~~~~~~~~~~~~~~~~~~~~~~~~$C_\sigma={\sf diag}(\sigma_{1}^2,\sigma_{2}^2,\ldots,\sigma_{N}^2)$
\\
\hline
{\bf Bias} $b_k= \alpha_k \nabla J_k(w^{\star})$& 
$\widecheck{b}=(V_R\otimes I_M) b,$~~~~~~
\begingroup
\renewcommand{\arraystretch}{1}
$\mathscr{P}[\,\widecheck{b}\,]=
\Big[
\|\widecheck{b}_1\|^2, \|\widecheck{b}_2\|^2,\ldots, \|\widecheck{b}_{N-1}\|^2
\Big]^{\top}$ (the equality holds since the bias is not random)
\endgroup
\\
\hline
{\bf Network error matrix} &
$J=\Lambda + U$ is the reduced Jordan matrix in \eqref{eq:Jordanrep}$,$~~~~~~~~~~~~~~~~~
$E_0=\left( (1-\zeta)I_{N-1} + \zeta\,
\displaystyle{\frac{\Lambda\Lambda^*}{|\lambda_2|}}\right) + \displaystyle{\frac{2\zeta}{1-|\lambda_2|}}\,U$
\\	
\hline
{\bf Compression parameters} &
$\omega_k$ is the compression factor of agent $k,$~~~~ 
$\bar{\Delta} =\|V^{-1}\|^2 \!\!\!\displaystyle{\max_{k=1,2,\ldots,N} \pi^2_k \,\omega_k},
~~~~
\widecheck{\Delta} =\|V^{-1}\|^2 \!\!\!\displaystyle{\max_{\substack{\ell=2,3,\ldots,N\\k=1,2,\ldots,N}}} 
|v_{\ell k}|^2 \,\omega_k$
\\
\hline
    \end{tabular}
\vspace*{5pt}
    \caption{Useful symbols used throughout the article.}
        \label{tab:Notation}
  \end{center}
\end{table*}

\begin{table*}[t]
  \begin{center}
    \begin{tabular}{|l|l|l|} 
      \hline
$[T_s]_{11}=\mu^2\,\|V^{-1}\|^2\,\sum_{\ell=1}^N \pi_{\ell} \,\alpha^2_{\ell}\,\beta_{\ell}^2 $  &
$[T_\delta]_{11}=2\,\mu^2\,\sigma_{11}^2$  &
$[T_q]_{11}=1 - \mu\,\zeta\,\nu$  
\\
\hline
$[T_s]_{12}=\mu^2\,\|V^{-1}\|^2\,\sum_{\ell=1}^N \pi_{\ell}\,\alpha^2_{\ell}\,\beta_{\ell}^2\, \mathds{1}_{N-1}^\top$  &
$[T_\delta]_{12}=2\,\mu^2\, \sigma_{12}^2\,\mathds{1}_{N-1}^\top$  &
$[T_q]_{12}=\mu\,\zeta\,\displaystyle{\frac{\sigma_{12}^2}{\nu}} \,\mathds{1}_{N-1}^{\top}$ 
\\	\hline
$[T_s]_{21}=\mu^2\,N\|V^{-1}\|^2\,V_{2R} \,C_\alpha\,C_\beta\, \mathds{1}_{N}$  &
$[T_\delta]_{21}=8\,\mu^2\, \sigma_{21}^2 \,\mathds{1}_{N-1}$  &
$[T_q]_{21}= \mu^2\,\zeta\,    \displaystyle{\frac{6\,\sigma_{21}^2}{1-|\lambda_2|}}\,\mathds{1}_{N-1}$ 
\\	\hline
$[T_s]_{22}=\mu^2\,N\|V^{-1}\|^2\,V_{2R} \,C_\alpha\,C_\beta\, \mathds{1}_{N}\mathds{1}_{N-1}^\top$  &
$[T_\delta]_{22}=8(I_{N-1}+U) + 8\,\mu^2\, \sigma_{22}^2\,\mathds{1}_{N-1}\mathds{1}_{N-1}^{\top}$  &
$[T_q]_{22}=E_0+\mu^2\,\zeta\,\displaystyle{\frac{6\,\sigma_{22}^2 \mathds{1}_{N-1}\mathds{1}_{N-1}^{\top}}{1-|\lambda_2|}}$
\\	\hline
$\bar{x}_s=\mu^2\,\sum_{\ell=1}^N \pi_{\ell} \,\alpha_{\ell}^2\, \sigma^2_{\ell}$  &
$\bar{x}_\delta=0$  &
$\bar{x}_q=0$  
\\	\hline
$\widecheck{x}_s=\mu^2\,N V_{2R}\,C_\alpha\,C_\sigma \, \mathds{1}_{N-1}$ &
$\widecheck{x}_\delta=8\,\mu^2\,\mathscr{P}[\,\widecheck{b}\,]$ &
$\widecheck{x}_q=\mu^2\,\zeta\,\displaystyle{\frac{6}{1-|\lambda_2|}}\,\mathscr{P}[\,\widecheck{b}\,]$  
\\	\hline
    \end{tabular}
\vspace*{5pt}
    \caption{Useful matrices and vectors appearing in Lemmas~\ref{lem:gradnoisetx}, \ref{lem:qnoisetx}, and~\ref{lem:qstaterecextended}. The symbols used in the present table are defined in Table~\ref{tab:Notation}.}
        \label{tab:TransferMaTable}
  \end{center}
\end{table*}

\endgroup

We start with three lemmas that characterize the interplay over time (in terms of energy) of three main quantities in the network transformed domain: the gradient noise $\widehat{\bm{s}}_i$, the quantization error $\widehat{\bm{\delta}}_i$ , and the quantized iterates $\widehat{\bm{q}}_i$. 

The first lemma relates the gradient noise to the quantized iterates.
\begin{lemma}[Gradient Noise Energy Transfer]
\label{lem:gradnoisetx}
The average energy of the transformed gradient noise extended vector $\widehat{\bm{s}}_i$ evolves over time according to the following inequality:
\beq
\boxed{
\mathscr{P}[\mu\,\widehat{\bm{s}}_i]\preceq T_s \, \mathscr{P}[\widehat{\bm{q}}_{i-1}] + x_s
}
\label{eq:lemma1}
\eeq
where the transfer matrix $T_s$ and the driving vector $x_s=[\bar{x}_s,\widecheck{x}_s]^{\top}$ are defined in Table~\ref{tab:TransferMaTable}.
\end{lemma}
\begin{IEEEproof}
See Appendix~\ref{Gradient noise bound lemma proof}.
\end{IEEEproof}

The second lemma relates the quantization error to the quantized iterates and to the gradient noise. 
\begin{lemma}[Quantization Error Energy Transfer]
\label{lem:qnoisetx}
The average energy of the transformed quantization-error extended vector $\widehat{\bm{\delta}}_i$ evolves over time according to the following inequality:
\beq
\boxed{
\mathscr{P}[\widehat{\bm{\delta}}_i]\preceq T_\delta \, \mathscr{P}[\widehat{\bm{q}}_{i-1}] 
+\mathscr{P}[\mu\,\widehat{\bm{s}}_i]
+ x_\delta 
}
\label{eq:lemma2}
\eeq
where the transfer matrix $T_\delta$ and the driving vector $x_\delta=[\bar{x}_\delta,\widecheck{x}_\delta]^{\top}$ are defined in Table~\ref{tab:TransferMaTable}.

\end{lemma}
\begin{IEEEproof}
See Appendix~\ref{Quantization noise bound lemma proof}.
\end{IEEEproof}

The third lemma relates the quantized iterates to the quantization error and the gradient noise.
\begin{lemma}[Quantized Iterates Energy Transfer]
\label{lem:qstaterecextended}
Let
\beq
\mu \,\zeta< \frac{2}{\eta+\nu},
\label{eq:mustabnew}
\eeq
where $\nu$ is the global-strong-convexity constant introduced in \eqref{sqNablaStrConvex} and $\eta$ is the average Lipschitz constant in \eqref{eq:LipCvxConst}. 
Let
\beq
\Delta= 
\begin{bmatrix}
\bar{\Delta}
\\
\\
\widecheck{\Delta}\mathds{1}_{N-1}
\end{bmatrix},
\eeq
where the constants $\bar{\Delta}$ and $\widecheck{\Delta}$ are defined in Table~\ref{tab:Notation}.
Then, the average energy of the transformed extended vector $\widehat{\bm{q}}_i$ evolves over time according to the following inequality:
\beq
\boxed{
\begin{array}{ll}
&\mathscr{P}[\widehat{\bm{q}}_i]\preceq T_q \, \mathscr{P}[\widehat{\bm{q}}_{i-1}]\\
\\
&+\zeta^2
\Delta\,\mathds{1}_N^{\top}
\,\mathscr{P}[\widehat{\bm{\delta}}_i]
+ \zeta^2\,\mathscr{P}[\mu\,\widehat{\bm{s}}_i]
+x_q
\end{array}
}
\label{eq:lemma3}
\eeq
where the transfer matrix $T_q$ and the driving vector $x_q=[\bar{x}_q,\widecheck{x}_q]^{\top}$ are defined in Table~\ref{tab:TransferMaTable}.
\end{lemma}
\begin{IEEEproof}
See Appendix~\ref{Recursion extended lemma proof}.
\end{IEEEproof}

Combining Lemmas~\ref{lem:gradnoisetx}, \ref{lem:qnoisetx}, and~\ref{lem:qstaterecextended} we arrive at a recursion on the quantized iterates $\widehat{\bm{q}}_i$, as stated in the next theorem. 
Unless otherwise specified, all matrices, vectors and constants in the statement of the theorem can be found in Tables~\ref{tab:Notation} and~\ref{tab:TransferMaTable}. 
\begin{theorem}[Recursion on Quantized Iterates]
\label{Quantized state recursion} 
Let 
\beq
T=
\underbrace{
\begin{bmatrix}
\tau&\tau_{12}\mathds{1}_{N-1}^{\top}
\\
\\
0& E
\end{bmatrix}}_{T_0}
+
v_{\mu,\zeta}\,\mathds{1}_N^{\top},
\label{eq:Trepresfirstappear}
\eeq
where:
\beq
\tau\triangleq 1-\mu\,\zeta\,\nu,\quad\tau_{12}\triangleq 16\,\zeta^2\,\bar{\Delta} + \mu\,\zeta\, \frac{\sigma_{12}^2}{\nu},
\label{eq:tautau12firstappear}
\eeq
\beq
E=E_0 + 16\,\zeta^2\,\widecheck{\Delta}\,\mathds{1}_{N-1} \mathds{1}_{N-1}^{\top},
\label{eq:Edefintheorem}
\eeq
and
\beq
v_{\mu,\zeta}=
\phi\,\begin{bmatrix}
\mu^2\,\zeta^2
\\
\\
\mu^2\,\zeta\,\mathds{1}_{N-1}
\end{bmatrix},
\label{eq:varepsilondef00}
\eeq
with $\phi$ being a positive scalar that embodies the constants appearing in the $\mu^2$-terms of the transfer matrices $T_s$, $T_{\delta}$ and $T_q$ in Table~\ref{tab:TransferMaTable}. The evaluation of $\phi$ is rather cumbersome and is detailed in Appendix~\ref{app:proofTheorem1}.
Let also
\beq
x=
\begin{bmatrix}
\bar{x}\\
\widecheck{x}
\end{bmatrix}
=
x_q + \zeta^2\Delta\,\mathds{1}_N^{\top}\, (x_\delta + x_s) + \zeta^2 x_s,
\label{eq:Tandx}
\eeq
where the driving vectors $x_s$, $x_{\delta}$ and $x_q$ are defined in Table~\ref{tab:TransferMaTable}.
Then, the average energy of the extended vector $\widehat{\bm{q}}_i$ obeys the following inequality:
\beq
\boxed{
\mathscr{P}[\widehat{\bm{q}}_{i}]
\preceq 
T \, \mathscr{P}[\widehat{\bm{q}}_{i-1}] + x
}
\label{eq:systemRecursion}
\eeq
\end{theorem}

\begin{IEEEproof}
See Appendix~\ref{app:proofTheorem1}. 
\end{IEEEproof}

By inspection of \eqref{eq:Trepresfirstappear}, we see that the transfer matrix $T$ can be written as the sum of an upper-diagonal matrix $T_0$ and a rank-one perturbation of order $\mu^2$. 
In particular, the upper-diagonal structure of $T_0$ implies that the evolution relating the network error components $\mathscr{P}[\widecheck{\bm{q}}_{i-1}]$ to $\mathscr{P}[\widecheck{\bm{q}}_{i}]$ takes place only through matrix $E$. Accordingly, such matrix will be referred to as the network error matrix. Moreover, it is worth noticing that $E$ is independent of the step-size $\mu$.

It is tempting to conclude from \eqref{eq:Trepresfirstappear} that the rank-one perturbation can be neglected as $\mu\rightarrow 0$. Were the matrix $T_0$ independent of $\mu$, this conclusion would be obvious. 
However, since $T_0$ does depend upon $\mu$, proving that for small $\mu$ the recursion in Theorem~\ref{Quantized state recursion} can be examined by replacing $T$ with $T_0$ is not necessarily true. 
We will be able to show that this is actually the case, but to this end we need to carry out the demanding technical analysis reported in the appendices. 

Nevertheless, to gain insight on how recursion \eqref{eq:systemRecursion} and the quantities in Table~\ref{tab:TransferMaTable} are relevant to the mean-square behavior of the ACTC strategy, let us simply assume for now that we can replace $T$ with $T_0$. Under this assumption, by developing the inequality recursion \eqref{eq:systemRecursion} we would arrive at the following inequality:
\beq
\label{eq:simplersystemRecursion}
\mathscr{P}[\widehat{\bm{q}}_{i}]
\preceq 
T_0^i \, \mathscr{P}[\widehat{\bm{q}}_{0}] + \sum_{j=0}^{i-1}T_0^j x.
\eeq
Assume that $T_0$ is stable. Then, from \eqref{eq:simplersystemRecursion} we would have:
\begin{align}
\label{eq:limitsimplersystemRecursion}
&\limsup_{i\rightarrow\infty}\mathscr{P}[\widehat{\bm{q}}_{i}]=
\limsup_{i\rightarrow\infty}
\begin{bmatrix}
\E\|\bar{\bm{q}}_i\|^2
\\
\\
\mathscr{P}[\widecheck{\bm{q}}_i]
\end{bmatrix}
\nonumber\\
&    \preceq 
(I - T_0)^{-1}  x=\begin{bmatrix}
       O(1/\mu) & O(1/\mu)\, \mathds{1}_{N-1}^{\top} \\
       \\
       0 & O(1)\,\mathds{1}_{N-1} \mathds{1}_{N-1}^{\top}
    \end{bmatrix}\,
    \begin{bmatrix}
    \bar{x}
    \\
    \\
    \widecheck{x}
    \end{bmatrix},
\end{align}
where we exploited the upper triangular shape of $T_0$ to evaluate the inverse $(I-T_0)^{-1}$. 
Examining \eqref{eq:Tandx} and Table~\ref{tab:TransferMaTable}, we see that all the entries of $x$ scale as $\mu^2$. 
Accordingly, from \eqref{eq:limitsimplersystemRecursion} we obtain the following important conclusions regarding the mean-square stability of the ACTC strategy. 

The term $\E\|\bar{\bm{q}}_i\|^2$ scales as $O(\mu)$, whereas the energy of the network error component $\E\|\widecheck{\bm{q}}_i\|^2$ is a higher-order term scaling as $O(\mu^2)$.

Since all entries in $x$ scale as $\mu^2$, at the leading order in $\mu$ the behavior of $\E\|\bar{\bm{q}}_i\|^2$ is determined by the $(1,1)$-entry of $(I-T_0)^{-1}$ multiplied by $\bar{x}$, i.e., the first entry in $x$.
Inspecting Table~\ref{tab:TransferMaTable}, we see that:
\begin{align}
\bar{x}&=\bar{x}_q + \bar{x}_s +
\zeta^2\Delta\mathds{1}_N^{\top}
\begin{bmatrix}
\bar{x}_\delta + \bar{x}_s\\
\widecheck{x}_\delta + \widecheck{x}_s
\end{bmatrix}\nonumber\\
&=\bar{x}_s +
\zeta^2\Delta\mathds{1}_N^{\top}
\begin{bmatrix}
\bar{x}_s\\
\widecheck{x}_\delta + \widecheck{x}_s
\end{bmatrix}.
\label{eq:insights}
\end{align}
Equation \eqref{eq:insights} contains useful information as regards the specific interplay between the compression stage and the mean-square behavior of the ACTC strategy. 
In the absence of compression, the contribution due to $\Delta$ disappears (since with unquantized data $\omega_k=0$ for all agents), and the only relevant term would then be the first component $\bar{x}_s$ corresponding to the driving vector of the gradient noise energy transfer in Lemma~\ref{lem:gradnoisetx}. This behavior matches perfectly the behavior of the classical (i.e., unquantized) ATC strategy.

On the other hand, when compression comes into play, another term arises, which mixes both components of $x_{\delta}$ and $x_s$, owing to the $\Delta\mathds{1}_N^{\top}$ perturbation in \eqref{eq:insights}. 
In particular, in the additional term due to quantization we can recognize the following relevant quantities. 
First, we see that the gradient noise component contributes again to the ACTC error through the quantity $\zeta^2\,\bar{\Delta}\,\bar{x}_s$. 

Second, and differently from the classical ATC, also the network component $\widecheck{x}_s$ of the gradient noise is now injected into the ACTC error by the quantization mechanism.
Finally, there is a term depending on the driving vector $\widecheck{x}_{\delta}$ pertaining to the quantization error examined in Lemma~\ref{lem:qnoisetx}. 
Notably, from Table~\ref{tab:TransferMaTable} we see that $\widecheck{x}_{\delta}$ depends on the bias term $\mathscr{P}[\,\widecheck{b}\,]$. This means that, in the absence of bias (e.g., when the true cost functions of all agents are minimized at the same location), the term $\widecheck{x}_{\delta}$ is zero. 

In summary, the steady-state error contains a classical term determined by the gradient noise, plus an additional term arising from data compression. 
In the latter error, backpropagation of the quantization error lets additional components of the gradient noise and the bias seep into the ACTC evolution, determining an increase of the mean-square-deviation.

The qualitative arguments illustrated above will be made rigorous in the appendices, providing technical guarantees for the ACTC strategy to be mean-square stable, with a steady-state mean-square-deviation on the order of $O(\mu)$, as stated in the next theorem. 
In order to state the theorem, it is necessary to introduce a useful function that will be critical to evaluate the mean-square stability of the ACTC strategy.

Following the canonical Jordan decomposition illustrated in Appendix~\ref{app:collection}, we denote by $\lambda_n$ the eigenvalue of $A$ associated with the $n$-th Jordan block of $A$, by $L_n$ the dimension of this block, and by $B$ the number of blocks.
Let
\beq
a_n \triangleq\frac{2 |\lambda_2| }{1-|\lambda_2|}\,\frac{1}{|\lambda_2|-|\lambda_n|^2},
\label{eq:angammacoeff}
\eeq
and let us introduce the function:
\beq
\gamma(A)\triangleq
\sum_{n=2}^B 
\frac{|\lambda_2|}{|\lambda_2|-|\lambda_n|^2}
\,
\left(\frac{a_n^{L_n+1}-1}{(a_n-1)^2}+\frac{L_n+1}{a_n-1}
\right).
\label{eq:gammaAdef0}
\eeq

\begin{theorem}[Mean-Square Stability]
\label{th:MSEstab}
Let
\beq
\zeta<\frac{1}{16 \widecheck{\Delta} \gamma(A)}, 
\label{eq:zetastabound}
\eeq
and let
\beq
\mu<\frac{2}{\zeta\,(\eta + \nu)},\quad \mu<\mu^{\star},
\label{eq:mustabound}
\eeq
where $\mu^{\star}$ is the positive\footnote{Equation \eqref{eq:mu0eq2maintext} is in the form $a \mu^2 + b\mu - c=0$, where $a$, $b$ and $c$ are positive. Thus, the equation admits two real-valued roots of opposite sign.} root of the equation:
\beq
\mu^2
\left(
1+\frac{\sigma^2_{12}}{\nu^2}
\right)
\,
\frac{\gamma(A)}{1-16\,\zeta\,\widecheck{\Delta}\,\gamma(A)}
+\mu\,\zeta\frac{1+16 \bar{\Delta}}{\nu} - \frac{1}{\phi}=0,
\label{eq:mu0eq2maintext}
\eeq
with $\bar{\Delta}$ and $\widecheck{\Delta}$ being defined in Table~\ref{tab:Notation}, and $\phi$ is a positive scalar that embodies the constants appearing in the $\mu^2$-terms of matrices $T_s$, $T_{\delta}$ and $T_q$ in Table~\ref{tab:TransferMaTable}. The evaluation of $\phi$ is rather cumbersome and is detailed in Appendix~\ref{app:proofTheorem1}.
Then the ACTC strategy is mean-square stable, namely, 
\beq
\limsup_{i\rightarrow\infty}
\E\|\bm{w}_{k,i}-w^{\star}\|^2<\infty.
\label{eq:merestability}
\eeq
Moreover, in the small step-size regime the mean-square-deviation is of order $\mu$, namely,
\beq
\boxed{
\limsup_{i\rightarrow\infty}
\E\|\bm{w}_{k,i}-w^{\star}\|^2=O(\mu) \textnormal{ as $\mu\rightarrow 0$}
}
\label{eq:orderofstability}
\eeq
\end{theorem}

\begin{IEEEproof}
See Appendix~\ref{app:proofofTheorem2}.
\end{IEEEproof}

\subsection{Insights from Theorem~\ref{th:MSEstab}}
The stability analysis leading to Theorem~\ref{th:MSEstab} and carried out in Lemmas~\ref{lem:Estab} (Appendix~\ref{subsec:stabE}) and~\ref{lem:Tstab} (Appendix~\ref{subsec:stabT}) is conducted in the transformed (complex) $z$-domain exploiting the formalism of resolvent matrices. This turns out to be a powerful approach that allows us to get necessary and sufficient conditions for stability.

In particular, the condition on $\zeta$ in \eqref{eq:zetastabound} is a necessary and sufficient condition for the stability of matrix $E$, which is in turn related to the speed of decay of the network transient, namely, to how fast the network agents coordinate among themselves to converge to a small neighborhood of the (true) minimizer $w^{\star}$. 
Condition \eqref{eq:zetastabound} reveals the main utility of introducing the parameter $\zeta$ in the ACTC algorithm. Notably, this parameter is not present (i.e., $\zeta=1$) in the unquantized version of the diffusion algorithm, i.e., in the ATC algorithm. However, Eq. \eqref{eq:zetastabound} shows that not necessarily the value $\zeta=1$ grants stability. 

Examining the structure of $\gamma(A)$ in \eqref{eq:gammaAdef0}, we see that the RHS of \eqref{eq:zetastabound} depends on two main elements, namely, $i)$ a constant $\widecheck{\Delta}$ that contains the quantization blow-up factors $\omega_k$; $ii)$ the eigenstructure of $A$, and particularly the second largest magnitude eigenvalue $\lambda_2$. 
Let us examine in detail the role played by each of these elements.

Regarding the quantization constant $\widecheck{\Delta}$, we see that poorer resolutions (i.e., higher values of $\widecheck{\Delta}$) go against stability, an effect that can be compensated by choosing smaller values for $\zeta$. This means that $\zeta$ is useful to compensate for the quantization error that seeps into the recursion of the individual agent's errors.  

Regarding the eigenstructure of $A$, the fundamental role of matrix $E$ for mean-square stability is summarized by the function $\gamma(A)$ in \eqref{eq:gammaAdef0}. This function provides an accurate stability threshold by capturing the full eigenstructure of $A$ through the eigenvalues $\lambda_n$, the size and the number of Jordan blocks.
In this way, we are given the flexibility of providing accurate stability thresholds for different types of combination matrices. 
Let us illustrate these useful features in relation to some relevant cases and existing results. 

A detailed stability analysis of the classical (i.e., uncompressed) ATC diffusion strategy under the general assumptions considered in this work was originally carried out in~\cite{ChenSayedTIT2015part1,Sayed}. Such analysis  relies on a more general Jordan decomposition (with arbitrary $\epsilon$ replacing the ones on the superdiagonal), which is instrumental to get an $\ell_1$ upper bound on the spectral radius, which in turn provides a sufficient condition for mean-square stability. 
One feature of this sufficient condition is that the stability range for $\mu$ scales exponentially as $\epsilon^{-N}$, with $\epsilon<1$, a condition that becomes stringent for large-scale networks. 

The analysis conducted in this work is different, and focuses on the resolvent matrices in the $z$-transformed complex domain. 
Thanks to this approach, we obtain a (necessary and sufficient) condition for stability that accounts for the entire eigenspectrum of $A$ through the function $\gamma(A)$ in \eqref{eq:gammaAdef0}. 
We will now explain how such eigenstructure plays a role for different types of combination matrices, and how the actual conditions for stability are in fact milder than the aforementioned exponential scaling.

{\em Diagonalizable Matrices.} 
For simplicity, let us consider the case that, excluding $\lambda_1=1$, all the remaining eigenvalues are equal, namely, $\lambda_n=\lambda_2$ for $n>1$. Note that this yields $a_n=const.$ in \eqref{eq:angammacoeff}. 
If $A$ is diagonalizable, we have $B=N$ and $L_n=1$ for all $n>1$, which, using \eqref{eq:gammaAdef0}, yields:
\beq
\gamma(A)\propto N,
\eeq 
i.e., $\gamma(A)$ scales linearly with $N$.\footnote{
We remark that, in the case of diagonalizable $A$, the linear scaling is a tight estimate, since a known result about rank-one perturbations of diagonal matrices allows us to evaluate the spectral radius in an exact manner as the sum of the spectral radius of the unperturbed matrix plus $N$ times the size of the perturbation~\cite{bib:diagonalrankone}.}

{\em ``Very'' Non-Diagonalizable Matrices.} 
Consider the opposite case where $B=2$ and $L_n=N-1$, namely, apart from the first Jordan block (i.e., the one associated with the single eigenvalue $\lambda_1=1$), we have only another block associated with $\lambda_n=\lambda_2$. Under this setting, 
we see from \eqref{eq:gammaAdef0} that  
\beq
\gamma(A)\sim \left(\frac{2 |\lambda_2|}{(1-|\lambda_2|)^2}\right)^N,
\label{eq:gammaAbadscaling}
\eeq
which implies an exponential scaling with $N$.

{\em Typical Non-Diagonalizable Matrices.}
The exponential scaling observed in \eqref{eq:gammaAbadscaling} is clearly not desirable for stability, since, in light of \eqref{eq:zetastabound}, it would significantly reduce the stability range. 

However, typical non-diagonalizable matrices adopted in distributed optimization applications seldom feature the extreme eigenstructure described above. 
As a matter of fact, if we perform the Jordan decomposition of typical combination matrices, we see that the size of the Jordan blocks is usually modest, and in any case it does not increase linearly with $N$. 
This means that the exponents in \eqref{eq:gammaAdef0} would be determined by the maximum Jordan block size, and not by the network size. Moreover, taking into account the fact that for typical combination matrices many eigenvalue magnitudes are considerably smaller than the second largest magnitude, the stability thresholds obtained through \eqref{eq:gammaAdef0} are significantly far from exhibiting an exponential scaling with $N$. 

We note also that larger values of $|\lambda_n|$, and particularly of $|\lambda_2|$, go against the stability of the network error matrix $E$, which means that $\zeta$ is useful to regulate the stability when the network component evolves more slowly, i.e., when $|\lambda_2|$ is closer to $1$.

In summary, the network error convergence depends upon the eigenstructure of $A$ and the quantizer's resolution. 
While the role of the eigenstructure of $A$ (and, hence, of the network connectivity) is common to the standard ATC strategy, one distinguishing feature of the ACTC strategy is represented by the fact that the spectral radius of $E$ increases due to the presence of the additional factor $\widecheck{\Delta}$. This means that the agreement among agents slows down due to backpropagation of the quantization error. 
However, and remarkably, this slowdown does not preclude the possibility of accurate performance given sufficient time for learning. 
Paralleling classical coding results from Shannon's theory, we could say that {\em the price of compression is not the impossibility of learning, rather a slowdown in the convergence}.

\section{Transient Analysis}
The next result refines the mean-square stability result in Theorem~\ref{th:MSEstab} to charactetize the learning dynamics of the ACTC strategy. In particular, we will characterize the transient phases of the algorithm before it converges to the steady state.
\begin{theorem}[ACTC Learning Behavior]
\label{th:transient}
Let
\beq
\rho_{\rm{cen}}\dfz (1 - \mu\,\zeta\,\nu)^2,
\eeq
where $\nu$ is the global strong convexity constant in \eqref{sqNablaStrConvex}. 
Assume that
\beq
\zeta<\frac{1}{16 \widecheck{\Delta} \gamma(A)}, 
\label{eq:zetastaboundtheorem3}
\eeq
let $\rho(E)$ be the spectral radius of $E$ (which under assumption \eqref{eq:zetastaboundtheorem3} has been proved to be smaller than $1$), and set $\epsilon>0$ such that:
\beq
\rho_{\rm{net}}=\rho(E)+\epsilon.
\eeq
Using the definitions of the compression factor $\Omega$ in \eqref{eq:omegamax}, of the entries $\{\pi_{\ell}\}$ of the Perron eigenvector in \eqref{eq:Perronvecfirstdef}, of the scaled step-sizes $\{\alpha_{\ell}\}$ in \eqref{eq:alphascalestepdef0}, and of the gradient-noise variances $\{\sigma^2_{\ell}\}$ in \eqref{eq:gradNoiseBound}, in the small-$\mu$ regime the evolution of the mean-square-deviation of the individual agent $k$ can be cast in the form, for all $i>0$:
\beq
\boxed{
\begin{array}{lll}
\E\|\bm{w}_{k,i} - w^{\star}\|^2
\leq
O(1) \,\rho_{\rm{net}}^i + O(1)\,\rho_{\rm{cen}}^i  + O(\mu)\,\rho_{\rm{cen}}^{i/4}
\\
\\
\mu\,\zeta\left(
\displaystyle{\frac{\sum_{\ell=1}^N \pi_{\ell} \,\alpha_{\ell}^2 \sigma_{\ell}^2}{2\nu}} 
+c_q\,\Omega\,(1+\Omega)
\right) +  O(\mu^{3/2})
\end{array}
}
\nonumber
\eeq
\beq
\phantom{x}
\label{eq:mainmainres}
\eeq
where the Big-O terms depend in general on the particular agent $k$, except for the $O(1)$ term multiplying $\rho_{\rm{cen}}^i$, and $c_q$ is a positive constant independent of $\mu$ and $i$.
\end{theorem}
\begin{IEEEproof}
See Appendix~\ref{app:Theorem3}.
\end{IEEEproof}

Thanks to \eqref{eq:mainmainres}, we arrive at a sharp and revealing description of the learning behavior of compressed diffusion strategies over adaptive networks. 
In fact, the individual terms arising in \eqref{eq:mainmainres} admit the following physical interpretation:
\begin{align}
&\underbrace{O(1)\,\rho_{\rm{net}}^i }_{\substack{\textnormal{network convergence {\bf to}}\\ \textnormal{the centralized solution}}}
+
\underbrace{O(1)\,\rho_{\rm{cen}}^i}_{\substack{\textnormal{convergence {\bf of} the} \\ \textnormal{centralized solution}}} 
+
\underbrace{O(\mu)\,\rho_{\rm{cen}}^{i/4} }_{\substack{\textnormal{higher-order correction} \\ \textnormal{relative to the transient phase}}}
\nonumber\\
\nonumber\\
&+ 
\overbrace{
\mu\,\zeta\left(
\underbrace{
\displaystyle{\frac{\sum_{\ell=1}^N \pi_{\ell} \,\alpha_{\ell}^2 \sigma_{\ell}^2}{2\nu}}
}_{\textnormal{uncompressed ACTC}}
+
\underbrace{
c_q\,\Omega\,(1+\Omega)
}_{\textnormal{compression loss}}
\right)
}^{\textnormal{steady-state error}}
+O(\mu^{3/2}),
\label{eq:mainmainres2}
\end{align}
which allows us to examine closely the distinct learning phases as detailed in the following remarks.

{\em --- Transient Phases}. 
First, we notice that the network rate $\rho_{\rm{net}}$ depends only on the stability parameter $\zeta$, and on the network connectivity properties through the eigenspectrum of the combination matrix $A$. 
As a result, we see that for sufficiently small $\mu$ we have that $\rho_{\rm{cen}}>\rho_{\rm{net}}$. 
Accordingly, for small step-sizes $\mu$, the transient associated with the convergence of the network solution toward the centralized solution dies out earlier (Phase I). After this initial transient, a second transient dominates (Phase II), which is relative to the slower (since $\rho_{\rm{cen}}>\rho_{\rm{net}}$) process that characterizes the convergence of the centralized solution to the steady-state. 
Remarkably, these two distinct phases of adaptive diffusion learning have already been identified in the context of adaptive learning over networks {\em without communication constraints}~\cite{ChenSayedTIT2015part1,ChenSayedTIT2015part2}.

{\em --- Compression Loss}.
After transient Phase II, the ACTC behavior is summarized in the following {\em upper bound} on the steady-state mean-square-deviation:
\begin{align}
{\sf MSD}_{\rm{ACTC}}
&=
\mu\,\zeta\left(
\displaystyle{\frac{\sum_{\ell=1}^N \pi_{\ell} \,\alpha_{\ell}^2 \sigma_{\ell}^2}{2\nu}}
+
c_q\,\Omega\,(1+\Omega)
\right)\nonumber\\
&+ O(\mu^{3/2}).
\label{eq:MSDACTC}
\end{align}
First of all, we remark that the product $\mu\,\zeta$ stays fixed once we set a prescribed convergence rate $\rho_{\rm{cen}}$. 
Then, for a given convergence rate, we see that the mean-square-deviation is composed of two main terms. 
The first term does not depend on the amount of compression, and is proportional to an average over the Perron weights $\{\pi_\ell\}$ of the scaled gradient noise powers $\{\alpha_{\ell}^2\sigma^2_{\ell}\}$, further divided by the global strong convexity constant $\nu$. 
The second term on the RHS of \eqref{eq:MSDACTC} is the {\em compression loss}, which is in fact an increasing function of the compression factor $\Omega$. 
The limiting case $\Omega=0$ corresponds to the setting {\em without compression}, i.e., to the ACTC algorithm in \eqref{ACTC} where the compression operator is the identity operator, formally:
\beq
{\sf MSD}_{\rm{ACTC}} - {\sf MSD}_{\rm{unc.\,ACTC}}=
\mu\,\zeta \,c_q\,\Omega\,(1+\Omega)
 + O(\mu^{3/2}),
\label{eq:MSDiffbituncACTC00}
\eeq
where we referred to the strategy with $\Omega=0$ as to the {\em uncompressed} ACTC. 

In particular, if we choose as compression operators the randomized quantizers examined in Sec.~\ref{sec:Alistarh}, we can obtain an explicit connection between the mean-square-deviation and the bit-rate. 
In fact, examining \eqref{eq:omegainsight}, we see that either:
\beq
\frac{M}{L^2}\leq 1\Rightarrow \Omega=\frac{M}{L^2}\Rightarrow \Omega + \Omega^2\leq 2\,\Omega=\frac{2 M}{L^2}\\
\eeq
or
\beq
\frac{M}{L^2}>1\Rightarrow \Omega=\frac{\sqrt{M}}{L}\Rightarrow \Omega + \Omega^2\leq 2\,\Omega^2=\frac{2 M}{L^2}.
\eeq
In summary, in both cases we can write:
\beq
\Omega\,(1+\Omega)\leq \frac{2 M}{L^2}
\leq
\frac{2 M}{(2^{r_{\min}} - 1)^2},
\label{eq:bitratebound}
\eeq
where we denoted by $r_{\min}$ the minimum bit-rate across agents.
From \eqref{eq:MSDiffbituncACTC00} and \eqref{eq:bitratebound} we conclude that the increase in mean-square-deviation from the uncompressed to the standard ACTC is given by:
\beq
{\sf MSD}_{\rm{ACTC}} - {\sf MSD}_{\rm{unc.\,ACTC}}\leq
\mu\,\zeta  \frac{c_q M}{(2^{r_{\min}} - 1)^{2}}
\!+\! O(\mu^{3/2}),
\label{eq:MSDiffbituncACTC}
\eeq
which reveals the following remarkable analogy with the fundamental laws of high-resolution quantization: for small step-sizes, the {\em excess mean-square-deviation} due to quantization scales exponentially with the bit-rate as $2^{-2 r_{\min}}$~\cite{GershoGrayBook}.

{\em --- Comparison Against Classical ATC}.
From the ACTC algorithm in \eqref{ACTC}, it is readily seen that the {\em uncompressed} ACTC (i.e., the ACTC with compression operator equal to the identity operator) coincides with the classical ATC only for the case $\zeta=1$, yielding:\footnote{Actually, in \eqref{eq:summaryunc} ${\sf MSD}^{\rm{ATC}}$ is an {\em upper bound} on the exact mean-square-deviation. This upper bound would be obtained by applying the mean-square stability analysis carried out in~\cite{ChenSayedTIT2015part1}. An exact evaluation (and not a bound) for the mean-square-deviation is instead provided in~\cite{ChenSayedTIT2015part2}. However, since the present work focuses on the mean-square stability analysis for the ACTC scheme and not on a steady-state analysis, the appropriate term of comparison is \eqref{eq:summaryunc}, and not the refined value that would be obtained from~\cite{ChenSayedTIT2015part2}.}  
\begin{numcases}{}
{\sf MSD}_{\rm{unc.\,ACTC}}=\mu\,\zeta \,\displaystyle{\frac{\sum_{\ell=1}^N \pi_{\ell} \,\alpha_{\ell}^2 \sigma_{\ell}^2}{2\nu}} + O(\mu^{3/2})
\nonumber\\
\label{eq:summaryunc}\\
{\sf MSD}_{\rm{ATC}}=\mu \,\displaystyle{\frac{\sum_{\ell=1}^N \pi_{\ell} \,\alpha_{\ell}^2 \sigma_{\ell}^2}{2\nu}} + O(\mu^{3/2})
\nonumber
\end{numcases}
However, to compare fairly the two strategies, we need to set the same desired value for the rate. 
Let ${\sf R}$ be the desired rate, we have the following relationships: 
\begin{numcases}{\!\!\!\!\!\!}
{\sf R} \approx 1-2\mu\,\zeta\,\nu\Rightarrow \mu\,\zeta\approx\frac{1-{\sf R}}{2\nu}
~~~\textnormal{[unc. ACTC]}
\nonumber\\
\label{eq:ACTCATCrateformulas}\\
{\sf R} \approx 1-2\mu\,\nu\Rightarrow \mu\approx\frac{1-{\sf R}}{2\nu}~~~~~~~\textnormal{[ATC]}\nonumber
\end{numcases}
which, substituted in \eqref{eq:summaryunc} yields:
\beq
{\sf MSD}_{\rm{unc.\,ACTC}} - {\sf MSD}_{\rm{ATC}}=O(\mu^{3/2}),
\label{eq:MSDiffACTCuncATC}
\eeq
where we remark that the higher-order terms $O(\mu^{3/2})$ do not cancel out since, in general, the $O(\mu^{3/2})$ corrections are different for the uncompressed ACTC and the ATC strategies. 
In particular, the stabilization parameter $\zeta$ typically entails a slight increase in the mean-square-deviation that is incorporated in the $O(\mu^{3/2})$ correction.\footnote{These types of higher-order differences among similar forms of distributed implementations are commonly encountered in the pertinent literature, for example, when one compares the ATC strategy against the Combine-Then-Adapt strategy or against the consensus strategy~\cite{Sayed}.}  

Equation \eqref{eq:MSDiffACTCuncATC} highlights that, in the small-$\mu$ regime, the uncompressed ACTC and classical ATC are equivalent at the leading order in $\mu$. 
Therefore, joining \eqref{eq:MSDiffbituncACTC00}, \eqref{eq:ACTCATCrateformulas}, and \eqref{eq:MSDiffACTCuncATC}, we conclude that:
\beq
{\sf MSD}_{\rm{ACTC}} - {\sf MSD}_{\rm{ATC}}=
\frac{1-{\sf R}}{2\nu} c_q\,\Omega\,(1+\Omega)
 + O(\mu^{3/2}).
\label{eq:MSDiffbituncACTC11}
\eeq
In particular, for the randomized quantizers in Sec.~\ref{sec:Alistarh}, from \eqref{eq:MSDiffbituncACTC} we get:
\beq
\boxed{
{\sf MSD}_{\rm{ACTC}} - {\sf MSD}_{\rm{ATC}}\leq
\frac{1-{\sf R}}{2\nu}\frac{c_q M}{(2^{r_{\min}} - 1)^{2}}
 + O(\mu^{3/2})
}
\nonumber
\eeq
\beq
\phantom{x}
\label{eq:MSDiffbitACTCATC}
\eeq
namely, for sufficiently small $\mu$ the excess of mean-square-deviation of the ACTC strategy w.r.t. to the classical ATC strategy scales as $\approx 2^{-2 r_{\min}}$.

\begin{figure*}[t]
\centering
\[
\begin{array}{cc}
\includegraphics[width=0.5\linewidth]{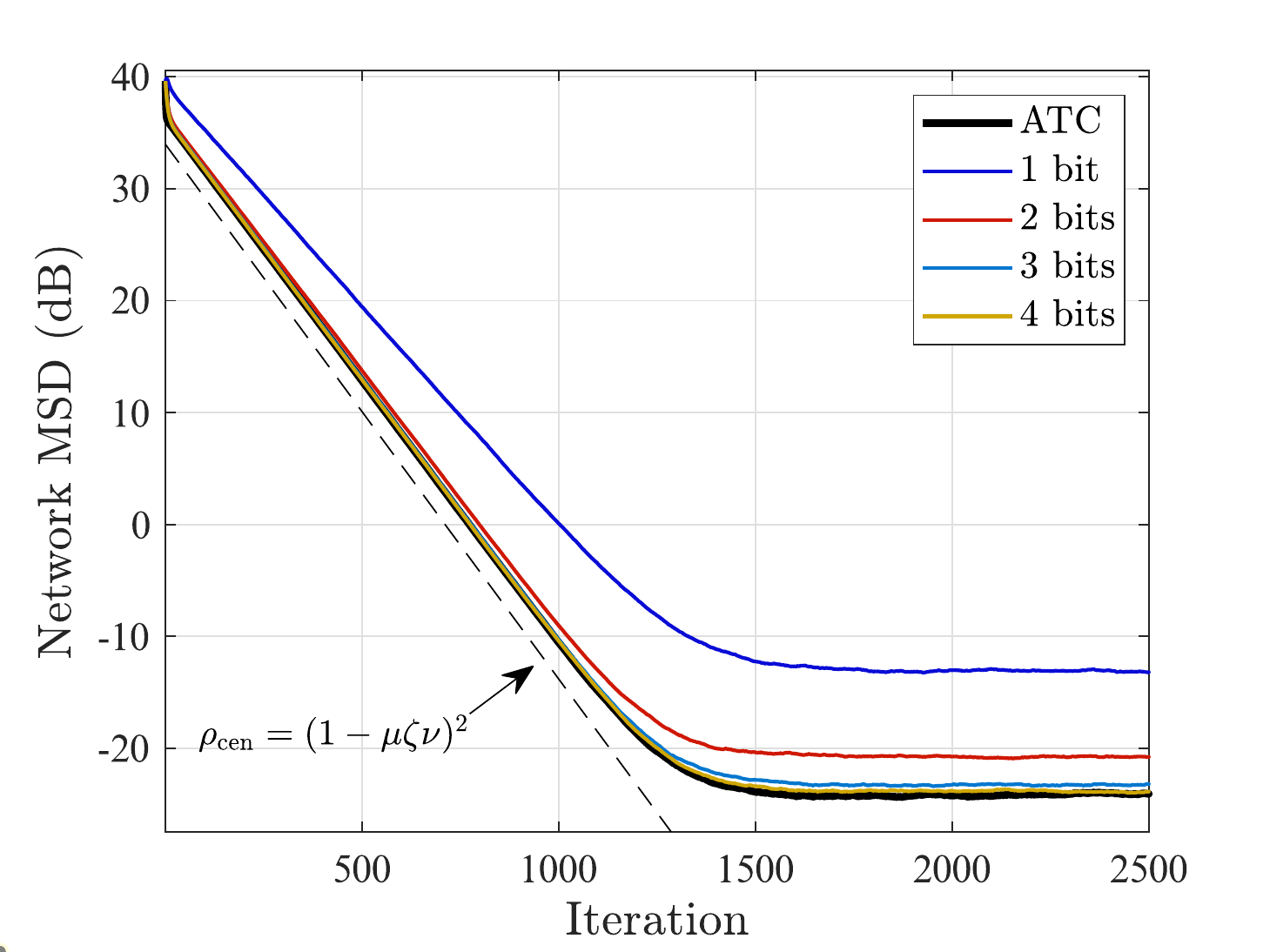}
\includegraphics[width=0.5\linewidth]{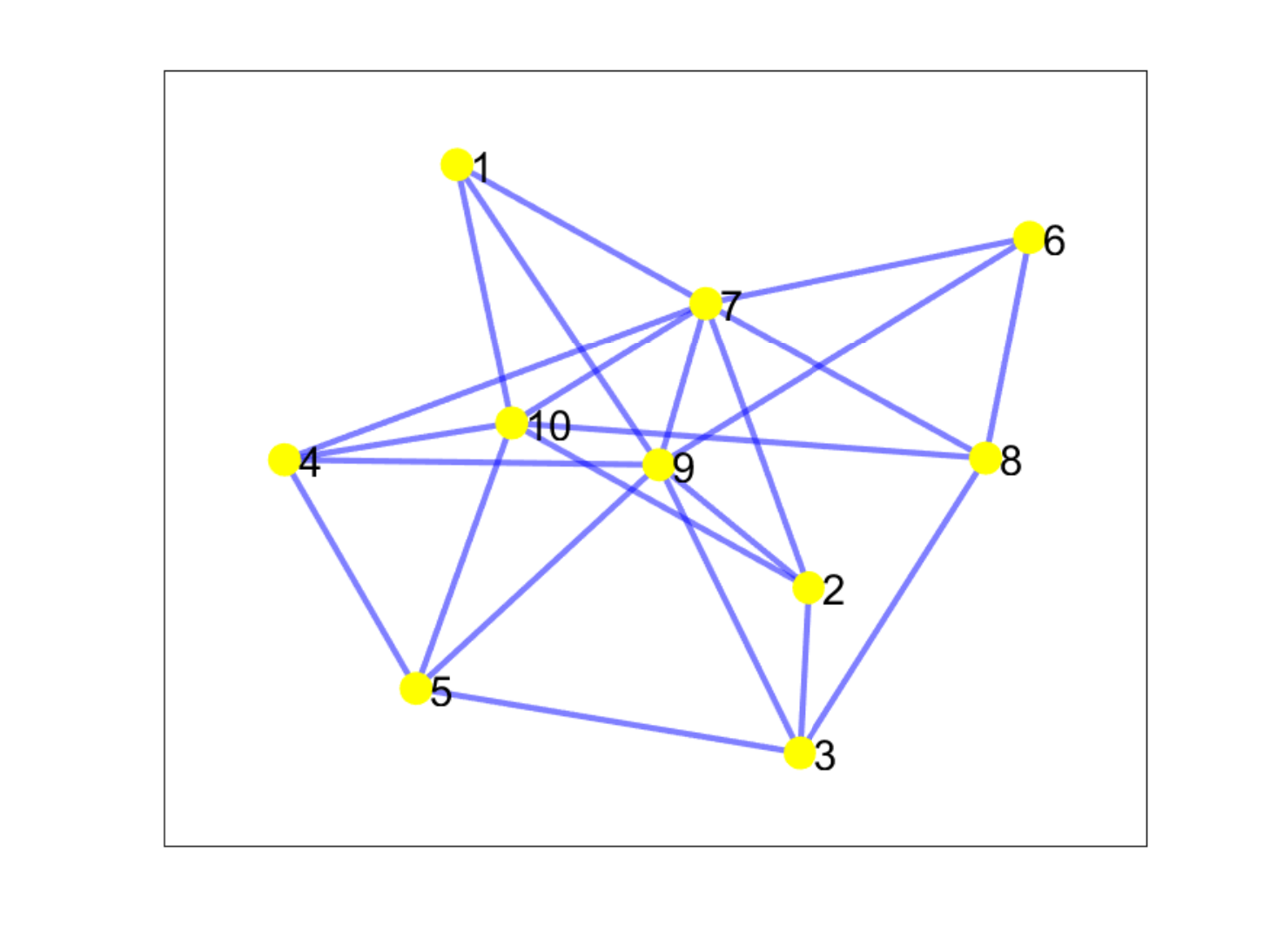}
\end{array}
\]
\caption{{\em Left plot}. 
ACTC network mean-square-deviation in \eqref{eq:netMSD} as a function of the iteration $i$, for different values of the bit-rate $r$. We considered the distributed regression problem in Sec.~\ref{sec:experiments}, with dimensionality $M=50$, Gaussian regressors $\bm{u}_{k,i}$ with diagonal matrices and variances drawn as independent realizations from a uniform distribution in $(1,4)$, and Gaussian disturbances $\bm{v}_{k,i}$ with variances drawn as independent realizations from a uniform distribution in $(0.25,1)$. 
The ACTC algorithm is run with stability parameter $\zeta = 0.25$, and with equal step-sizes $\mu_k=\mu=4 \times10^{-3}$. 
All agents use the randomized quantizer in Sec.~\ref{sec:AlistarhQuant} with bit-rate $r$. 
The mean-square-deviations are estimated by means of $10^2$ Monte Carlo runs.
{\em Right plot}. Network topology used in the experiments, on top of which we build a Metropolis combination matrix to run the ACTC algorithm. All nodes have a self-loop, not shown in the figure.
}
\label{fig:1}
\end{figure*}

\section{Illustrative Examples}
\label{sec:experiments}

As an application of the ACTC diffusion strategy, we consider the scenario where $N$ agents, interconnected through a network satisfying Assumptions \ref{Strong Connectivity} and \ref{Stochastic combination matrix}, aim at solving a regression problem in a distributed way. Each individual agent observes a flow of streaming information. Specifically, at each time $i$, agent $k$ observes data $\bm{d}_{k,i}\in\mathbb{R}$ and regressors $\bm{u}_{k,i}\in\mathbb{R}^M$, which obey the following linear regression model:
\beq
\bm{d}_{k,i} = \bm{u}_{k,i}^{\top}w^{\star} + \bm{v}_{k,i} \quad k=1,\ldots,N,
\label{eq:linregmain}
\eeq
where $w^{\star}\in\mathbb{R}^{M}$ is an unknown (deterministic) parameter vector and $\bm{v}_{k,i}\in\mathbb{R}$ acts as noise. 
We assume that processes $\{\bm{u}_{k,i}\}$ and $\{\bm{v}_{k,i}\}$ have mean equal to zero, are independent both over time and space (i.e., across the agents), with second-order statistics given by, respectively:
\beq
R_{u,k}=\E[\bm{u}_{k,i} \bm{u}_{k,i}^{\top}],\qquad\E \bm{v}^2_{k,i}=\sigma^2_{v,k}.
\eeq
The goal is to to estimate the unknown $w^{\star}$, which, by applying straightforward manipulations to \eqref{eq:linregmain}, can be seen to obey the relationship:
\beq
r_{du,k}=R_{u,k} w^{\star},
\label{eq:rduRu}
\eeq
where $r_{du,k}=\E[\bm{d}_{k,i} \bm{u}_{k,i}]$. 
In principle, each agent could perform estimation of $w^{\star}$ by solving the optimization problem:
\beq
\min_{w\in\mathbb{R}^M}\E\left(\bm{d}_{k,i}-\bm{u}_{k,i}^{\top} w\right)^2,
\label{eq:individualMSE}
\eeq
which in turn corresponds to adopting the following quadratic loss and cost functions:
\begin{align}
&L_k(w;\{\bm{d}_{k,i}, \bm{u}_{k,i}\}) = 
(\bm{d}_{k,i} - \bm{u}_{k,i}^{\top} w)^2,\\
&J_k(w) = \E \left[ L_k(w;\{\bm{d}_{k,i}, \bm{u}_{k,i}\})\right].
\end{align}
There are several reasons why the agents can be interested in solving the regression problem in a cooperative fashion. 
First of all, it was shown in~\cite{Sayed} that, under suitable design, cooperation is beneficial in terms of inference performance. 
Even more remarkably, in many cases the local regression problem \eqref{eq:individualMSE} can be ill-posed if the agents' regressors do not contain sufficient information. This is an issue classically known as {\em collinearity}, which basically implies that the regression covariance matrix $R_{u,k}$ is singular, and many $w$ exist that solve \eqref{eq:individualMSE}. 
This behavior can be easily grasped by noticing that
\beq
\nabla J_k(w)=2 (R_{u,k} w - r_{du,k}),
\eeq 
and, examining \eqref{eq:rduRu}, we see that if $R_{u,k}$ is not invertible, $w^{\star}$ is a solution to the optimization problem, but not the only one.
Accordingly, reliable inference about $w$ is impaired by collinearity. Technically, when the regression covariance matrix of agent $k$ is singular, the cost function $J_k(w)$ is {\em not} strongly convex, and the true $w^{\star}$ is {\em one among} the minimizers of \eqref{eq:individualMSE}. 

However, if we now replace \eqref{eq:individualMSE} with its global counterpart:
\beq
\min_{w\in\mathbb{R}^M}\E\left[
\sum_{k=1}^N p_k\left(\bm{d}_{k,i}-\bm{u}_{k,i}^{\top} w\right)^2\right],
\label{eq:coopMSE}
\eeq
it is readily seen that a single agent with a non-singular $R_{u,k}$ (i.e., with a strongly-convex cost function) is able to enable successful inference! 
Notably, such a minimal requirement is sufficient to our ACTC strategy to solve the regression problem in a {\em distributed} way and under {\em communication constraints}.

\subsection{Role of Compression Degree}
In Fig.~\ref{fig:1}, we examine the learning performance of the ACTC strategy as a function of the iteration $i$, for different values of quantization bits. 
The simulations were run under the following setting. 
The regression problem has dimensionality $M=50$. 
The covariance matrices $R_{u,k}$ are all diagonal, and the associated regressors' variances are drawn as independent realizations from a uniform distribution with range $(1,4)$. 
The noise variances $\sigma^{2}_{v,k}$ are drawn as independent realizations from a uniform distribution with range $(0.25,1)$. 
The network is made of $N=10$ agents, connected through the topology displayed in Fig.~\ref{fig:1}, and with Metropolis combination policy. 
Under this setting, we run the ACTC algorithm with stability parameter $\zeta = 0.25$, and with equal step-sizes $\mu_k=\mu=4 \times10^{-3}$. Likewise, the common number of bits employed by the agents will be denoted by $r$, and we will examine the ACTC behavior for different values of $r$, namely, ranging from $1$ to $4$ bits. The compression operator is the randomized quantizer described in Sec.~\ref{sec:AlistarhQuant}. 
All errors are estimated by means of $10^2$ Monte Carlo runs.

\begin{figure}[t]
\centering
\includegraphics[width=0.6\linewidth]{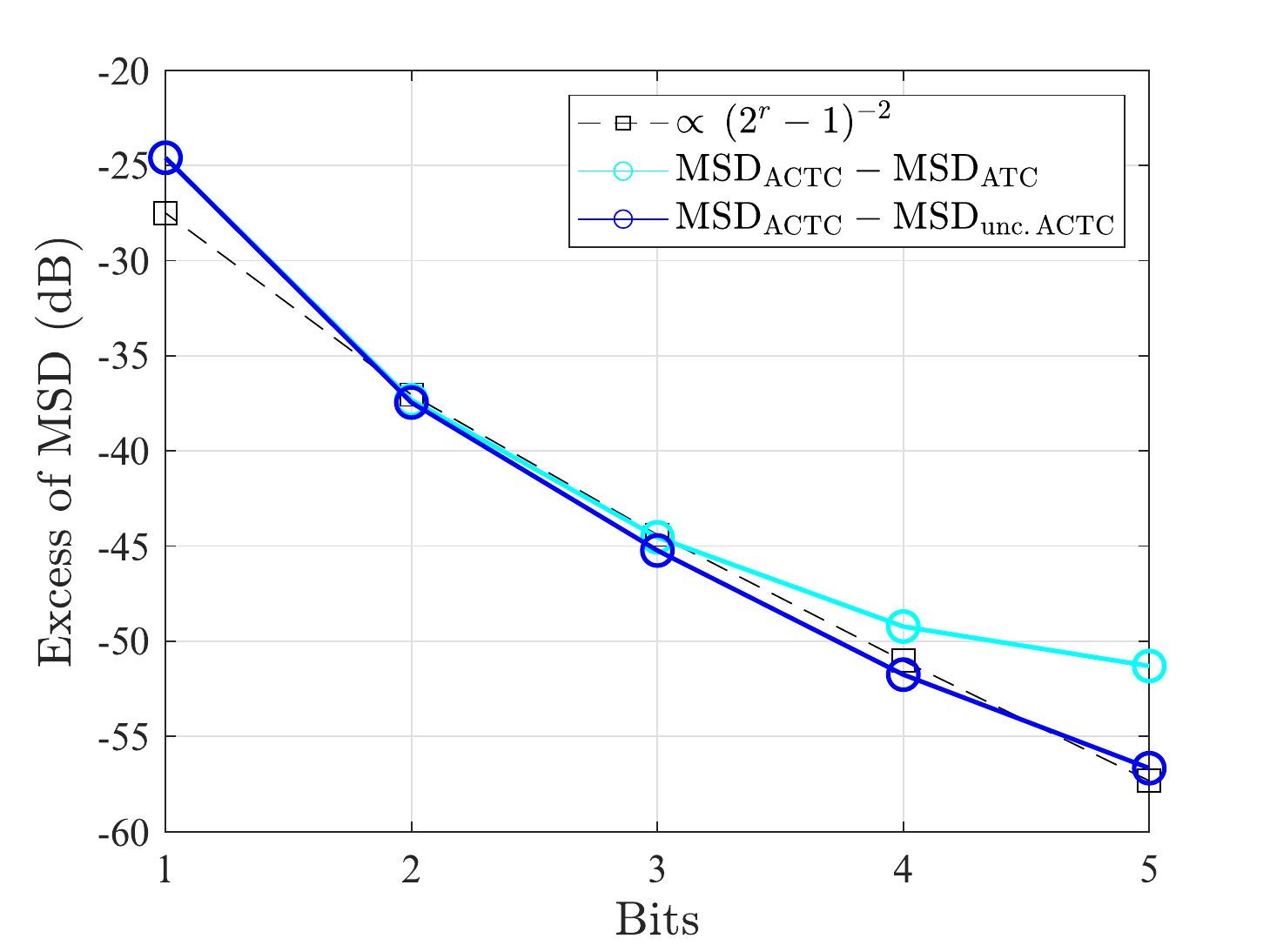}
\caption{Difference between the network mean-square-deviation of the ACTC strategy and the mean-square-deviation of the uncompressed strategies. The label ``uncompressed ACTC'' stems for the ACTC strategy in \eqref{ACTC} with $\bm{\mathcal{Q}}(x)=x$, whereas ``ATC'' stems for the classical ATC in~\cite{Sayed}. 
The dashed curve is obtained by depicting a curve proportional to $(2^r - 1)^{-2}$, with proportionality constant set so as to match the second point of the uncompressed ACTC curve.
The relevant parameters of the ACTC strategy and of the distributed regression problem are set as in Fig.~\ref{fig:1}, but for the dimensionality $M=10$.}
    \label{fig:2}
\end{figure}

As a first performance index, we examine the {\em network} ACTC learning performance, i.e., the mean-square-deviation averaged over all agents:
\beq
\frac1 N \E\|\widetilde{\bm{w}}_{i}-w^\star\|^2.
\label{eq:netMSD}
\eeq 
The behavior observed in Fig.~\ref{fig:1} summarizes sharply the essential characteristics of the ACTC algorithm, as captured by Theorem~\ref{th:transient}: $i)$ for all quantizer's resolutions, the mean-square-deviation has a transient that is essentially governed by the predicted rate $\rho_{\rm{cen}}=(1-\mu\,\zeta\,\nu)^2$ (dashed line); $ii)$ some higher-order discrepancies are absorbed in an initial (much faster) transient; $iii)$ the ACTC errors corresponding to different bits  converge to different steady-state error values that, yet for relatively low bit-rates, approach the performance of the ATC (i.e., unquantized) strategy. 

With reference to the same setting of Fig.~\ref{fig:1}, but for a smaller dimensionality $M=10$, in Fig.~\ref{fig:2} we display the {\em excess of mean-square-deviation} of the ACTC strategy w.r.t. the uncompressed ACTC, see  \eqref{eq:MSDiffbituncACTC00}, and w.r.t. classical ATC, see \eqref{eq:MSDiffbituncACTC11}. 
We remark that the curves in Fig.~\ref{fig:1} {\em do not represent mean-square-deviations}, but the difference between the mean-square-deviation attained by the ACTC diffusion strategy, and the mean-square-deviation that would be attainable in the absence of data compression by the uncompressed ACTC and the classical ATC. Accordingly, such excess of error summarizes only the effect of data compression, and is therefore expected to reduce as the bit-rate increases (while the overall mean-square-deviation cannot vanish since we are in a stochastic-gradient environment).   
We see that the curves scale with the number of bits as $(2^{r} - 1)^{-2}$. This result is in perfect accordance with the predictions of Theorem~\ref{th:transient} --- see \eqref{eq:MSDiffbituncACTC} and \eqref{eq:MSDiffbitACTCATC}. 
As we discussed when commenting on \eqref{eq:MSDiffbituncACTC11}, the uncompressed ACTC strategy features a slight increase in mean-square-deviation, contained in the higher-order $O(\mu^{3/2})$ correction. 
Such small increase becomes visible only when comparable to the quantization error, i.e., when the quantization error becomes negligible. In the particular example of Fig.~\ref{fig:2}, this happens when the quantization error is $\approx -55 \rm{dB}$.

\begin{figure}[t]
    \centering
    \includegraphics[width=0.6\linewidth]{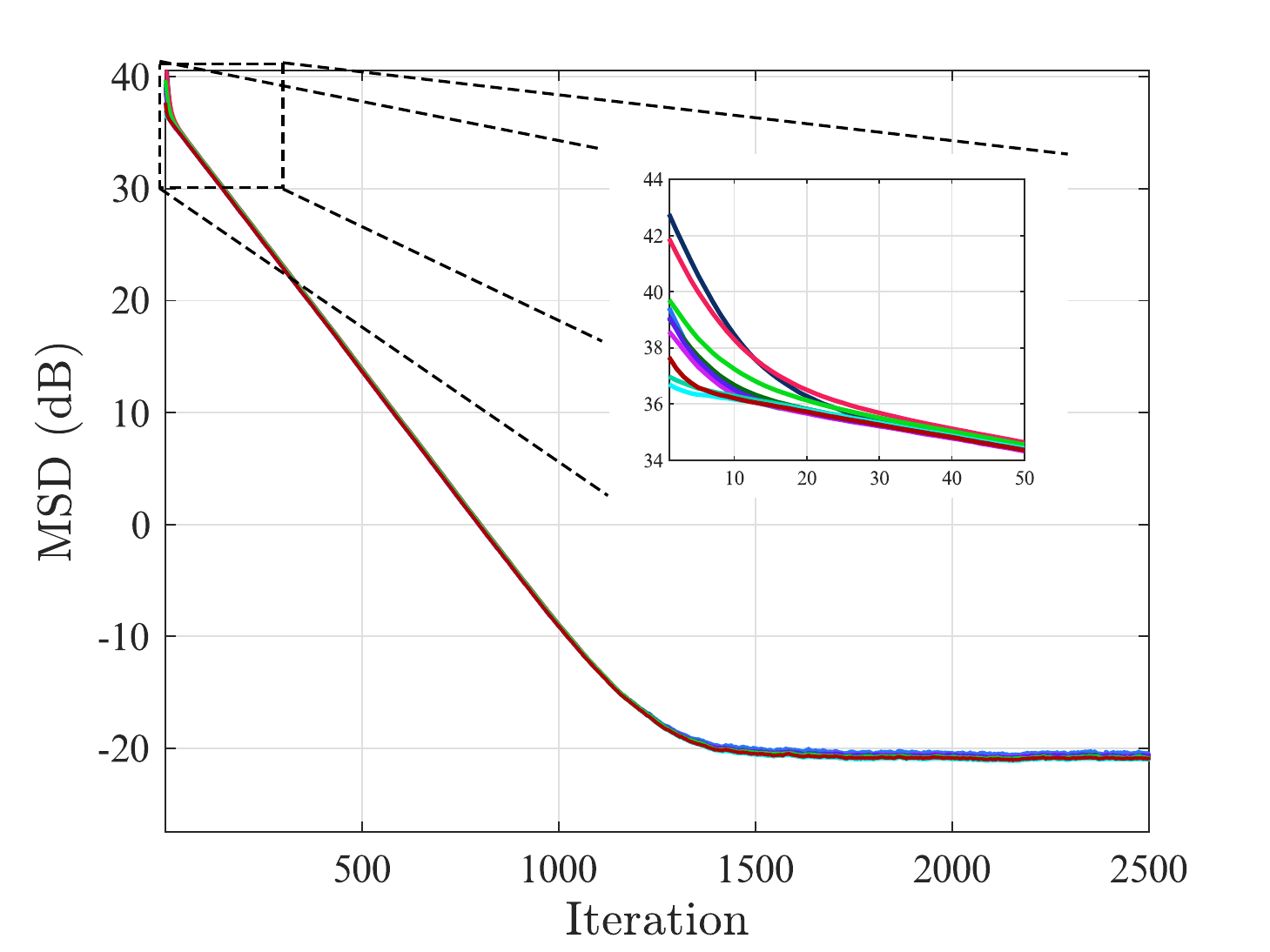}
    \caption{ACTC mean-square-deviation of the {\em individual} agents achieved using $r=2$ bits, under the same setting of Fig.~\ref{fig:1}. The inset plot zooms in on the faster initial transient needed by the agents to reach a coordinated evolution.}
    \label{fig:3}
\end{figure}

In Fig.~\ref{fig:3}, we continue by examining the performance of the {\em individual} agents. 
In accordance with our results, {\em all agents behave similarly}, both in terms of transient and steady-state behavior. 
As shown by \eqref{eq:mainmainres}, initial discrepancies between the agents (see the inset plot) are absorbed into a faster network transient, after which all agents act in a coordinated manner, and converge to the steady-state value.
 
\begin{figure}[t]
    \centering
    \includegraphics[width=0.6\linewidth]{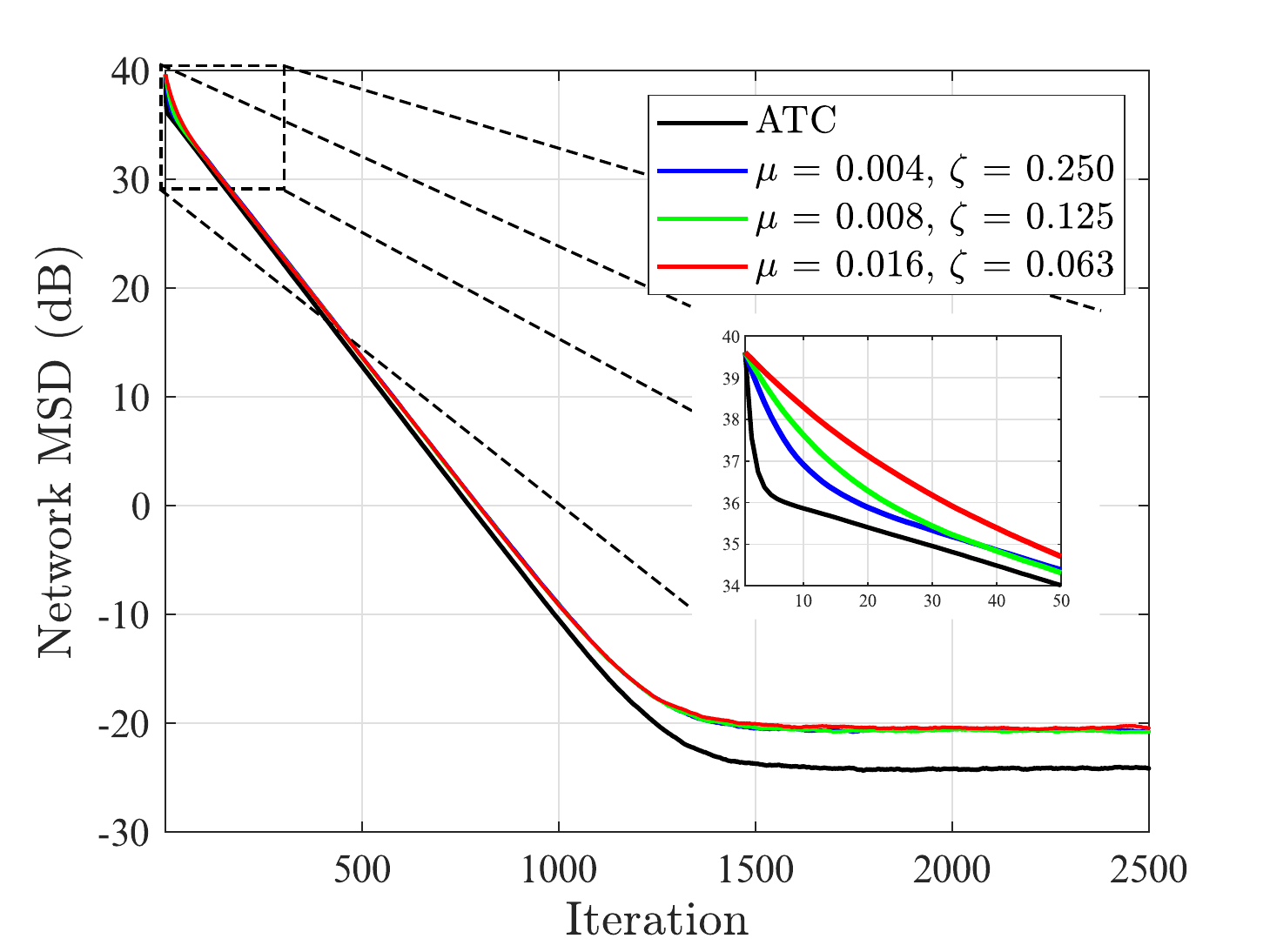}
    \caption{Network mean-square-deviation of the ACTC strategy, for different $(\mu,\zeta)$ pairs guaranteeing the same value of the product $\mu\,\zeta$ (and, hence, the same value of convergence rate $\rho_{\rm{cen}}$). The simulation setting is the same as in Fig.~\ref{fig:1}. The inset plot highlights the impact of $\zeta$ on the faster initial transient needed by the agents to reach a coordinated evolution.}
    \label{fig:4}
\end{figure}

Finally, in Fig.~\ref{fig:4} we examine the joint role of the step-size $\mu$ and of the stabilizing parameter $\zeta$. 
Again, the theoretical predictions are confirmed, since we see that by keeping the product $\mu\,\zeta$ constant, all curves behave equally in terms of rate $(1-\mu\,\zeta\,\nu)^2$ and steady-state error, with $\zeta$ playing some role only in the faster initial network transient (see inset plot). 

It is useful to evaluate the saving, in terms of bits, achieved with the ACTC strategy. 
To this end, we must recall that the randomized quantizers in Sec.~\ref{sec:AlistarhQuant} compress finely (say, with machine precision $32$ bits) the norm of the vectors to be quantized, send one additional bit for the sign of each entry, and then apply random quantization with $r$ bits to each entry --- see \eqref{eq:overallbitbudget}. 
Accordingly, given a dimensionality $M$, a number of iterations $i_{\max}$, and a number of quantization bits $r$, the overall bit expense of each agent is:
\beq
r_{\rm{tot}}=\left(32+M\times (r+1)\right) i_{\max}.
\eeq
Applying this formula to the setting in Fig.~\ref{fig:1}, we see that, for the time necessary to enter reliably the steady state ($i_{\max}\approx 2500$), and referring to the coarser scheme that uses only $2$ bits, we get:
\beq
r_{\rm{ACTC}}^{\textnormal{(tot)}}=\left(32+50\times 2\right) \times 2500=455 \textnormal{ kbit}.
\eeq
This value should be compared against the expense required by the plain ATC strategy, where each entry of the vector to be quantized is represented by $32$ bits, yielding:
\beq
r_{\rm{ATC}}^{\textnormal{(tot)}}=32\times 50\times 2500=4 \textnormal{ Mbit},
\eeq 
implying a remarkable gain of about one order of magnitude in terms of bit rate. 
This gain should be evaluated in relation to the loss, in terms of mean-square-deviation, arising from data compression. 
Inspecting Fig.~\ref{fig:1}, we see that we loose $\approx 4$ dB, which is definitely tolerable, especially in the light of the remarkable bit-rate savings.

\begin{figure}[t]
    \centering
    \includegraphics[width=0.6\linewidth]{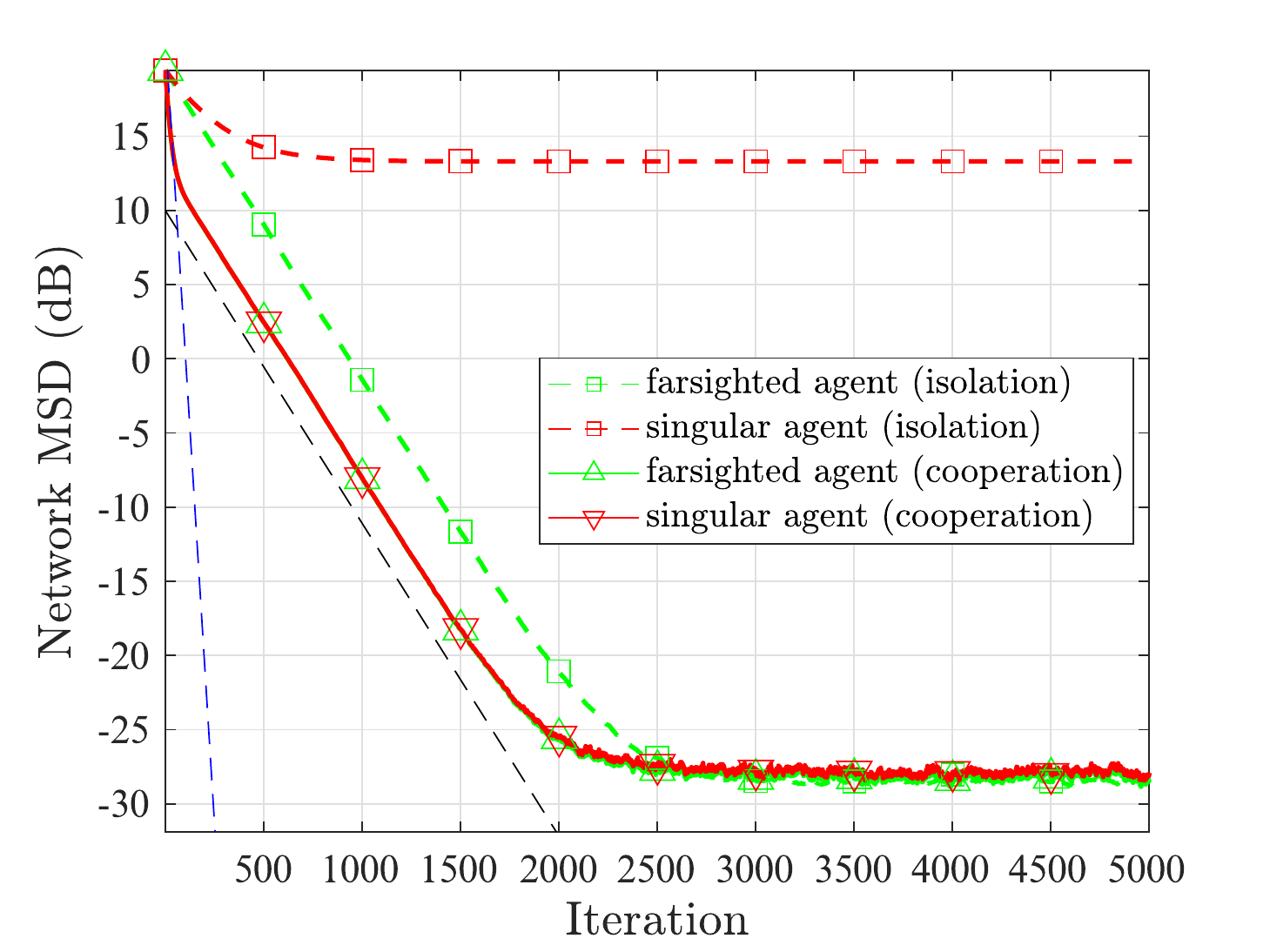}
    \caption{
ACTC network mean-square-deviation in \eqref{eq:netMSD} as a function of the iteration $i$, for the case of a {\em doubly-stochastic} combination matrix obtained with the Metropolis rule. 
We considered the distributed regression problem in Sec.~\ref{subsec:unident} with $N-1$ singular agents having locally unidentifiable regression problems, and one farsighted agent having a locally well-posed regression problem. 
The dimensionality $M=10$, and the parameters of the Gaussian regressors $\bm{u}_{k,i}$ and of the Gaussian disturbances $\bm{v}_{k,i}$ are drawn as in Fig.~\ref{fig:1}, but for the fact that the regressors' matrices of the singular agents have two equal columns (which yields local unidentifiability at these agents). 
The ACTC algorithm is run with stability parameter $\zeta = 0.8$, and with equal step-sizes $\mu_k=\mu=5 \times10^{-3}$. 
All agents use the randomized quantizer in Sec.~\ref{sec:AlistarhQuant} with bit-rate $r=3$. 
All errors are estimated by means of $10^2$ Monte Carlo runs.}
    \label{fig:5}
\end{figure}

\begin{figure}[t]
    \centering
    \includegraphics[width=0.6\linewidth]{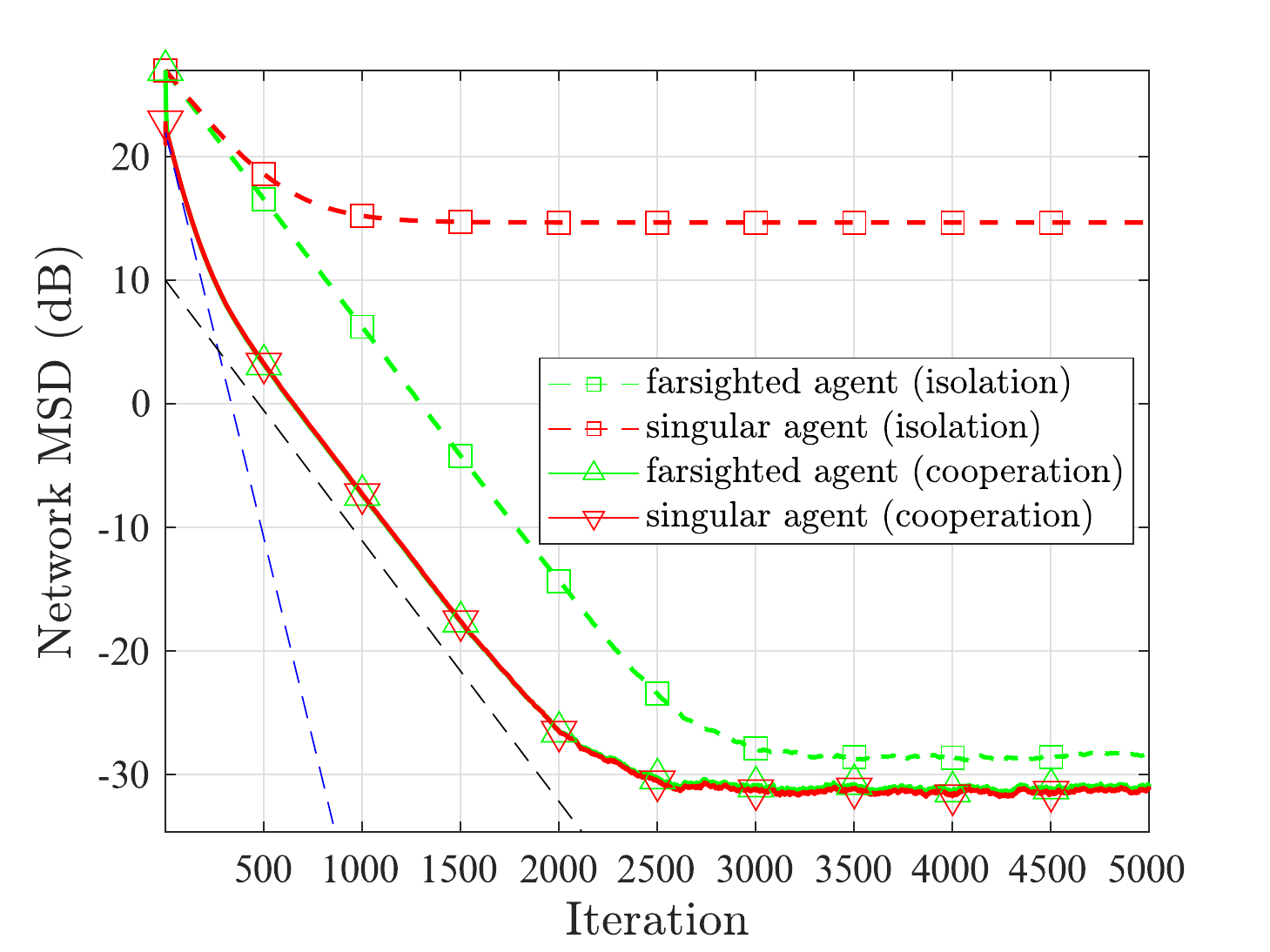}
    \caption{
    ACTC network mean-square-deviation in \eqref{eq:netMSD} as a function of the iteration $i$, for a {\em left-stochastic} combination matrix obtained through the Metropolis-Hastings procedure to match a target Perron vector that optimizes the (unquantized) ATC performance. In order to match the convergence rate $\rho_{\rm{cen}}$ corresponding to Fig.~\ref{fig:5}, we here used $\zeta = 0.8$, and with equal step-sizes $\mu_k=\mu=16 \times10^{-3}$, while leaving all other system parameters unchanged.}
    \label{fig:6}
\end{figure}

\subsection{Unidentifiable Problem}
\label{subsec:unident}
We now move on to consider the challenging case where only one agent (say, agent $1$) has a locally identifiable regression problem. 

Technically, the regressors' matrix $R_{u,1}$ is invertible, whereas the remaining matrices $R_{u,k}$, for $k>1$, are singular. 
In particular, agents $k=2,3,\ldots,N$ solve a regression problem with two linearly dependent features. 
In the following, agent $1$ will be referred to as ``farsighted agent'', whereas the other agents as ``singular agents''.
Moreover, we assume that all agents have the same regressor and noise variances $\sigma^2_u$ and $\sigma^2_v$, and we consider equal step-sizes at all agents, such that the vector $p$ coincides with the Perron eigenvector.

Under this setting the local cost functions of the singular agents are {\em not} strongly convex, whereas the aggregate cost function in \eqref{eq:coopMSE} is strongly convex, with constant $\nu$ given by:
\beq
\nu=2 p_1 \sigma^2_u,
\eeq
where $p_1$ is the entry of the Perron eigenvector corresponding to the farsighted agent.

In Fig.~\ref{fig:5}, we consider a doubly-stochastic combination policy, namely, the Metropolis rule.
Five main conclusions arise. 
First, agent $1$ in isolation is able to learn fairly well, with a steady-state error $\approx - 26~{\rm dB}$, while the other agents, when in isolation, are unable to learn properly, with steady-state errors $\approx 18~{\rm dB}$.\footnote{Actually, these errors depend on the initial conditions since the singular agents have a {\em continuum} of minimizers.}  
Second, when organized into a network, {\em all agents are able to learn properly and even with only $3$ bits per iteration!}
Third, after the initial network transient, and before the final coordinated transient (dashed black line), an intermediate transient arises (dashed blue line), which can be shown to represent the time needed by the singular agents to align with the farsighted one.\footnote{This further transient is not visible in our formulas, since it is absorbed in the upper-bound corresponding to the slower transient dominated by $(1-\mu\,\zeta\,\nu)^2$. By examining the considered example, it is possible to evaluate analytically this transient, but the analysis is beyond the scope of this work.}
Fourth, despite the fact that $N-1$ agents have a singular regressors' matrix, they contribute to accelerate convergence to the steady state. However, we see that the cooperative steady-state performance is equivalent to the individual (i.e., non-cooperative) performance of the farsighted agent. We will now show that this conclusion is not a general conclusion, and depends on the particular combination policy.

To this end, in Fig.~\ref{fig:6} we consider the same setting of Fig.~\ref{fig:5}, but for the choice of the combination policy, which is now left-stochastic. Specifically, we use the Metropolis-Hastings rule to construct a combination matrix with a target Perron eigenvector --- see the procedure explained in~\cite[p. 630]{Sayed}. The target Perron eigenvector is chosen so as to minimize the steady-state error of the unquantized ATC strategy. Clearly, this does not guarantee that we are choosing the best Perron eigenvector for the ACTC strategy, but we will now see that this turns out to be a meaningful choice. 
In fact, we see from Fig.~\ref{fig:6} that, with the optimized combination policy, network cooperation achieves a twofold goal. 
As in the case of a doubly-stochastic policy, cooperation is beneficial to the singular agents. 
Moreover, it is also beneficial to the farsighted agent, which is now able to improve on the steady-state performance achieved without cooperation. 

In summary, the conducted experiments lead to a revealing conclusion as regards the role of topology on the learning performance. By suitable design of the combination matrix, the regularization action played by agent $1$ makes the singular agents capable of contributing more fully to the optimization problem, allowing all agents to achieve a mean-square-deviation that outperforms the non-cooperative performance achievable by the farsighted agent in isolation.  

\begin{figure}[t]
    \centering
    \includegraphics[width=0.6\linewidth]{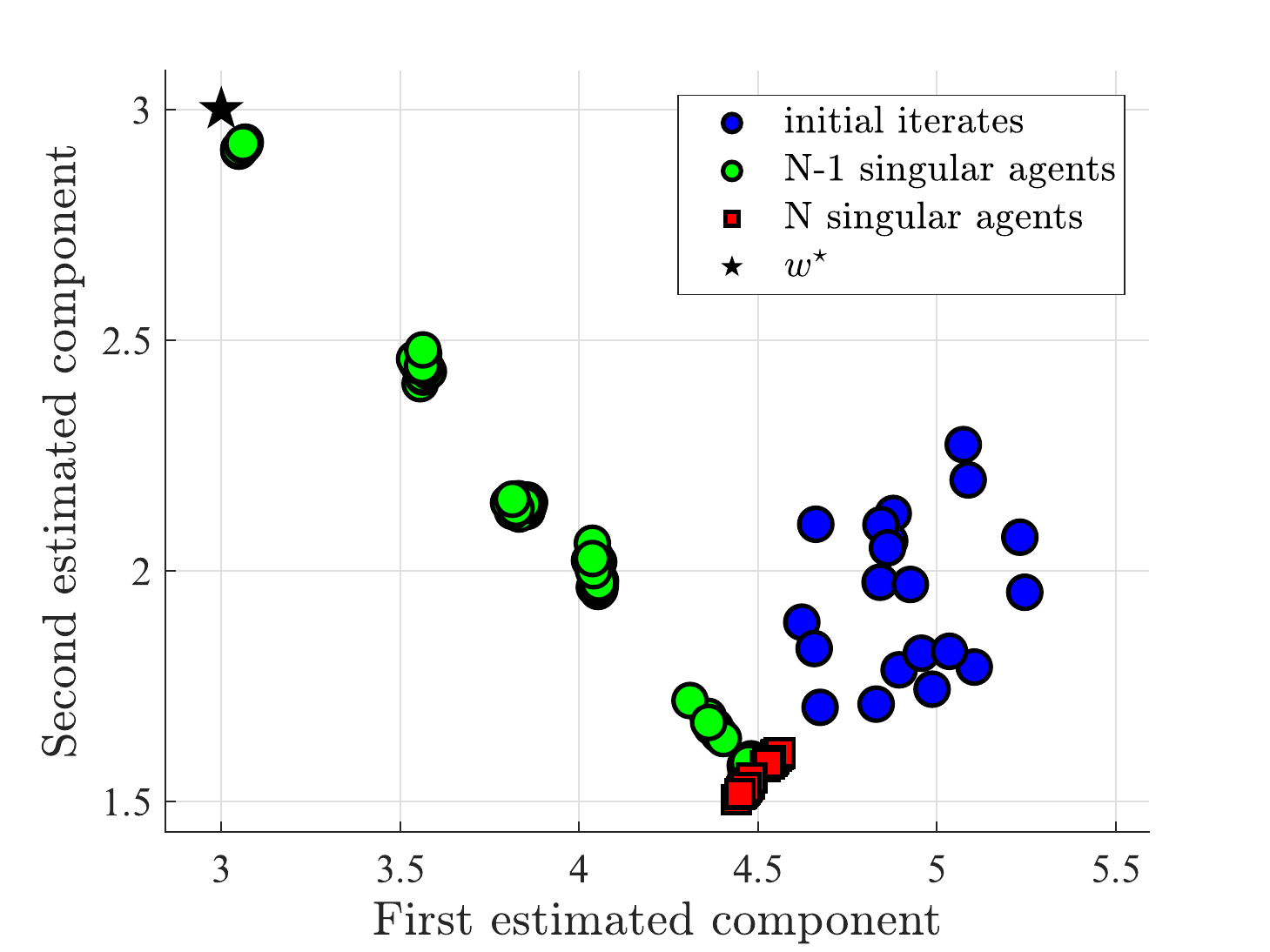}
    \caption{Illustrative example of time-evolution of the individual agents' iterates for the ACTC strategy.}
    \label{fig:7}
\end{figure}

Finally, the convergence behavior of the ACTC strategy is visually illustrated in Fig.~\ref{fig:7}, with reference to a simple example with dimensionality $M=2$, and $N=20$ agents. Let us consider first the case where the $N$ agents are all singular (red squares). In this case, we see that, moving from the initial iterates (blue circles), the singular agents follow wrong paths converging around the wrong point $(4.5, 1.5)\neq w^{\star}$. In contrast, the ACTC strategy (green circles) allows all agents to converge well to a small neighborhood of the true minimizer, after an initial transient where they need to coordinate with each other.  

\subsection{Comparison with Existing Strategies}
\begin{figure}[t]
    \centering
    \includegraphics[width=0.6\linewidth]{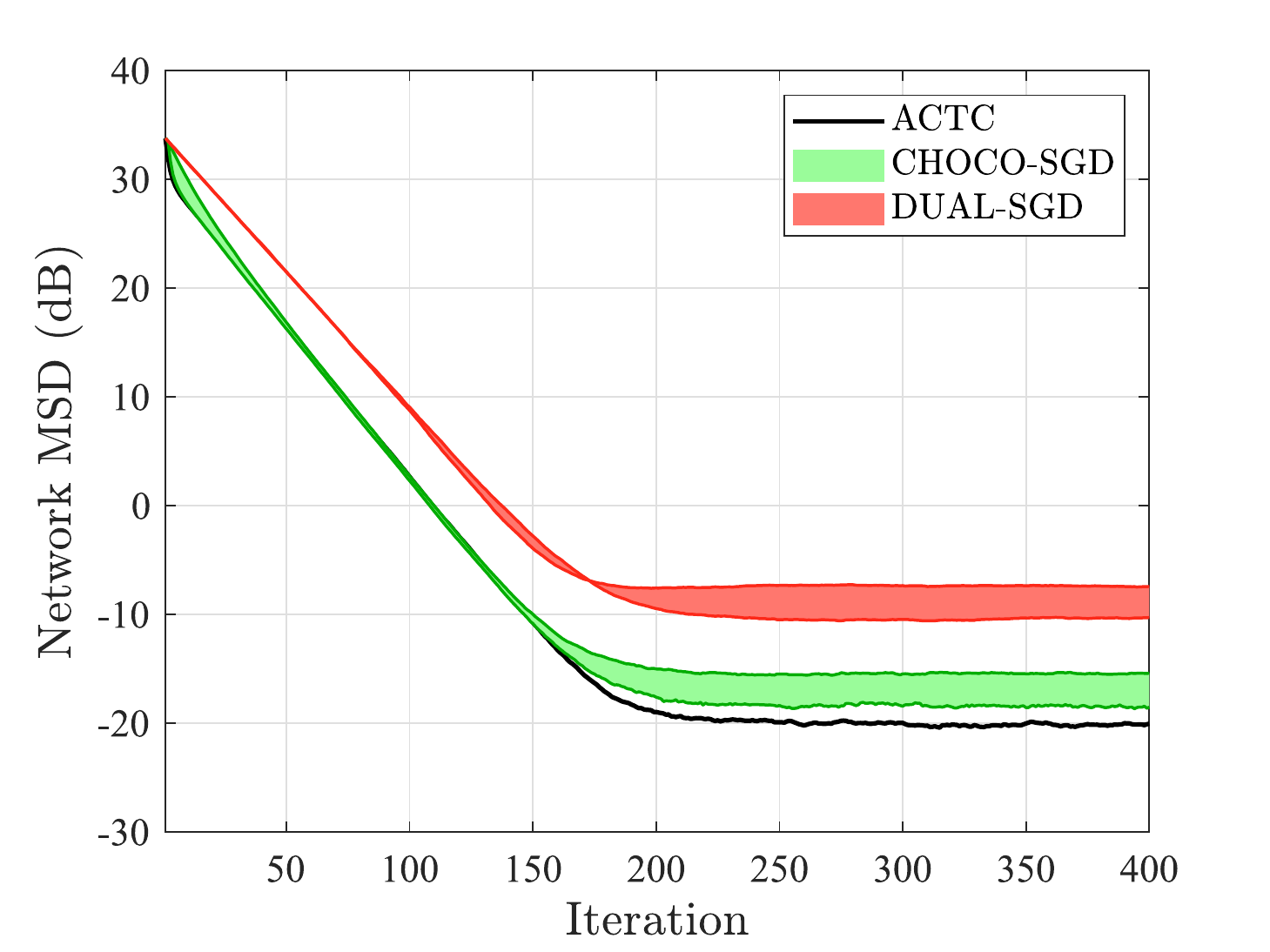}
    \caption{Comparison of the proposed ACTC strategy against known strategies, namely, CHOCO-SGD~\cite{KoloskovaStichJaggiICML2019} and DUAL-SGD~\cite{BitsForFree}. The shaded areas correspond to the best subsets of tuning parameters of CHOCO-SGD and DUAL-SGD explored during our experiments.}
    \label{fig:8}
\end{figure}

As we have illustrated in Sec.~\ref{sec:DistQua}, the present work generalizes the existing works on compressed distributed implementations under several aspects, including left-stochastic combination policies, lack of local strong convexity, diffusion strategies. For this reason, the existing theoretical results cannot cover the challenging setting considered in the present work, which required instead a significant additional effort. 
Nevertheless, even if formulated and studied under alternative settings, some of the existing algorithms can be practically applied to our setting. 
In particular, we select from the existing algorithms two particular up-to-date implementations that, as far as we know, constitute the actual benchmark performance, namely, CHOCO-SGD~\cite{KoloskovaStichJaggiICML2019} and its dual version, DUAL-SGD~\cite{BitsForFree}. Notably, the latter two algorithms have more tuning parameters than our algorithm. Even if the necessity itself of tuning more parameters might be considered a disadvantage of these strategies, in order to ensure a fair comparison we performed a fine tuning of all the parameters to guarantee best performance of CHOCO-SGD and DUAL-SGD. The shaded areas shown in Fig.~\ref{fig:8} correspond to the range of mean-square-deviations spanned by a subset of the parameters explored during the tuning phase.

Figure~\ref{fig:8} displays the comparison involving the proposed ACTC strategy and the aforementioned two strategies. 
Remarkably, for the same value of the transient time, the ACTC strategy outperforms both CHOCO-SGD and DUAL-SGD at steady state. 
In particular, we see that DUAL-SGD performs appreciably worser than the implementations in the primal domain. This conclusion is in perfect agreement with was shown in~\cite{SayedPrimalDual} for the uncompressed case. In fact, the core of DUAL-SGD is a primal-dual distributed strategy of Arrow-Hurwicz type, which was shown in~\cite[Corollary 3]{SayedPrimalDual} to converge, despite being a distributed cooperative strategy, at most to the non-cooperative performance. 
In contrast, both the ACTC and CHOCO-SGD strategies are able to exploit fully the distributed cooperation, which explains the performance improvement exhibited in Fig.~\ref{fig:8}.\footnote{In~\cite{BitsForFree}, with reference to a regularized logistic regression example, it is shown that DUAL-SGD and CHOCO-SGD perform similarly, but this comparison is in terms of function values and not in terms of iterates.}

We move on to examine the improvement of the proposed ACTC strategy on the existing CHOCO-SGD strategy. Also in this case, this improvement can be neatly explained in the light of known behavior observed in the uncompressed case, since the improvement matches well similar gains achievable when using diffusion (as ACTC does) as opposed to consensus (as CHOCO-SGD does). 
In fact, as observed in distributed optimization without compression~\cite{Sayed}, diffusion strategies can outperform consensus strategies, and, remarkably, from our experiments we observe the same behavior when these types of strategies are called to operate under communication constraints.

\section{Conclusion}

We considered a network of agents tasked to solve a certain distributed optimization problem from continual aggregation of {\em streaming observations}.
Fundamental features of our setting are {\em adaptation}, {\em local cooperation} and {\em data compression}. 
By adaptation we mean that the agents must be able to react promptly to drifts in the operational conditions, so as to adapt their inferential solution quickly. In this regard, stochastic-gradient algorithms with {\em constant step-size} become critical.
By local-cooperation we mean that each individual agent is allowed to implement a {\em distributed} algorithm by exchanging information with its neighbors.
Finally, data compression comes from the need of communicating information at a finite rate, owing to energy/bandwidth constraints. 

We introduced a novel strategy nicknamed as Adaptive-Compress-Then-Combine (ACTC), whose core is an {\em adaptive diffusion} strategy properly twinned with a {\em differential stochastic compression} strategy.
Our analysis is conducted under the challenging setting where: $i)$ communication is allowed to be unidirectional (i.e., over {\em directed} graphs); and $ii)$ the cost functions at the individual agents are allowed to be {\em non-convex}, provided that a global cost function obtained as linear combination of the local cost functions is {\em strongly convex}.
We obtained the following main results. 
First, we established that the proposed ACTC scheme is mean-square stable, and in particular that {\em each individual agent} is able to infer well the value of the parameter to be estimated, with a mean-square-deviation vanishing proportionally to the step-size.
Second, we characterized the learning behavior of each individual agent, obtaining analytical solutions that highlight the existence of two main transient phases, one (faster) relative to convergence of all agents to a {\em coordinated evolution}, the other (slower) relative to convergence of the coordinated estimate to the steady-state solution. 
Notably, these distinct learning phases were shown to emerge in diffusion strategies {\em without data compression}~\cite{ChenSayedTIT2015part1}. 
Therefore, our result implies that these distinct phases are preserved {\em despite the presence of compressed data, for any degree of compression}. 
Moreover, there are also distinguishing features arising from data compression, and the obtained analytical solutions are able to reflect well the role of the compression degree (e.g., quantization bits) in the final learning behavior.
A remarkable conclusion stemming from our analysis is that, for sufficiently small step-sizes, small errors are achievable for {\em any compression degree}. This behavior brings an interesting analogy with classical information-theoretic results: the information-rate limitation does not preclude the possibility of learning, but involves a reduction in the speed of convergence.

Another useful parallel can be drawn with the recently introduced paradigm of {\em exact diffusion}~\cite{YuanYingZhaoSayedSP2019part1,YuanYingZhaoSayedSP2019part2}, where no compression is present, and the {\em true} (i.e., non stochastic) gradient is available. 
Under this paradigm, diffusion strategies with constant step-size $\mu$ are enriched with an error compensation step, which allows them to attain a zero, rather than $O(\mu)$, mean-square-deviation~\cite{YuanYingZhaoSayedSP2019part1,YuanYingZhaoSayedSP2019part2}. 
Inspired by the structure of the ACTC algorithm in \eqref{ACTC}, it could be worth including in the exact diffusion algorithm the parameter $\zeta$ and a general nonlinearity $\bm{Q}_k(\cdot)$, and tuning these two quantities to speed up convergence.

There is still a lot of work to be done in the context of distributed adaptive learning under communication constraints. 
One advance regards a steady-state performance analysis aimed at obtaining exact formulas for the mean-square-deviation.
A further useful contribution is to extend results available under non-convex environments~\cite{VlaskiSayedTSP2021part1,VlaskiSayedTSP2021part2,VlaskiSayedTAC2021} to the case of compressed data.  
Under such setting, the traditional difficulties arising from the lack of convexity (e.g., the evolution of the stochastic-gradient iterates, their mean-square stability and steady-state performance) will be complicated by the complexity arising from the introduction of the nonlinear compression operator.
Finally, an open problem that we are currently investigating regards the trade-off between number of transmissions and quantization bits, i.e., how to perform jointly the design of the topology and the allocation of bit-rate budget to maximize the performance.

\appendices

\section{Collection of Useful Background Results}
\label{app:collection}
\subsection{Jordan Representation}
\label{app:Jordan}

Let $J_{\rm{tot}}$ be the matrix associated with the canonical Jordan decomposition of the combination matrix $A$, which can be represented as~\cite{Johnson-Horn}:
\beq 
J_{\rm{tot}} = {\sf diag}\{J_1,J_2, \ldots, J_{B}\},
\eeq
where $B$ is the number of Jordan blocks. 
As usual, the individual blocks can have different size, and the unspecified off-diagonal terms arising after block-diagonal concatenation are automatically set to zero.
For $n=1,2,\ldots,B$, we denote by $\lambda_n$ the eigenvalue associated with block $J_n$, and, without loss of generality, we assume that the eigenvalues are sorted in descending order of magnitude, namely,
\beq
|\lambda_1|>|\lambda_2|\geq |\lambda_3|\geq\ldots\geq|\lambda_B|.
\eeq
Each Jordan block takes on the form:
\beq
J_{n} \triangleq \begin{bmatrix}
\lambda_n & 1 & \\
& \ddots & \ddots & \\
& & \ddots & 1 \\
& & & \lambda_n
\end{bmatrix},
\eeq
and can be accordingly written as:
\beq
J_n=\lambda_n I_{L_n} + U_{L_n},
\eeq
where $L_n$ is the dimension of the $n$-th block, and $U_{L_n}$ is a square matrix of size $L_n$ that has all zero entries, but for the first diagonal above the main diagonal, which has entries equal to $1$. 

In view Assumptions~\ref{Strong Connectivity} and~\ref{Stochastic combination matrix}, the combination matrix $A$ has a unique largest magnitude eigenvalue that is $\lambda_1=1$, i.e., the first Jordan block is $J_1=1$.
The remaining $B-1$ Jordan blocks can be conveniently arranged in the reduced matrix:
\beq 
J = {\sf diag}\{J_2, \ldots, J_{B}\}.
\label{eq:redJordan}
\eeq
Moreover, letting
\beq
\Lambda\triangleq {\sf diag}\{\lambda_2 I_{L_2},\lambda_3 I_{L_3},\ldots,\lambda_B I_{L_B}\},
\label{eq:Lambdablocks}
\eeq
and
\beq
U\triangleq {\sf diag}\{U_{L_2},U_{L_3},\ldots,U_{L_B}\},
\label{eq:Ublocks}
\eeq
we end up with the following useful representation:
\beq
J=\Lambda + U.
\label{eq:Jordanrep}
\eeq

\subsection{Energy Operators}
\label{app:energyop}
\begin{definition}[Energy Vector Operator]
\label{def:EVO}
Let $x_1, x_2,\ldots, x_N$ be $N$ vectors of size $M\times 1$, and let 
\beq
x = 
\begin{bmatrix}
x_1\\
x_2\\
\vdots\\
x_N
\end{bmatrix}
\eeq
be the block vector of size $MN \times 1$ obtained by concatenating these vectors. 
The energy vector operator, ${\sf P}: \mathbb{C}^{MN} \rightarrow \mathbb{R}^N$, is defined as:
\beq
{\sf P}[x] = 
\begin{bmatrix}
\|x_1\|^2\\
\|x_2\|^2\\
\vdots\\
\|x_N\|^2
\end{bmatrix}.
\label{eq:vectEnergyDef}
\eeq
\hfill$\square$
\end{definition}
The stochastic counterpart of operator ${\sf P}[\cdot]$ is the operator $\mathscr{P}[\cdot]$ introduced in \eqref{eq:avenopdef}, which, for a random argument $\bm{x}$, can be written in terms of ${\sf P}[\cdot]$ simply as:
\beq
\mathscr{P}[\bm{x}]=\E{\sf P}[\bm{x}].
\eeq
The block-matrix counterpart of operator ${\sf P}[\cdot]$ is defined as follows.
\begin{definition}[Norm Matrix Operator]
\label{def:NMO}
Let $\{ X_{kn} \}$, with $k = 1, 2, \ldots, K$ and $n = 1, 2, \ldots, N$, a doubly-indexed sequence of $M \times M$ matrices, and consider the $MK\times MN$ matrix $X$ whose $(k,n)$-block is $\{ X_{kn} \}$. 
The norm matrix operator, ${\sf \bar{P}}: \mathbb{C}^{MK \times MN} \rightarrow \mathbb{R}^{K \times N}$, is defined as:
\beq
\matpow[X] \triangleq \begin{bmatrix}
    \|X_{11}\| & \ldots & \|X_{1N}\| \\ 
    \vdots &  & \vdots \\ 
    \|X_{K1}\| & \ldots & \|X_{KN}\|
\end{bmatrix}.
\label{eq:normMatrixOpDef}
\eeq
\hfill$\square$
\end{definition}

The operators \eqref{eq:vectEnergyDef} and \eqref{eq:normMatrixOpDef} are equipped with several useful properties. 
We now list those properties that will be exploited in the forthcoming proofs, and refer the Reader to~\cite{ChenSayedTIT2015part1} for the proof of these properties.

\begin{property}[Energy Vector Operator and Norm Matrix Operator Properties]
Let $x$ and $y$ be two block column vectors of size $M N \times 1$ constructed as in Definition~\ref{def:EVO}, and let $X$ and $Y$ be two block matrices of size $M K \times MN$ constructed as in Definition~\ref{def:NMO}. 
The following properties hold.

\begin{enumerate}
    \item[P1)] Nonnegativity: ${\sf P}[x] \succeq 0, \matpow[X] \succeq 0$.
    \item[P2)] Scaling: For any scalar $a \in \mathbb{C}$, ${\sf P}[ax] = |a|^2\,{\sf P}[x]$ and $\matpow[aX] = |a|\,\matpow[X]$.
    \item[P3)] Convexity: given a set of $L$ vectors $\{x^{(1)}, \ldots, x^{(L)}\}$ having the same structure of $x$, and a set of convex coefficients $\{a_1, \ldots, a_L \}$, then
    \beq
    {\sf P}[a_1 x^{(1)} + \ldots + a_L x^{(L)}] \preceq a_1{\sf P}[x^{(1)}] + \ldots + a_L {\sf P}[x^{(L)}].
    \eeq
    \item[P4)] Additivity under orthogonality: 
    Let $\bm{x}$ and $\bm{y}$ be two block column random vectors of size $M N \times 1$ constructed as in Definition~\ref{def:EVO}. If blocks $\bm{x}_k$ and $\bm{y}_k$ are orthogonal for all $k=1,2,\ldots,N$, namely if
    \beq
    \E\left[ \bm{x}_k^{\mathsf{H}}\bm{y}_k\right] = 0,
    \eeq 
    then we have that:
    \beq 
    \mathscr{P}[\bm{x} + \bm{y}] = \mathscr{P}[\bm{x}] + \mathscr{P}[\bm{y}].
    \eeq
    \item[P5)] Relation to Euclidean norm: $\mathds{1}_{N}^{\top}{\sf P}[x] = \|x\|^2$.
    \item[P6)] Linear transformations: Let $Q$ be a $K \times N$ block matrix with $M \times M$ blocks. 
    Applying the energy operator to the linear transformation $Qx$ we obtain: 
    \begin{align} 
    {\sf P}[Qx] & \preceq \|\matpow[Q]\|^2_{\infty}\,\matpow[Q]\,{\sf P}[x] 
    \label{eq:LTpowForm1}\\
    & \preceq  \|\matpow[Q]\|^2_{\infty}\, \mathds{1}_{N} \mathds{1}_{N}^{\top}\,{\sf P}[x].
    \label{eq:LTpowForm2}
    \end{align}
    \item [P7)]
    Stable Kronecker Jordan operator: 
Consider the extended version of the reduced Jordan matrix $J$ in \eqref{eq:redJordan}:  
$\mathcal{J} = J \otimes I_M$
Then, for any two block column vectors $x'$ and $y'$ of size $M (N-1) \times 1$ constructed as in Definition~\ref{def:EVO}, we have that:
    \beq 
    {\sf P}\left[\mathcal{J}x' + y'\right] \preceq \Gamma\,{\sf P}[x'] + \frac{2}{1-|\lambda_2|}{\sf P}[y'],
    \eeq
    where the $\Gamma$ is the following $(N-1) \times (N-1)$ matrix:
    \beq
    \Gamma 
        \triangleq \begin{bmatrix}
        |\lambda_2| & \displaystyle{\frac{2}{1-|\lambda_2|}} & \\
        & \ddots & \ddots & \\ 
        & & \ddots &  \displaystyle{\frac{2}{1-|\lambda_2|}} \\
        & & & |\lambda_2|
        \end{bmatrix}.
    \label{eq:stableKronJord}
    \eeq
\end{enumerate}
\end{property}

\subsection{Representation in Transformed Network Coordinates}
\label{app:usefulrep}
Before proving all the pertinent lemmas and theorems, it is useful to write down the ACTC strategy in terms of  the transformed variables introduced in Sec.~\ref{sec:NCT}.
Regarding the transformed quantized vector $\widehat{\bm{q}_i}$, from the second step in (\ref{eq:netACTC}) by $\mathcal{V}$, we readily get:
\beq
\widehat{\bm{q}}_i=\mathcal{V}\widetilde{\bm{q}}_i=\widehat{\bm{q}}_{i-1}+\zeta\,\mathcal{V}\,\bm{\mathcal{Q}}\big(\mathcal{V}^{-1}\,\widehat{\bm{\delta}}_i\big).
\label{eq:hatqrecur0}
\eeq
We note in passing that the term $\mathcal{V}\,\bm{\mathcal{Q}}\big(\mathcal{V}^{-1}\,\widehat{\bm{\delta}}_i\big)$ reflects well the inherent nonlinear behavior of the compression operators. 
In fact, the linear transformation $\mathcal{V}$ and the nonlinear operator $\bm{\mathcal{Q}}(\cdot)$ do not commute and, hence, the direct and inverse network transformation, $\mathcal{V}$ and $\mathcal{V}^{-1}$, do {\em not} compensate perfectly with each other.

Let us switch to the transformed quantization-error vector $\widehat{\bm{\delta}}_i$, and focus accordingly on the first step in \eqref{eq:netACTC}. We introduce the extended Jordan matrix:
\beq
\mathcal{J}_{\rm{tot}} \triangleq J_{\rm{tot}} \otimes I_M,
\eeq
which, in view of \eqref{aJordan}, allows us to write the extended combination matrix $\mathcal{A}$ in \eqref{hCal} as:
\beq
\mathcal{A}=\mathcal{V}^{-1} \mathcal{J}_{\rm{tot}} \mathcal{V}.
\eeq
Therefore, in view of \eqref{eq:netACTC} we can write:
\beq
\label{eq:decomposedBigDrivingMatrix}
(I_{MN} - \mu\,\bm{\mathcal{H}}_{i-1})\mathcal{A}^\top-I_{MN}
= \mathcal{V}^{-1}\,(\mathcal{J}_{\rm{tot}} - \mu\,\bm{\mathcal{G}}_{i-1} - I_{MN})\,\mathcal{V},
\eeq
where we introduced the matrix:
\beq
\bm{\mathcal{G}}_{i-1} \triangleq \mathcal{V}\,\bm{\mathcal{H}}_{i-1}\mathcal{A}^\top\,
        \mathcal{V}^{-1}.
\label{eq:ScalDef}
\eeq
Substituting now \eqref{eq:decomposedBigDrivingMatrix} into the first step of \eqref{eq:netACTC} and applying the network transformation, we get:
\beq
\widehat{\bm{\delta}}_i=\mathcal{V}\bm{\delta}_i
=(\mathcal{J}_{\rm{tot}} - I_{MN} - \mu\,\bm{\mathcal{G}}_{i-1} )\widehat{\bm{q}}_{i-1} -\mu\,\widehat{\bm{s}}_i -\mu\,\widehat{b}.
\label{eq:deltransformation}
\eeq
Furthermore, by using \eqref{hCal} and \eqref{otherCal} in \eqref{eq:ScalDef}, the matrix $\bm{\mathcal{G}}_{i-1}$ can be written as:
\begin{align}
&\bm{\mathcal{G}}_{i-1}
=
(V\otimes I_M) \bm{\mathcal{H}}_{i-1} (A^{\top} \otimes I_M) (V^{-1} \otimes I_M)\nonumber\\
&=
(V\otimes I_M) \bm{\mathcal{H}}_{i-1} (A^{\top} V^{-1} \otimes I_M),
\label{eq:preintermediateSblock}
\end{align}
where we used the property of the Kronecker product $(X\otimes Z)(Y\otimes Z)=XY\otimes Z$, holding for any three matrices $X, Y, Z$ with compatible dimensions. 
Exploiting now the partitioned structure of $V$ and $V^{-1}$ in \eqref{jordan}, we can write:
\begin{align}
&V\otimes I_M=
\begin{bmatrix}
\pi^{\top}\otimes I_M\\V_R\otimes I_M
\end{bmatrix},\label{eq:intermediateSblock1}
\\
&A^{\top} V^{-1}\otimes I_M=
\begin{bmatrix}
\mathds{1}_N\otimes I_M
&
A^{\top}V_L\otimes I_M
\end{bmatrix},
\label{eq:intermediateSblock2}
\end{align}
where in the last matrix we used the equality $A^{\top}\mathds{1}_N=\mathds{1}_N$, holding since $A$ is a left-stochastic matrix.
Using \eqref{eq:intermediateSblock1} and \eqref{eq:intermediateSblock2} in \eqref{eq:preintermediateSblock} we obtain the following block-decomposition for $\bm{\mathcal{G}}_{i-1}$:
\beq
\bm{\mathcal{G}}_{i-1}=
\begin{bmatrix} 
       \bm{G}_{11,i-1} & \bm{G}_{12,i-1}\\
        \bm{G}_{21,i-1} & \bm{G}_{22,i-1}
        \end{bmatrix},
\label{eq:dBlocks}
\eeq
where
\begin{align}
    \bm{G}_{11,i-1} &= \sum_{k=1}^{N}\pi_k\,\bm{H}_{k,i-1},
\label{d11Def}
\\
    \bm{G}_{12,i-1} &=(\pi^{\top} \otimes I_M)\,\bm{\mathcal{H}}_{i-1}\,(A^{\top} V_L \otimes I_M),
\label{d12Def}
\\
    \bm{G}_{21,i-1} &=(V_R \otimes I_M)\,\bm{\mathcal{H}}_{i-1}\,(\mathds{1}_{N} \otimes I_M),
\label{d21Def}
\\
    \bm{G}_{22,i-1} &=(V_R \otimes I_M)\,\bm{\mathcal{H}}_{i-1}\,(A^{\top}V_L \otimes I_M).
\label{d22Def}
\end{align}
Combining now \eqref{eq:deltransformation} with \eqref{eq:dBlocks}, we obtain:
    \begin{align}
        \begin{bmatrix}
        \bar{\bm{\delta}}_{i}\\
        \widecheck{\bm{\delta}}_{i}
        \end{bmatrix}
&=
        \begin{bmatrix}
         - \mu \bm{G}_{11,i-1} & -\mu \bm{G}_{12,i-1}\\
        -\mu \bm{G}_{21,i-1} & \mathcal{J}-I_{M(N-1)} - \mu \bm{G}_{22,i-1}
        \end{bmatrix}
    \nonumber\\ 
    & \times
        \begin{bmatrix}
        \bar{\bm{q}}_{i-1} \\
        \widecheck{\bm{q}}_{i-1}
        \end{bmatrix}
-\mu
        \begin{bmatrix}
        \bar{\bm{s}}_{i} \\
        \widecheck{\bm{s}}_{i}
        \end{bmatrix}
-\mu
        \begin{bmatrix}
        0 \\
        \widecheck{b}
        \end{bmatrix}.
\label{quantErrSystem}
    \end{align}
We conclude this section with a lemma that will be repeatedly used in the forthcoming proofs.

\begin{lemma}[Characterization of $\bm{\mathcal{G}}_{i-1}$]
\label{Characterization of the block matrix} 
The blocks of matrix $\bm{\mathcal{G}}_{i-1}$ in \eqref{eq:dBlocks}  satisfy the following bounds.
First, the $M\times M$ symmetric matrix $\bm{G}_{11,i-1}$ in \eqref{d11Def} fulfills the bounds:
\beq
\nu I_M \leq \bm{G}_{11,i-1}\leq \eta I_M,
\label{d11Bound}
\eeq
where $\nu$ is the global-strong-convexity constant introduced in \eqref{sqNablaStrConvex} and $\eta$ is the average Lipschitz constant in \eqref{eq:LipCvxConst}.

Second, a positive constant $\sigma_{12}$ exists such that:
\beq
\|\bm{G}_{12,i-1}\| \leq \sigma_{12}.
\label{d12Bound}
\eeq
Finally, the matrices $\matpow[\bm{G}_{21,i-1}]$ and $\matpow[\bm{G}_{22,i-1}]$ obtained by applying the norm matrix operator to the matrices in \eqref{d21Def} and \eqref{d22Def}, have bounded norm, in particular we have:
\beq
\|\matpow[\bm{G}_{21,i-1}]\|_{\infty} \leq \sigma_{21},
~~\|\matpow[\bm{G}_{22,i-1}]\|_{\infty} \leq \sigma_{22},
\label{d21PowBound}
\eeq
for some positive constants $\sigma_{21}$, and $\sigma_{22}$. 
\end{lemma}
\begin{IEEEproof}
The proof relies basically on the properties of the Hessian matrices $\bm{H}_{k,i-1}$, which arise from Assumptions~\ref{Individual cost function smoothness} and~\ref{Global Strong}. 
Let us focus on \eqref{d11Bound}.  
Using \eqref{sqNablaStrConvex} and \eqref{eq:Hessian} in \eqref{d11Def} we readily obtain:
\beq
\bm{G}_{11,i-1}=\sum_{k=1}^N p_k \,\int_{0}^{1} \nabla^2 J_k(w^{\star} - t\widetilde{\bm{w}}_{k,i-1})dt\geq \nu I_M,
\eeq
which proves the lower bound in \eqref{d11Bound}. 
The upper bound is obtained by observing that:
\begin{align}
\|\bm{G}_{11,i-1}\|
&\leq
\sum_{k=1}^N p_k \left\|
\int_{0}^{1} \nabla^2 J_k(w^{\star} - t\widetilde{\bm{w}}_{k,i-1})dt
\right\|
\nonumber\\
&\leq
\sum_{k=1}^N p_k 
\int_{0}^{1} \left\|\nabla^2 J_k(w^{\star} - t\widetilde{\bm{w}}_{k,i-1})\right\| dt
\leq \eta,
\end{align}
where the first inequality is the triangle inequality, the intermediate inequality is the mean-value inequality, and the last inequality follows by \eqref{eq:LipCvxConst}.

We continue by proving \eqref{d12Bound} and \eqref{d21PowBound}. 
First, we note that $\bm{G}_{12,i-1}$, $\bm{G}_{21,i-1}$, and $\bm{G}_{22,i-1}$ have the following common structure: 
\beq
( X \otimes I_M) {\sf diag}\{ \bm{H}_{1,i-1},\ldots, \bm{H}_{N,i-1}\} (Y \otimes I_M),
\eeq
for a suitable choice of the matrices $X$ and $Y$, having made explicit the definition of $\bm{\mathcal{H}}$ in \eqref{hCal}.
The bound in \eqref{d12Bound} follows readily from the Lipschitz property in \eqref{nablaLip}. 
Regarding \eqref{d21PowBound}, we observe that we can write:
\begin{align}
&( X \otimes I_M) {\sf diag}\{ \bm{H}_{1,i-1},\ldots, \bm{H}_{N,i-1}\}=\nonumber\\
&\begin{bmatrix}
x_{11} \bm{H}_{1,i-1} & x_{12} \bm{H}_{2,i-1} &\cdots&x_{1N} \bm{H}_{N,i-1}
\\
x_{21} \bm{H}_{1,i-1} & x_{22} \bm{H}_{2,i-1} &\cdots&x_{2N} \bm{H}_{N,i-1}\\
\vdots&&\vdots&
\\
x_{N1} \bm{H}_{1,i-1} & x_{N2} \bm{H}_{2,i-1} &\cdots&x_{NN} \bm{H}_{N,i-1}
\end{bmatrix}.
\end{align}
Therefore, for $\ell,\ell'=1,2,\ldots,N$, the $(\ell,\ell')$-block of the matrix $( X \otimes I_M) {\sf diag}\{ \bm{H}_{1,i-1},\ldots, \bm{H}_{N,i-1}\}(Y\otimes I_M)$ can be written as:
\beq
\sum_{k=1}^N x_{\ell k} \,y_{k\ell'} \,\bm{H}_{k,i-1},
\eeq
showing that each $M\times M$ block of $\bm{G}_{21,i-1}$ or $\bm{G}_{22,i-1}$ is a linear combination of the Hessian matrices $\bm{H}_{k,i-1}$. 
Now, by applying the Lipschitz property in \eqref{nablaLip} to the individual Hessian matrices $\bm{H}_{k,i-1}$ in \eqref{eq:Hessian} we see that each of these matrices has bounded norm. 
Applying the norm matrix operator to $\bm{G}_{21,i-1}$ or $\bm{G}_{22,i-1}$, we conclude that all entries of $\matpow[\bm{G}_{21,i-1}]$ and $\matpow[\bm{G}_{22,i-1}]$ are bounded, and, hence, so are $\|\matpow[\bm{G}_{21,i-1}]\|_{\infty}$ and $\|\matpow[\bm{G}_{22,i-1}]\|_{\infty}$, which concludes the proof of the lemma.
\end{IEEEproof}

\section{Proof of Lemma~\ref{lem:gradnoisetx}}
\label{Gradient noise bound lemma proof}

\begin{IEEEproof}
Each component of $\bm{s}_i$ in \eqref{eq:gradnoisedeffirstappear} fulfills the following chain of inequalities:
\begin{align}
        &\E\,\|\bm{s}_{k,i} \|^2= \alpha_k^2\,\E\,\|\bm{n}_{k,i}(\bm{w}_{k,i-1}) \|^2\nonumber\\
        &\overset{(a)}{\leq} \alpha_k^2\,\beta^2_k\,\E\,\|\widetilde{\bm{w}}_{k,i-1}\|^2 + \alpha_k^2\,\sigma^2_{k} \nonumber\\
        &= \alpha_k^2\,\beta^2_k\,\E\,\Big\|\sum_{\ell \in \mathcal{N}_k} a_{\ell k}\widetilde{\bm{q}}_{\ell,i-1}\Big\|^2 + \alpha_k^2\,\sigma^2_{k} \nonumber \\
        & \overset{(b)}{\leq} \alpha_k^2\,\beta^2_k\sum_{\ell =1}^{N} a_{\ell k}\E\,\|\widetilde{\bm{q}}_{\ell,i-1}\|^2 + \alpha_k^2\,\sigma^2_{k} \nonumber \\
        & \leq \alpha_k^2\,\beta^2_k\,\E\,\|\widetilde{\bm{q}}_{i-1}\|^2 + \alpha_k^2\,\sigma^2_{k} \nonumber\\
        &=\alpha_k^2\,\beta^2_k\,\E\,\|\mathcal{V}^{-1}\,\mathcal{V}\widetilde{\bm{q}}_{i-1}\|^2 + \alpha_k^2\,\sigma^2_{k} \nonumber \\
        &\overset{(c)}{\leq} \alpha_k^2\,\beta^2_k\,\|V^{-1}\|^2\,\E\,\|\widehat{\bm{q}}_{i-1}\|^2 + \alpha_k^2\,\sigma^2_{k} \nonumber \\
        & \overset{(d)}{=} \alpha_k^2\,\beta^2_k\,\|V^{-1}\|^2\,\mathds{1}^{\top}_N\,\mathscr{P}[\widehat{\bm{q}}_{i-1}] + \alpha_k^2\,\sigma^2_{k}, 
    \label{eq:singleAgentNoiseBound}
\end{align}
where $(a)$ follows by \eqref{eq:gradNoiseBound}; $(b)$ follows by Jensen's inequality since the weights $\{a_{\ell k}\}$ are convex; $(c)$ exploits the fact that $\widehat{\bm{q}}_{i-1}=\mathcal{V}\widetilde{\bm{q}}_{i-1}$ and the equality $\|\mathcal{V}^{-1}\|$=$\|(V\otimes I_M)^{-1}\|$=$\|V^{-1}\|$; and $(d)$ follows by property P5) of the energy operators.

Recalling that the transformed gradient noise vector $\widehat{\bm{s}}_i$ is equal to $(V\otimes I_M) \bm{s}_i$, the $\ell$-th block $\widehat{\bm{s}}_{\ell,i}$, for $\ell=1,2,\ldots,N$, is:
\beq
\widehat{\bm{s}}_{\ell,i}=\sum_{k=1}^{N} v_{\ell k}\, \bm{s}_{k,i},
\label{eq:hatgradnoiseexplicitformapp}
\eeq
where $v_{\ell k}$ is the $(\ell,k)$-entry of matrix $V$.
It is useful to examine separately the coordinated-evolution component $\bar{\bm{s}}_i=\widehat{\bm{s}}_{1,i}$ and the remaining components $\widehat{\bm{s}}_{\ell,i}$, for $\ell=2,3,\ldots,N$. 
To this end, we exploit the block decomposition of matrix $V$ in \eqref{jordan}.
Regarding $\bar{\bm{s}}_i$, since $v_{1 k}=\pi_k$, from \eqref{eq:hatgradnoiseexplicitformapp} we have that:
\begin{align}
&\E\|\bar{\bm{s}}_i\|^2=\E\|\widehat{\bm{s}}_{1,i}\|^2
=\E\left\|
\sum_{k=1}^{N} \pi_k\, \bm{s}_{k,i}
\right\|^2
\leq
\sum_{k=1}^{N} \pi_k\, 
\E\|\bm{s}_{k,i}\|^2
\nonumber\\
&\leq
\|V^{-1}\|^2\,\Big(\sum_{k=1}^N \pi_k\, \alpha^2_k \,\beta_k^2\Big)\,
\mathds{1}_N^{\top}\,\mathscr{P}[\widehat{\bm{q}}_{i-1}]
+ 
\sum_{k=1}^{N} \pi_k \,\alpha_k^2\, \sigma^2_{k},
\label{eq:barS1}
\end{align}
where the first inequality is Jensen's inequality with convex weights $\{\pi_k\}$, whereas the second inequality comes from \eqref{eq:singleAgentNoiseBound}.
Likewise, for $\ell=2,3,\ldots, N$, from \eqref{eq:hatgradnoiseexplicitformapp} we can write: 
\begin{align}
\E\|\widehat{\bm{s}}_{\ell,i}\|^2 
&=
N^2 \E\left\|
\frac 1 N\sum_{k=1}^{N} v_{\ell k}\, \bm{s}_{k,i} 
\right\|^2
\!\!\leq\!
N \,
\sum_{k=1}^{N} |v_{\ell k}|^2\, \E\|\bm{s}_{k,i}\|^2 
\nonumber \\
& \leq
N\,\|V^{-1}\|^2\,\Big(\sum_{k=1}^N |v_{\ell k}|^2\, \alpha^2_k\,\beta_k^2\Big)\,
\mathds{1}_N^{\top}\,\mathscr{P}[\widehat{\bm{q}}_{i-1}]\nonumber\\
&+ 
N\,\sum_{k=1}^{N} |v_{\ell k}|^2\, \alpha^2_k\, \sigma^2_{k},
\label{eq:hatsBound}
\end{align}
where the first inequality is Jensen's inequality with uniform weights $1/N$, whereas the second inequality comes from \eqref{eq:singleAgentNoiseBound}. 
Finally, by introducing the ``squared'' counterpart of the complex matrix $V_R$, whose entries, for $\ell=1,2,\ldots,N-1$ and $k=1,2,\ldots,N$ are:
\beq
[V_{2R}]_{\ell k}=\big| [V_R]_{\ell k} \big|^2,
\eeq
and recalling the definition of the diagonal matrices $C_\alpha$, $C_\beta$, and $C_\sigma$ in Table~\ref{tab:Notation}, from \eqref{eq:barS1} and \eqref{eq:hatsBound} it is readily seen that the claim in \eqref{eq:lemma1} has been in fact proved, with the characterization of matrix $T_s$ and of the quantities $\bar{x}_s$ and $\widecheck{x}_s$ as given in Table~\ref{tab:TransferMaTable}. 
\end{IEEEproof}

\section{Proof of Lemma~\ref{lem:qnoisetx}}
\label{Quantization noise bound lemma proof}
\begin{IEEEproof}
Applying the average energy operator $\mathscr{P}[\cdot]$ to \eqref{quantErrSystem}, we obtain:
\begin{equation}
\E\|\bar{\bm{\delta}}_{i}\|^2 = \mu^2\,\E\,\|\bar{\bm{s}}_{i}\|^2 +
\mu^2\,
\E\left\|\bm{G}_{11,i-1}\bar{\bm{q}}_{i-1} + \bm{G}_{12,i-1}\widecheck{\bm{q}}_{i-1}\right\|^2
\label{quantErrorBar}
\end{equation}
\begin{align}
\mathscr{P}[\widecheck{\bm{\delta}}_{i}]=\mu^2\mathscr{P}[\widecheck{\bm{s}}_{i}]+
\mathscr{P}\Big[ &(\mathcal{J}\!-\!I_{M(N-1)} - \mu\,\bm{G}_{22,i-1})\widecheck{\bm{q}}_{i-1}
\nonumber\\
&
-\mu\,\bm{G}_{21,i-1}\bar{\bm{q}}_{i-1} - \mu\,\widecheck{b}\Big],
\label{quantErrorCheck}
\end{align}
where the energy terms corresponding to the gradient noise are additive in view of property \eqref{eq:gradNoiseUnbiased} and property P4) of the energy operator.

Let us consider the second term on the RHS of \eqref{quantErrorBar}, for which we can write: 
    \begin{align}
        &        \| \bm{G}_{11,i-1}\bar{\bm{q}}_{i-1} + \bm{G}_{12,i-1}\widecheck{\bm{q}}_{i-1}\|^2
        \nonumber\\ 
        &\leq 
        2\| \bm{G}_{11,i-1} \bar{\bm{q}}_{i-1}\|^2 
        +2\|\bm{G}_{12,i-1}\widecheck{\bm{q}}_{i-1}\|^2
        \nonumber\\
        &\leq        
        2\sigma^2_{11}\| \bar{\bm{q}}_{i-1}\|^2+
        2\sigma_{12}^2 \|\widecheck{\bm{q}}_{i-1}\|^2
        \nonumber\\
        &=
        2\sigma^2_{11}\| \bar{\bm{q}}_{i-1}\|^2+
        2\sigma_{12}^2 \mathds{1}^{\top}_{N-1}{\sf P}[\widecheck{\bm{q}}_{i-1}],
    \label{eq:quantErrorBarAlmostFinished}
\end{align}
where the first inequality is an application of Jensen's inequality, the second inequality comes from \eqref{d11Bound} and \eqref{d12Bound}, and the final equality comes from property P5) of the energy operator ${\sf P}[\cdot]$.
Taking expectations in \eqref{eq:quantErrorBarAlmostFinished} and using the result in \eqref{quantErrorBar} we obtain:
\begin{align}
\E\|\bar{\bm{\delta}}_{i}\|^2
&\leq
2\,\mu^2\,\sigma^2_{11} \E\|\bar{\bm{q}}_{i-1}\|^2
\nonumber\\
&+
2\,\mu^2\,\sigma_{12}^2 \mathds{1}^{\top}_{N-1}\mathscr{P}[\widecheck{\bm{q}}_{i-1}]
+
\mu^2\,\E\,\|\bar{\bm{s}}_{i}\|^2.
\label{eq:lemma2bar} 
\end{align}

Let us move on to examine \eqref{quantErrorCheck}. 
First of all, we appeal to the Jordan matrix representation in \eqref{eq:Jordanrep} to write:
\beq
\mathcal{J}-I_{M(N-1)}=(\Lambda-I_{N-1})\otimes I_M + U\otimes I_M
\triangleq
\mathcal{D}+\mathcal{U}.
\label{eq:JminuIdecompose}
\eeq
Then, the following chain of inequalities holds:
\begin{align}
&{\sf P}\left[ (\mathcal{J}-I_{M(N-1)}-\mu\bm{G}_{22,i-1})\widecheck{\bm{q}}_{i-1} - \mu\bm{G}_{21,i-1}\bar{\bm{q}}_{i-1} - \mu\,\widecheck{b}\right]\nonumber\\
&=
{\sf P}\left[ (\mathcal{D}+\mathcal{U}-\mu\bm{G}_{22,i-1})\widecheck{\bm{q}}_{i-1} - \mu\bm{G}_{21,i-1}\bar{\bm{q}}_{i-1} - \mu\,\widecheck{b}\right]\nonumber\\
&\overset{(a)}{\leq}
2\,{\sf P}[\mathcal{D}\,\widecheck{\bm{q}}_{i-1}] \nonumber\\ 
&+
2\,{\sf P}\left[ (\mathcal{U}-\mu\bm{G}_{22,i-1})\widecheck{\bm{q}}_{i-1} - \mu\bm{G}_{21,i-1}\bar{\bm{q}}_{i-1} - \mu\,\widecheck{b}\right]
\nonumber\\
&\overset{(b)}{\leq}
2\,{\sf P}[\mathcal{D}\,\widecheck{\bm{q}}_{i-1}] + 
8\,{\sf P}[ \mathcal{U}\,\widecheck{\bm{q}}_{i-1}]\nonumber\\
&
+8\mu^2\,{\sf P}[ \bm{G}_{22,i-1}\widecheck{\bm{q}}_{i-1}]+8\mu^2\,{\sf P}[ \bm{G}_{21,i-1}\bar{\bm{q}}_{i-1}]
+8\mu^2\,{\sf P}[\,\widecheck{b}\,]\nonumber\\
&\overset{(c)}{\leq}
2\,\|\matpow[\mathcal{D}]\|_{\infty}\, \matpow[\mathcal{D}] \,{\sf P}[\widecheck{\bm{q}}_{i-1}] + 
8\,\|\matpow[\mathcal{U}]\|_{\infty}\, \matpow[\mathcal{U}] \,{\sf P}[ \widecheck{\bm{q}}_{i-1}]\nonumber\\
&+8\mu^2\,\|\matpow[\bm{G}_{21,i-1}]\|_{\infty}^2\,\mathds{1}_{N-1} {\sf P}[ \bar{\bm{q}}_{i-1}]\nonumber\\
&
+8\mu^2\,\|\matpow[\bm{G}_{22,i-1}]\|_{\infty}^2\,\mathds{1}_{N-1} \mathds{1}_{N-1}^{\top} {\sf P}[ \widecheck{\bm{q}}_{i-1}]
+8\mu^2\,{\sf P}[\,\widecheck{b}\,]
\nonumber\\
&\overset{(d)}{\leq}
8(I_{N-1} + U) 
{\sf P}[ \widecheck{\bm{q}}_{i-1}]
\nonumber\\
&+8\mu^2\,\|\matpow[\bm{G}_{21,i-1}]\|_{\infty}^2\,\mathds{1}_{N-1} {\sf P}[ \bar{\bm{q}}_{i-1}]\nonumber\\
&+8\mu^2\,\|\matpow[\bm{G}_{22,i-1}]\|_{\infty}^2\, \mathds{1}_{N-1} \mathds{1}_{N-1}^{\top} {\sf P}[ \widecheck{\bm{q}}_{i-1}]
+8\mu^2\,{\sf P}[\,\widecheck{b}\,]\nonumber\\
&\overset{(e)}{\leq}
8(I_{N-1} + U) 
{\sf P}[ \widecheck{\bm{q}}_{i-1}]
+8\,\mu^2\sigma_{21}^2\mathds{1}_{N-1} {\sf P}[ \bar{\bm{q}}_{i-1}]
\nonumber\\
&+8\,\mu^2\sigma_{22}^2 \mathds{1}_{N-1} \mathds{1}_{N-1}^{\top} {\sf P}[ \widecheck{\bm{q}}_{i-1}]
+8\,\mu^2 {\sf P}[\,\widecheck{b}\,],
\label{eq:lemma2checkalmost}
\end{align}
where $(a)$ follows by the convexity property P3) of the energy operator applied with weights $1/2$; 
$(b)$ follows by the same property applied with weights $1/4$; 
$(c)$ follows by property P6) of the energy operator, respectively in form \eqref{eq:LTpowForm1} as regards the first two terms, and in form \eqref{eq:LTpowForm2} as regards the remaining terms; $(d)$ follows by observing that, due to the peculiar shape of $\mathcal{D}$ and $\mathcal{U}$, one has the identities:
\begin{align}
&\|\matpow[\mathcal{D}]\|_{\infty}=\|I_{N-1} - \Lambda\|=\max_{n=2,3,\ldots, N} |1-\lambda_n(A)|,\\ 
&\matpow[\mathcal{D}]\preceq  \|I_{N-1} - \Lambda\| \, I_{N-1},\\
&\|\matpow[\mathcal{U}]\|_{\infty}=1, ~~~
\matpow[\mathcal{U}]=U,
\end{align}
and by the inequality:
\beq
\max_{n=2,3,\ldots, N} |1-\lambda_n(A)|\leq 1+|\lambda_2|<2.
\label{eq:LambdaBound}
\eeq
Finally, the inequality in $(e)$ follows by the bounds in \eqref{d21PowBound}.

Taking expectations in \eqref{eq:lemma2checkalmost} and then using \eqref{quantErrorCheck} we get:
\beq
\begin{split}
& \mathscr{P}[\widecheck{\bm{\delta}}_{i}] 
\preceq
8\,\mu^2\sigma_{21}^2\mathds{1}_{N-1}\mathscr{P}[ \bar{\bm{q}}_{i-1}]
+8(I_{N-1} + U) 
\mathscr{P}[ \widecheck{\bm{q}}_{i-1}]
\\
&
+8\,\mu^2\sigma_{22}^2 \mathds{1}_{N-1} \mathds{1}_{N-1}^{\top} \mathscr{P}[ \widecheck{\bm{q}}_{i-1}]
+\mu^2\,\mathscr{P}[\widecheck{\bm{s}}_{i}]
+8\,\mu^2 {\sf P}[\,\widecheck{b}\,].        
    \end{split}
\label{eq:lemma2check}
\eeq
Examining jointly \eqref{eq:lemma2bar} and \eqref{eq:lemma2check}, we see that we have in fact proved \eqref{eq:lemma2}, with the matrix $T_\delta$ and the quantities $\bar{x}_{\delta}$ and $\widecheck{x}_\delta$ as given in Table~\ref{tab:TransferMaTable}.
\end{IEEEproof}

\section{Proof of Lemma~\ref{lem:qstaterecextended}}
\label{Recursion extended lemma proof}
We start with an auxiliary lemma that will be then used to prove Lemma~\ref{lem:qstaterecextended}.
\begin{lemma}[Quantized State Decomposition]
\label{Quantized state decomposition}
Let 
\begin{align}
\bar{\Delta} &\triangleq\|V^{-1}\|^2\, \max_{k=1,2,\ldots,N} \pi^2_k \,\omega_k,
\label{eq:deltabardef}\\
\widecheck{\Delta} &\triangleq\|V^{-1}\|^2\, \max_{\substack{\ell=2,3,\ldots,N\\k=1,2,\ldots,N}} 
|v_{\ell k}|^2 \,\omega_k,
\label{eq:deltacheckdef}
\end{align}
where $v_{\ell k}$ is the $(\ell,k)$-entry of the transformation matrix $V$ in \eqref{jordan}. 
Then, for any $\zeta \in (0,1)$ we have that:
\beq
\boxed{
\mathscr{P}[\widehat{\bm{q}}_{i}] \preceq 
\mathscr{P}[\widehat{\bm{q}}_{i-1} + \zeta\widehat{\bm{\delta}}_i]
+  \zeta^2 \Delta \mathds{1}_N^{\top} \mathscr{P}[\widehat{\bm{\delta}}_i]
}
\label{quantizedStateDecomp}
\eeq
where
\beq
\Delta\triangleq 
\begin{bmatrix}
\bar{\Delta}
\\
\\
\widecheck{\Delta}\mathds{1}_{N-1}
\end{bmatrix}.
\eeq
\end{lemma}

\begin{IEEEproof}
By adding and subtracting $\zeta \widehat{\bm{\delta}}_i$ in \eqref{eq:hatqrecur0} we can write:
\begin{align}
\widehat{\bm{q}}_i= 
\underbrace{\widehat{\bm{q}}_{i-1}+\zeta\,\widehat{\bm{\delta}}_i}_{\bm{x}}
+
\underbrace{
\zeta 
\left(
\mathcal{V}\,\bm{\mathcal{Q}}\big(\mathcal{V}^{-1}\,\widehat{\bm{\delta}}_i\big)
-\widehat{\bm{\delta}}_i
\right)}_{\bm{y}}.
\label{eq:hatqrecur}
\end{align}
Consider now two realizations of $\widehat{\bm{q}}_{i-1}$ and $\widehat{\bm{\delta}}_i$, which means that $\bm{x}=x$ becomes deterministic and that $\bm{y}$ contains only the randomness arising from the stochastic compression operator $\bm{\mathcal{Q}}(\cdot)$.
From the quantizer's unbiasedness property \eqref{unbiasedComp} we have that:
\beq
\E\left[\mathcal{V}\,\bm{\mathcal{Q}}(\mathcal{V}^{-1}\,x)\right]=
\mathcal{V}\,\E\left[\bm{\mathcal{Q}}(\mathcal{V}^{-1}\,x)\right]=x,
\eeq
or:
\beq
\E\left[
x^{\mathsf{H}} \bm{y}
\right]=0.
\label{eq:orthogonprinc3}
\eeq
Since the randomness in the compression operator is independent of all the other random mechanisms in the system and is independent over time, the orthogonality condition in \eqref{eq:orthogonprinc3} holds also when expectations are computed w.r.t. all random variables, which allows us to apply property P4) of the energy operator in \eqref{eq:hatqrecur}, yielding:
\begin{align}
\mathscr{P}\left[\widehat{\bm{q}}_{i}\right] = 
\mathscr{P}\left[\widehat{\bm{q}}_{i-1}+\zeta \widehat{\bm{\delta}}_i\right]
+\zeta^2\mathscr{P}\left[
\mathcal{V}\,\bm{\mathcal{Q}}(\mathcal{V}^{-1}\,\widehat{\bm{\delta}}_i)-\widehat{\bm{\delta}}_i
\right].\nonumber\\
\label{eq:orthogonprinc2}
\end{align}
On the other hand, recalling that $\bm{\delta}_i=\mathcal{V}^{-1} \widehat{\bm{\delta}}_i$, we can write:
    \begin{align}
&\mathcal{V}\,\bm{\mathcal{Q}}(\mathcal{V}^{-1}\,\widehat{\bm{\delta}}_i)-\widehat{\bm{\delta}}_i=
\mathcal{V}\,\left(\bm{\mathcal{Q}}(\bm{\delta}_i)-\bm{\delta}_i\right)
\nonumber\\
&= 
        \begin{bmatrix}
            v_{11}I_M  & \dots & v_{1N}I_M\\
            &  \vdots \\
            v_{N1}I_M  & \dots & v_{NN}I_M
        \end{bmatrix} 
        \begin{bmatrix}
            \bm{Q}(\bm{\delta}_{1,i})-\bm{\delta}_{1,i}\\
            \vdots \\
            \bm{Q}(\bm{\delta}_{N,i})-\bm{\delta}_{N,i}
        \end{bmatrix} 
        \nonumber\\
       & = 
        \begin{bmatrix}
            \sum_{k=1}^{N}  v_{1k} \Big(\bm{Q}(\bm{\delta}_{k,i})-\bm{\delta}_{k,i}\Big)\\
            \vdots \\
            \sum_{k=1}^{N}  v_{Nk} \Big(\bm{Q}(\bm{\delta}_{k,i})-\bm{\delta}_{k,i}\Big)
        \end{bmatrix}.
\label{matrixProdLemma4}
    \end{align} 
The expected energy of the $\ell$-th block in \eqref{matrixProdLemma4} is:
\begin{align}
& \E\, \left\|
\sum_{k=1}^{N}  v_{\ell k} \Big(\bm{Q}(\bm{\delta}_{k,i})-\bm{\delta}_{k,i}\Big)
\right\|^2 \nonumber\\
&\overset{(a)}{=}\sum_{k=1}^{N}  |v_{\ell k}|^2\,
\E\left\|\bm{Q}(\bm{\delta}_{k,i})-\bm{\delta}_{k,i}
\right\|^2 
\nonumber\\
&\overset{(b)}{\leq}
\sum_{k=1}^{N} |v_{\ell k}|^2\,\omega_k \,\E\,\| \bm{\delta}_{k,i} \|^2,
    \label{normSimplQuantErr}
\end{align}
where $(a)$ follows from the fact that the compression operators are independent across agents and unbiased, and $(b)$ follows by the non blow-up property in \eqref{boundVarianceComp}.
Recalling that the first row of matrix $V$ is the (transposed) Perron eigenvector, the first entry ($\ell=1$) in \eqref{normSimplQuantErr} can be upper bounded by:
\beq
\max_{k=1,2,\ldots,N} \pi^2_k\,\omega_k\,\,
\underbrace{\E\|\bm{\delta}_{i}\|^2}_{\mathcal{V}^{-1}\,\widehat{\bm{\delta}}_{i}}
\leq
\bar{\Delta}\,\E\| \widehat{\bm{\delta}}_{i}\|^2,
\label{eq:barDelta2}
\eeq
where the last inequality follows by the definition of $\bar{\Delta}$ in \eqref{eq:deltabardef}. 
Likewise, the other entries ($\ell\neq 1$) in \eqref{normSimplQuantErr} can be upper bounded by:
\beq
\max_{\substack{\ell=2,3,\ldots,N\\k=1,2,\ldots,N}} 
|v_{\ell k}|^2\,\omega_k\,\,
\E\|\bm{\delta}_{i}\|^2
\leq
\widecheck{\Delta}\,\E\| \widehat{\bm{\delta}}_{i}\|^2,
\label{eq:checkDelta2}
\eeq
having used the definition of $\widecheck{\Delta}$ in \eqref{eq:deltacheckdef}. 
The claim of the lemma follows by joining \eqref{eq:orthogonprinc2}, \eqref{normSimplQuantErr}, \eqref{eq:barDelta2} and \eqref{eq:checkDelta2}, and using property P5) of the energy operator.
\end{IEEEproof}
\begin{IEEEproof}[Proof of Lemma~\ref{lem:qstaterecextended}]
The first term on the RHS in (\ref{quantizedStateDecomp}) can be represented in block form as follows:
\beq
\mathscr{P}[\widehat{\bm{q}}_{i-1}+\zeta\widehat{\bm{\delta}}_i]=
\begin{bmatrix}
\E\|\bar{\bm{q}}_{i-1}+\zeta\bar{\bm{\delta}}_i\|^2\\
\\
\mathscr{P}[\widecheck{\bm{q}}_{i-1}+\zeta\widecheck{\bm{\delta}}_i]
\end{bmatrix}.
\label{eq:lemma3startingpoint}
\eeq 
Let us start by examining the first block in \eqref{eq:lemma3startingpoint}. 
Exploiting the block decomposition in \eqref{quantErrSystem} we can write:
\begin{align}
&\bar{\bm{q}}_{i-1}+\zeta\bar{\bm{\delta}}_i=
\nonumber\\
    &=
    (I_M - \mu\,\zeta\,\bm{G}_{11,i-1})\bar{\bm{q}}_{i-1} - \mu\,\zeta\bm{G}_{12,i-1}\widecheck{\bm{q}}_{i-1} - \mu\,\zeta\,\bar{\bm{s}}_{i}.
\label{eq:qbarfirstdec}
\end{align}   
First of all, using \eqref{eq:gradNoiseUnbiased} we have the equality:
\begin{align}
&\E\|\bar{\bm{q}}_{i-1}+\zeta\bar{\bm{\delta}}_i\|^2
=
\mu^2\zeta^2\E\|\bar{\bm{s}}_i\|^2
\nonumber\\
&+\E\|(I_M - \mu\zeta\,\bm{G}_{11,i-1})\bar{\bm{q}}_{i-1} - \mu\zeta\bm{G}_{12,i-1}\widecheck{\bm{q}}_{i-1}\|^2.
\label{eq:firstqbarineq}
\end{align}
Let us now examine the spectral radius of $I_M - \mu\zeta\,\bm{G}_{11,i-1}$.
Using \eqref{d11Bound} we can write:
\begin{align}
\|I_M - \mu\zeta\,\bm{G}_{11,i-1}\|^2
\leq
\max\Big\{(1-\mu\zeta \eta)^2, (1-\mu\zeta \nu)^2\Big\}.\nonumber\\
\label{eq:radImminusmu}
\end{align}
We have the following chain of equivalent relationships:\footnote{Condition \eqref{eq:mustabnew} is not the tightest condition one can use to guarantee stability of $I_M - \mu\zeta\bm{G}_{11,i-1}$. Some examples of how to get a better constant can be found in~\cite{ChenSayedTIT2015part1,Sayed}, and, with straightforward algebra, we can get the refined upper bound $\|I_M - \mu\zeta\,\bm{G}_{11,i-1}\|\leq 1 - \mu\zeta\nu+1/2(\mu\zeta\eta)^2$. However, in our analysis the additional $O(\mu^2)$ term is expected to bring little information.
In fact, as we will see in Lemma~\ref{lem:Tstab} further ahead, a number of $O(\mu^2)$ terms will be collected into a large correction constant $\phi$ that characterizes the stability analysis, with a stability threshold $\mu^{\star}$ being usually smaller than the factor $2/(\eta+\nu)$ that will be obtained from our characterization of $I_M - \mu\zeta\bm{G}_{11,i-1}$.}
\begin{align}
(1-\mu\zeta \nu)^2 &> (1-\mu\zeta \eta)^2
\nonumber\\
&\Updownarrow 
\nonumber\\
(\mu\zeta \nu)^2 -2\mu\zeta\nu
&>(\mu\zeta \eta)^2 -2\mu\zeta\eta 
\nonumber\\
&\Updownarrow
\nonumber\\
\mu\zeta \nu^2 -2 \nu
&>
\mu\zeta \eta^2 -2\eta 
\nonumber\\
&\Updownarrow
\nonumber\\
\mu\zeta \big(\eta^2 - \nu^2\big) 
&<
2 \big(\eta-\nu\big)
\nonumber\\
&\Updownarrow
\nonumber\\
&\!\!\!\!\!\!\!\textnormal{Eq. \eqref{eq:mustabnew}},
\label{eq:ineqseqmustab}
\end{align}
where the last implication is true because $\nu\leq\eta$ in view of \eqref{eq:LipCvxConst}. 
Since all implications in \eqref{eq:ineqseqmustab} hold in both directions, we have in fact proved that:
\beq
\|I_M - \mu\zeta\,\bm{G}_{11,i-1}\|^2 \leq (1-\mu\zeta \nu)^2.
\label{eq:radImminusmu2}
\eeq
Moreover, since $\nu\leq\eta$, we also have:
\beq
\frac{2}{\eta+\nu}<\frac{1}{\nu},
\eeq
which, using \eqref{eq:mustabnew}, implies that $1-\mu\zeta\nu>0$, finally yielding, in view of \eqref{eq:radImminusmu2}:
\beq
\|I_M - \mu\zeta\,\bm{G}_{11,i-1}\| \leq 1-\mu\zeta \nu.
\eeq
This upper bound will be useful in characterizing the last term in \eqref{eq:firstqbarineq}, which can be manipulated as follows, for $0<t<1$:
\begin{align}
&\|(I_M - \mu\zeta\,\bm{G}_{11,i-1})\bar{\bm{q}}_{i-1} - \mu\zeta\bm{G}_{12,i-1}\widecheck{\bm{q}}_{i-1}\|^2
\nonumber\\
&=
\left\|(1-t)\frac{I_M - \mu\zeta\,\bm{G}_{11,i-1}}{1-t}\bar{\bm{q}}_{i-1} - t \frac {\mu\zeta\bm{G}_{12,i-1}}{t}\widecheck{\bm{q}}_{i-1}\right\|^2
\nonumber\\
&\leq
\frac{\|(I_M - \mu\zeta\,\bm{G}_{11,i-1})\bar{\bm{q}}_{i-1}\|^2}{1-t} 
+
\mu^2\zeta^2\frac{\|\bm{G}_{12,i-1}\widecheck{\bm{q}}_{i-1}\|^2}{t}
\nonumber\\
&\leq
(1-\mu\zeta\nu)^2\frac{\|\bar{\bm{q}}_{i-1}\|^2}{1-t} 
+
\mu^2\zeta^2\sigma_{12}^2\frac{\|\widecheck{\bm{q}}_{i-1}\|^2}{t}
\nonumber\\
&=
(1-\mu\zeta\nu)\|\bar{\bm{q}}_{i-1}\|^2
+
\mu\zeta\frac{\sigma_{12}^2}{\nu}\|\widecheck{\bm{q}}_{i-1}\|^2,
\label{eq:lemma3baralmost}
\end{align}
where the first inequality is an application of Jensen's inequality, whereas the second inequality follows by setting $t=\mu\zeta\nu$, and by using \eqref{d12Bound}.
Taking expectations in \eqref{eq:lemma3baralmost} and using the result in \eqref{eq:firstqbarineq} we get:
\begin{align}
\E\|\bar{\bm{q}}_{i-1}+\zeta\bar{\bm{\delta}}_i\|^2
&\leq
(1-\mu\zeta\nu)\E\|\bar{\bm{q}}_{i-1}\|^2
\nonumber\\
&+
\mu\zeta\frac{\sigma_{12}^2}{\nu}\E\|\widecheck{\bm{q}}_{i-1}\|^2
+
\mu^2\zeta^2\E\|\bar{\bm{s}}_i\|^2.
\label{eq:lemma3bar}
\end{align}
We continue by examining the second block in \eqref{eq:lemma3startingpoint}. 
Using the block decomposition in \eqref{quantErrSystem} we can write:
\begin{align}
&\widecheck{\bm{q}}_{i-1}+\zeta\widecheck{\bm{\delta}}_i=
(1-\zeta)\widecheck{\bm{q}}_{i-1} \nonumber\\
&+ \zeta
\underbrace{
\left(-\mu\,\bm{G}_{21,i-1}\bar{\bm{q}}_{i-1} + (\mathcal{J}-\mu\,\bm{G}_{22,i-1})\widecheck{\bm{q}}_{i-1} -\mu\,\widecheck{\bm{s}}_{i}-\mu\,\widecheck{b}\right)}_{\bm{y}},
\label{eq:qcheckfirstdec}
\end{align}
and, using \eqref{eq:gradNoiseUnbiased} along with property P4) of the energy operator, we get:
\beq
\mathscr{P}[\widecheck{\bm{q}}_{i-1}+\zeta\widecheck{\bm{\delta}}_i]=\mu^2\zeta^2 \mathscr{P}[\widecheck{\bm{s}}_{i}]
+
\mathscr{P}[(1-\zeta)\widecheck{\bm{q}}_{i-1}+ \zeta\bm{y}].
\label{eq:qcheckfirstdec2}
\eeq
On the other hand, by the convexity property P3) of the energy operator we have:
\beq
{\sf P}[(1-\zeta)\widecheck{\bm{q}}_{i-1}+ \zeta\bm{y}]\preceq
(1-\zeta){\sf P}[\widecheck{\bm{q}}_{i-1}]+\zeta{\sf P}[\bm{y}].
\label{eq:lemma3cvx}
\eeq
Making explicit the definition of $\bm{y}$ in \eqref{eq:qcheckfirstdec}, we can write the following chain of inequalities: 
\begin{align}
&{\sf P}[\bm{y}]={\sf P}\left[
\mathcal{J}\widecheck{\bm{q}}_{i-1}
-\mu\,\bm{G}_{21,i-1}\bar{\bm{q}}_{i-1} - \mu\,\bm{G}_{22,i-1}\widecheck{\bm{q}}_{i-1} - \mu\,\widecheck{b})\right]  
\nonumber\\ & 
\overset{(a)}{\preceq} 
\underbrace{
\left(\frac{\Lambda \Lambda^*}{|\lambda_2|} + \frac{2\,U}{1-|\lambda_2|}\right)
}_{\Gamma}
\,{\sf P}[\widecheck{\bm{q}}_{i-1}] 
\nonumber\\
&+ \frac{2\mu^2}{1 - |\lambda_2|}{\sf P}\left[\bm{G}_{21,i-1}\bar{\bm{q}}_{i} + \bm{G}_{22,i-1}\widecheck{\bm{q}}_{i} +\widecheck{b}\right]  
\nonumber\\ & 
\overset{(b)}{\preceq} \Gamma{\sf P}[\widecheck{\bm{q}}_{i-1}] 
+\frac{2\mu^2}{1-|\lambda_2|}
\nonumber\\
& \times\Big( 
3\|\matpow[\bm{G}_{21,i-1}]\|^2_{\infty}\,\mathds{1}_{N-1}\|\bar{\bm{q}}_{i-1}\|^2
\nonumber\\
&+
3\|\matpow[\bm{G}_{22,i-1}]\|^2_{\infty}\, \mathds{1}_{N-1} \mathds{1}_{N-1}^{\top} {\sf P}[\widecheck{\bm{q}}_{i-1}]
+3{\sf P}[\,\widecheck{b}\,]
\Big)
\nonumber\\
&\overset{(c)}{\preceq} 
\Gamma {\sf P}[\widecheck{\bm{q}}_{i-1}] 
+\frac{6\mu^2}{1-|\lambda_2|}
\nonumber\\
& \times\Big( 
\sigma_{21}^2\mathds{1}_{N-1} \|\bar{\bm{q}}_{i-1}\|^2
+
\sigma_{22}^2 \mathds{1}_{N-1} \mathds{1}_{N-1}^{\top} {\sf P}[\widecheck{\bm{q}}_{i-1}]
+{\sf P}[\,\widecheck{b}\,]
\Big),
\label{eq:qcheckPowSmartDecomp}
\end{align}
where step $(a)$ applies property P7) of the energy operator, step $(b)$ applies property P3) with weights $1/3$, and step $(c)$ uses \eqref{d21PowBound}.

Letting
\beq
E_0=\left( 
(1-\zeta) I_{N-1} + \zeta\,\frac{\Lambda\Lambda^*}{|\lambda_2|}
\right) + \displaystyle{\frac{2\zeta}{1-|\lambda_2|}}\,U,
\label{eq:E0def0}
\eeq
taking expectations in \eqref{eq:qcheckPowSmartDecomp}, and using the result in \eqref{eq:lemma3cvx} and then in \eqref{eq:qcheckfirstdec}, we obtain:
\begin{align}
&\mathscr{P}[\widecheck{\bm{q}}_{i-1}+\zeta\widecheck{\bm{\delta}}_i]
\preceq
\mu^2\zeta \frac{6\,\sigma_{21}^2}{1-|\lambda_2|}\,\mathds{1}_{N-1} \,\E\|\bar{\bm{q}}_{i-1}\|^2+
\nonumber\\
&
\left(E_0
+\mu^2\zeta \frac{6\,\sigma_{22}^2}{1-|\lambda_2|} \mathds{1}_{N-1} \mathds{1}_{N-1}^{\top}
\right)
\mathscr{P}[\widecheck{\bm{q}}_{i-1}]
\nonumber\\
&+
\mu^2\zeta\,\frac{6}{1-|\lambda_2|}{\sf P}[\,\widecheck{b}\,]
+\mu^2\zeta^2 \mathscr{P}[\widecheck{\bm{s}}_{i}].
\label{eq:lemma3check}
\end{align}
Calling upon Lemma~\ref{Quantized state decomposition} along with \eqref{eq:lemma3bar} and \eqref{eq:lemma3check}, we see that \eqref{eq:lemma3} holds true, with the matrix $T_q$ and the quantities $\bar{x}_q$ and $\widecheck{x}_q$ as given in Table~\ref{tab:TransferMaTable}.

\end{IEEEproof}

\section{Proof of Theorem~\ref{Quantized state recursion}}
\label{app:proofTheorem1}
\begin{IEEEproof}
In \eqref{eq:lemma3}, we can replace $\mathscr{P}[\widehat{\bm{\delta}}_i]$ with the RHS of \eqref{eq:lemma2} to get an inequality that relates $\mathscr{P}[\widehat{\bm{q}}_i]$ to $\mathscr{P}[\widehat{\bm{q}}_{i-1}]$ and $\mathscr{P}[\mu\,\widehat{\bm{s}}_i]$. Replacing now $\mathscr{P}[\mu\,\widehat{\bm{s}}_i]$ with the RHS of \eqref{eq:lemma1} we get the inequality recursion in \eqref{eq:systemRecursion}, with the driving vector $x$ defined in \eqref{eq:Tandx} and with the matrix $T$ replaced by the matrix:
\beq
T_{\rm{tmp}}=T_q + \zeta^2\Delta\,\mathds{1}_N^{\top}\, T_\delta +
\zeta^2 \left(\Delta\, \mathds{1}_N^{\top} + I_N\right) T_s.
\label{eq:Tapp}
\eeq
The claim of the theorem will be proved if we show that the matrix $T_{\rm{tmp}}$ is upper bounded by the matrix $T$ appearing in \eqref{eq:Trepresfirstappear}.

The term $T_q$ can be upper bounded as:
\begin{align}
T_q\preceq
\begin{bmatrix}
1-\mu\,\zeta\,\nu & \mu\,\zeta\, \displaystyle{\frac{\sigma_{12}^2}{\nu}} \mathds{1}_{N-1}^{\top}
\\
\\
\phi^{(q)}\,\mu^2\,\zeta\,\mathds{1}_{N-1}& E_0 + \phi^{(q)}\,\mu^2\,\zeta\,\mathds{1}_{N-1}\mathds{1}_{N-1}^{\top}
\end{bmatrix},\nonumber\\
\label{eq:Tqexpl}
\end{align}
where for brevity we introduced the bounding constant $\phi^{(q)}$.

Likewise, concerning the term involving $T_s$ we can write:
\beq
\left(\Delta\,\mathds{1}_N^{\top} + I_N\right) T_s
\preceq \phi^{(s)}\mu^2\,\mathds{1}_N\mathds{1}_N^{\top},
\label{eq:Tsexpl}
\eeq
where $\phi^{(s)}$ is a suitable constant.

Regarding the term involving $T_\delta$, by exploiting Table~\ref{tab:TransferMaTable} we have:
\begin{align}
&\Delta \mathds{1}_N^{\top}\, T_\delta=
\begin{bmatrix}
\bar{\Delta}&\bar{\Delta}\,\mathds{1}_{N-1}^{\top}
\\
\\
\widecheck{\Delta}\,\mathds{1}_{N-1}&\widecheck{\Delta}\,\mathds{1}_{N-1}\mathds{1}_{N-1}^{\top}
\end{bmatrix}
\begin{bmatrix}
[T_{\delta}]_{11}&[T_{\delta}]_{12}
\\
\\
[T_{\delta}]_{21}&[T_{\delta}]_{22}
\end{bmatrix}\nonumber\\
&
\preceq 
\begin{bmatrix}
0
&
8\,\bar{\Delta}\,\mathds{1}_{N-1}^{\top}\,(I_{N-1} + U)
\\
\\
0
&
8\,\widecheck{\Delta}\,\mathds{1}_{N-1} \mathds{1}_{N-1}^{\top}\,(I_{N-1} + U)
\end{bmatrix}
+\phi^{(\delta)}\mu^2\,\mathds{1}_N\mathds{1}_N^{\top}\nonumber\\
&\preceq
\begin{bmatrix}
0
&
16\,\bar{\Delta}\,\mathds{1}_{N-1}^{\top}
\\
\\
0
&
16\,\widecheck{\Delta}\,\mathds{1}_{N-1} \mathds{1}_{N-1}^{\top}
\end{bmatrix}
+\phi^{(\delta)}\mu^2\,\mathds{1}_N\mathds{1}_N^{\top},
\label{eq:Tdexpl}
\end{align}
where we used the bound:
\beq
\mathds{1}_{N-1}^{\top}(I_{N-1}+U)
\preceq
2\,\mathds{1}_{N-1}^{\top},
\eeq
and where $\phi^{(\delta)}$ is a suitable constant that upper bounds the terms of order $\mu^2$.

If we now introduce the maximal constant
\beq
\phi=\max\left\{\phi^{(s)},\phi^{(\delta)},\phi^{(q)}\right\},
\eeq 
and the rank-one perturbed version of $E_0$:
\beq
E=E_0 + 16\,\zeta^2\,\widecheck{\Delta}\,\mathds{1}_{N-1} \mathds{1}_{N-1}^{\top},
\label{eq:Erankonedef}
\eeq
by using \eqref{eq:Tqexpl}, \eqref{eq:Tsexpl}, and \eqref{eq:Tdexpl} in \eqref{eq:Tapp}, we end up with the following bound: \beq
T_{\rm{tmp}}\preceq 
\begin{bmatrix}
\tau&\tau_{12}\mathds{1}_{N-1}^{\top}
\\
\\
0& E
\end{bmatrix}
+
v_{\mu,\zeta}\,\mathds{1}_N^{\top},
\label{eq:Trepres}
\eeq
where the quantities $\tau$, $\tau_{12}$, and $v_{\mu,\zeta}$ are defined in \eqref{eq:tautau12firstappear} and \eqref{eq:varepsilondef00}, respectively. The claim of the theorem now follows from the definition of $T$ in \eqref{eq:Trepresfirstappear}.
\end{IEEEproof}

\section{Stability of $E$}
\label{subsec:stabE}
\begin{lemma}[Stability of $E$]
\label{lem:Estab}
Let
\beq
a_n \triangleq\frac{2 |\lambda_2| }{(1-|\lambda_2|)}\frac{1}{|\lambda_2|-|\lambda_n|^2},
\eeq
and
\beq
\gamma(A)\triangleq
\sum_{n=2}^B 
\frac{|\lambda_2|}{|\lambda_2|-|\lambda_n|^2}
\,
\left(\frac{a_n^{L_n+1}-1}{(a_n-1)^2}+\frac{L_n+1}{a_n-1}
\right).
\label{eq:gammaAdef}
\eeq
Then, the matrix $E$ defined in \eqref{eq:Edefintheorem} has spectral radius less than $1$ if, and only if:
\beq
\zeta<\frac{1}{16 \widecheck{\Delta} \gamma(A)}.
\label{eq:zetastab}
\eeq
\end{lemma}
\begin{IEEEproof}
We introduce the resolvent of matrix $E$:
\beq
\mathcal{R}_E(z)= (z I_{N-1} - E)^{-1},
\eeq
which is well-posed for $z$ distinct from the eigenvalues of $E$. 
The stability of $E$ will be proved if we show that:
\beq
\|\mathcal{R}_E(z)\|_1<\infty \quad \textnormal{for all $|z|\geq 1$}.
\label{eq:REzstab}
\eeq 
In order to prove \eqref{eq:REzstab}, we examine first the resolvent of the {\em unperturbed} matrix $E_0$ in Table~\ref{tab:Notation} --- see also \eqref{eq:E0def0}, namely,
\beq
\mathcal{R}_{E_0}(z)= (z I_{N-1} - E_0)^{-1}.
\eeq
Since $E_0$ is upper triangular and all its diagonal elements are positive values strictly less than one, we conclude that $E_0$ is stable, which further implies that the resolvent $\mathcal{R}_{E_0}(z)$ is bounded for $|z|\geq 1$. 

We continue by relating the resolvent of $E$ to the resolvent of the {\em unperturbed} matrix $E_0$. 
Exploiting the structure of $E$ in \eqref{eq:Erankonedef}, we see that $E$ is given by $E_0$ plus an additive rank-one perturbation. Since we have shown that in the range of interest $|z|\geq 1$ the matrix $z I_{N-1} - E_0$ is invertible, we can apply the Sherman-Morrison identity to $z I_{N-1} - E$, obtaining~\cite{Johnson-Horn}: 
\begin{align}
\mathcal{R}_E(z)\!=\!\mathcal{R}_{E_0}(z)  
\!+\! 16\,\zeta^2\widecheck{\Delta}\,
\frac{\mathcal{R}_{E_0}(z)\mathds{1}_{N-1}\mathds{1}_{N-1}^{\top}
\mathcal{R}_{E_0}(z)}{1-16\,\zeta^2\,\widecheck{\Delta}\,\mathds{1}_{N-1}^{\top}\mathcal{R}_{E_0}(z)\mathds{1}_{N-1}},
\nonumber\\
\label{eq:E0resolvent0}
\end{align}
where the identity holds if, and only if, the denominator on the RHS of \eqref{eq:E0resolvent0} is not zero. 
In particular, if the denominator is zero $z I_{N-1} - E$ is not invertible.
Therefore, to prove \eqref{eq:REzstab} we must examine the behavior of the complex scalar function:
\beq
g_E(z)\triangleq 
\mathds{1}_{N-1}^{\top}\mathcal{R}_{E_0}(z)\mathds{1}_{N-1},
\label{eq:gzdenom}
\eeq
over the entire range $|z|\geq 1$.
To this end, it is critical to characterize the resolvent of $E_0$.
Exploiting \eqref{eq:Lambdablocks}, \eqref{eq:Ublocks}, and \eqref{eq:Jordanrep}, we can represent $E_0$ as:
\beq
E_0=
{\sf diag}\left(E_2,E_3,\ldots,E_B
\right),
\eeq
where, for $n=2,3,\ldots,B$, we introduced the $L_n\times L_n$ matrices:
\beq
E_n=
\underbrace{
\left(
(1-\zeta) + \zeta\,\frac{|\lambda_n|^2}{|\lambda_2|} 
\right)}_{r_n}\,I_{L_n}
+
\underbrace{\frac{2\zeta}{1-|\lambda_2|}}_{u_n}\,U_{L_n}.
\eeq
By computing the inverse of the block-diagonal matrix $zI_{N-1} - E_0$ as the block-diagonal matrix of the individual inverse matrix-blocks, we have that:
\beq
\mathcal{R}_{E_0}(z)={\sf diag}\Big(
\mathcal{R}_{E_2}(z), \mathcal{R}_{E_3}(z),\ldots,\mathcal{R}_{E_B}(z)
\Big).
\label{eq:Re0diag}
\eeq
On the other hand, for Jordan-type matrices like $zI_{N-1} - E_n$, the inverse is known to be in the form:
\begin{align}
\mathcal{R}_{E_n}(z)=\frac{1}{z - r_n}
\begin{bmatrix}
    1 & \displaystyle{\frac{u_n}{z-r_n}} & \ldots &\!\!\!\left(\displaystyle{\frac{u_n}{z-r_n}}\right)^{L_n-1}\\
    0 & 1 & \ldots & \!\!\!\left(\displaystyle{\frac{u_n}{z-r_n}}\right)^{L_n-2} \\ 
    0& 0 & \ddots &  \!\!\!\!\!\!\!\!\!\!\!\!\!\!\!\!\!\!\!\!\!\!\!\vdots\\
    0& 0 & \ldots& \!\!\!\!\!\!\!\!\!\!\!\!\!\!\!\!\!\!\!\!\!\!\!1
    \end{bmatrix}.\nonumber\\
\label{eq:zIminusE0expand}
\end{align}
Using \eqref{eq:Re0diag} and \eqref{eq:zIminusE0expand} in \eqref{eq:gzdenom}, we obtain:
\beq
g_E(z)=\frac{1}{z-r_n}\,\sum_{n=2}^B \sum_{j=0}^{L_n-1} (L_n-j)\,\left(\displaystyle{\frac{u_n}{z-r_n}}\right)^j.
\label{eq:RE0zexpl}
\eeq
Applying the triangular inequality in \eqref{eq:RE0zexpl}, and noticing that for any two complex numbers $z_1$ and $z_2$, we have $|z_1-z_2|\geq |z_1|-|z_2|$, we can write the following inequality (in the range $|z|\geq 1$):
\beq
|g_E(z)|
\leq
\frac{1}{1-r_n}\sum_{n=2}^B \sum_{j=0}^{L_n-1} (L_n-j)\left(\displaystyle{\frac{u_n}{1-r_n}}\right)^j
=g_E(1).
\label{eq:boundResE0}
\eeq
Equation \eqref{eq:boundResE0} implies that a sufficient condition for the stability of $E$ is:
\beq
16\,\zeta^2 \,\widecheck{\Delta} g_E(1)<1.
\label{eq:true}
\eeq
We now show this is also a necessary condition by reductio ad absurdum. 
Assume that \eqref{eq:true} is violated, namely that (we rule out the equality since it obviously correspond to instability):
\beq
16 \zeta^2 \widecheck{\Delta} g_E(1)>1.
\label{eq:absurd}
\eeq
Were \eqref{eq:absurd} true, there would certainly exist one value $z^\star\in\mathbb{R}$, with $z^\star>1$, such that the denominator on the RHS of \eqref{eq:E0resolvent0} is equal to zero. This is because the function $g_E(z)$ is analytic over the domain $|z|\geq 1$ and, in particular, it is continuous on the real axis and vanishes as $z\rightarrow \infty$. 
This implies that \eqref{eq:true} is a necessary and sufficient condition for the stability of $E$. 

Let us now recast \eqref{eq:true} in a more explicit form. First of all, the inner summation can be computed by standard results on geometric series, yielding:
\begin{align}
&\sum_{j=0}^{L-1} (L-j) a^j=
\frac{a^{L+1}-1}{(a-1)^2}+\frac{L+1}{a-1}.
\end{align}
Let now
\beq
a_n = \displaystyle{\frac{u_n}{1-r_n}}=\frac{2 |\lambda_2| }{(1-|\lambda_2|)}\frac{1}{|\lambda_2|-|\lambda_n|^2},
\eeq
we have that
\beq
g_E(1)=\frac{\gamma(A)}{\zeta},
\label{eq:g1eq}
\eeq
where $\gamma(A)$ is defined in \eqref{eq:gammaAdef}.
In view of \eqref{eq:E0resolvent0}, \eqref{eq:gzdenom}, and \eqref{eq:true}, Eq. \eqref{eq:g1eq} implies that the matrix $E$ is stable if, and only if, Eq. \eqref{eq:zetastab} is verified.
\end{IEEEproof}

\section{Stability of $T$}
\label{subsec:stabT}
\begin{lemma}[Stability of $T$]
\label{lem:Tstab}
Let
\beq
\zeta<\frac{1}{16 \widecheck{\Delta} \gamma(A)}, \quad \mu<\frac{2}{\zeta\,(\eta + \nu)},
\label{eq:stabcondinLemma}
\eeq
and let $\mu^{\star}$ be the positive root of the equation:
\beq
\mu^2
\left(
1+\frac{\sigma^2_{12}}{\nu^2}
\right)
\,
\frac{\gamma(A)}{1-16\,\zeta\,\widecheck{\Delta}\,\gamma(A)}
+\mu\,\zeta\frac{1+16 \bar{\Delta}}{\nu} - \frac{1}{\phi}=0.
\label{eq:mu0eq2}
\eeq
Then, the matrix $T$ in \eqref{eq:Trepresfirstappear} has spectral radius less than $1$ if, and only if, $\mu<\mu^{\star}$.
\end{lemma}
\begin{IEEEproof}
The matrix $T_0$ in \eqref{eq:Trepresfirstappear} is stable since $\tau<1$ in view of the second inequality in \eqref{eq:stabcondinLemma}, and $E$ is stable in view of the first inequality in \eqref{eq:stabcondinLemma} and Lemma~\ref{lem:Estab}.
Then, the eigenvalues of $T_0$ lie all strictly within the unit disc, and, hence, the resolvent $\mathcal{R}_{T_0}(z)$ exists. 
Accordingly, considering the resolvent of matrix $T$:
\beq
\mathcal{R}_T(z)=(z I_N - T)^{-1}.
\eeq
and exploiting the structure of $T$ in \eqref{eq:Trepresfirstappear} (i.e., $T_0$ plus an additive rank-one perturbation), from the Sherman-Morrison identity we have~\cite{Johnson-Horn}:
\beq
\mathcal{R}_T(z)=\mathcal{R}_{T_0}(z)
+ \frac{\mathcal{R}_{T_0}(z)\,v_{\mu,\zeta}\,\mathds{1}_N^{\top}\,\mathcal{R}_{T_0}(z)}
{1 - \mathds{1}_N^{\top}\,\mathcal{R}_{T_0}(z)\,v_{\mu,\zeta}},
\label{eq:matinvlemmapply}
\eeq
where the formula is valid if, and only if, the denominator on the RHS of \eqref{eq:matinvlemmapply} is not zero. 
Moreover, if the denominator is zero, then $(z I_N - T)$ is not invertible.
Therefore, the stability of $T$ will be proved if we show that the denominator on the RHS of \eqref{eq:matinvlemmapply} is not zero over the range $|z|\geq 1$. 
To this end, we will now examine the resolvent of the unperturbed matrix $T_0$. 

From \eqref{eq:Trepresfirstappear} we see that $T_0$ is block upper-triangular, which implies that we can compute the inverse $\mathcal{R}_{T_0}(z)=(z I_N - T_0)^{-1}$ as:
\beq
\mathcal{R}_{T_0}(z)=
\begin{bmatrix}
\displaystyle{\frac{1}{z-\tau}}&\displaystyle{\frac{\tau_{12}\mathds{1}^{\top}_{N-1}\mathcal{R}_E(z)}{z-\tau}}\\
\\
0&\mathcal{R}_{E}(z)
\end{bmatrix},
\label{eq:zIminusT}
\eeq
and, exploiting the definition of $v_{\mu,\zeta}$ in \eqref{eq:varepsilondef00} we get:
\begin{align}
&g_T(z)\triangleq\mathds{1}_N^{\top}\,\mathcal{R}_{T_0}(z)\,v_{\mu,\zeta}=\phi \mu^2
\times\Bigg(
\frac{\zeta^2}{z-\tau}
\nonumber\\
&
+
\frac{\zeta\tau_{12}\mathds{1}_{N-1}^{\top}\mathcal{R}_E(z)\mathds{1}_{N-1}}{z-\tau}
+\zeta\mathds{1}_{N-1}^{\top}\mathcal{R}_E(z)\mathds{1}_{N-1}
\Bigg).
\label{eq:gTzdef}
\end{align}
Using \eqref{eq:E0resolvent0} and \eqref{eq:gzdenom}, we have the following identity:
\begin{align}
\mathds{1}_{N-1}^{\top}\mathcal{R}_E(z)\mathds{1}_{N-1}&=
g_E(z)
+ 16\,\zeta^2\,\widecheck{\Delta}\,
\frac{g_E^2(z)}{1-16\,\zeta^2\,\widecheck{\Delta}\,g_E(z)}\nonumber\\
&=
\frac{g_E(z)}{1-16\,\zeta^2\,\widecheck{\Delta}\,g_E(z)},
\end{align}
which, applied in \eqref{eq:gTzdef}, yields:
\begin{align}
g_T(z)&=
\phi \mu^2
\Bigg\{
\frac{\zeta^2}{z-\tau}
\nonumber\\
&+\zeta\left(
\frac{\tau_{12}}{z-\tau}+1
\right)
\,
\frac{g_E(z)}{1-16\,\zeta^2\,\widecheck{\Delta}\,g_E(z)}
\Bigg\}.
\end{align}
Accordingly, by triangle inequality we have that:
\begin{align}
&|g_T(z)|
\leq
\phi \mu^2\nonumber\\
&\times\left\{
\frac{\zeta^2}{|z-\tau|}
+\zeta\left(
\frac{\tau_{12}}{|z-\tau|}+1
\right)
\,
\frac{|g_E(z)|}{|1-16\,\zeta^2\,\widecheck{\Delta}\,g_E(z)|}
\right\}\nonumber\\
&\leq
\phi\mu^2\left\{
\frac{\zeta^2}{|z|-\tau}
+\zeta\left(
\frac{\tau_{12}}{|z|-\tau}+1
\right)
\,
\frac{|g_E(z)|}{1-16\,\zeta^2\,\widecheck{\Delta}\,|g_E(z)|}
\right\}.\nonumber\\
\label{eq:chaineqTstab}
\end{align}
where we used the inequality $|z_1-z_2|\geq |z_1|-|z_2|$, and the fact that, since \eqref{eq:zetastab} is verified by hypothesis, the denominator of the last fraction in \eqref{eq:chaineqTstab} is positive in view of \eqref{eq:boundResE0} and \eqref{eq:g1eq}. 
On the other hand, we know from \eqref{eq:boundResE0} that $|g_E(z)|\leq g_E(1)$ in the range $|z|\geq 1$, which, applied in \eqref{eq:chaineqTstab}, allows us to write:
\begin{align}
&|g_T(z)|\leq g_T(1)\nonumber\\
&=
\phi\mu^2\left\{
\frac{\zeta^2}{1-\tau}
+\zeta\left(
\frac{\tau_{12}}{1-\tau}+1
\right)
\,
\frac{g_E(1)}{1-16\,\zeta^2\,\widecheck{\Delta}\,g_E(1)}
\right\}.\nonumber\\
\label{eq:gTz2}
\end{align}
Reasoning as done in the proof of Lemma~\ref{lem:Estab}, a necessary and sufficient condition for the stability of $T$ is:
\beq
g_T(1)<1. 
\label{eq:gT1ineq}
\eeq
To this aim, let us apply the definitions in \eqref{eq:tautau12firstappear} and \eqref{eq:g1eq} in \eqref{eq:gTz2}, to obtain:
\begin{align}
&g_T(1)=\phi\nonumber\\
&\times\left\{
\mu\,\zeta\frac{1+16 \bar{\Delta}}{\nu}
+\mu^2
\left(
1+\frac{\sigma^2_{12}}{\nu^2}
\right)
\,
\frac{\gamma(A)}{1-16\,\zeta\,\widecheck{\Delta}\,\gamma(A)}
\right\}.\nonumber\\
\label{eq:gT1finalform}
\end{align}
In view of \eqref{eq:gT1finalform}, inequality \eqref{eq:gT1ineq} is true if, and only if, the (positive) step-size $\mu$ is smaller than the positive root of the equation in \eqref{eq:mu0eq2}, and the proof of the lemma is complete.
\end{IEEEproof}

\subsection{Bounds on the Powers of $T$}
\label{sec: Gamma bound lemma proof}
The stability established in Lemma~\ref{lem:Tstab} allows to conclude that the matrix powers $T^i$ can be uniformly bounded w.r.t. to $i$. However, the bound would depend on the matrix $T$, and, in particular, would depend on the step-size $\mu$. Since we are interested in characterizing the small-$\mu$ behavior of the ACTC mean-square-deviation, it is essential to establish how such bound behaves as $\mu\rightarrow 0$. To this end, Lemma~\ref{lem:Tstab} alone does not provide enough information, and we need to resort to the powerful framework of Kreiss stability~\cite{Trefethen-Embree}. 

Preliminarily, it is necessary to introduce the concept of {\em Kreiss constant}. 
Given the resolvent $\mathcal{R}_X(z)$ associated with a matrix $X$, the Kreiss constant relative to $X$ is defined as follows~\cite{Trefethen-Embree}:
\beq
K_X\dfz\sup_{z\in\mathbb{C}:|z|>1} (|z|-1) \,\|\mathcal{R}_X(z)\|,
\label{eq:Kreissdef}
\eeq
and it is useful to bound (from above and from below) the norm of matrix powers as follows:
\beq
K_X\leq\sup_{i \geq 0} \|X^i\|\leq N e K_X,
\label{eq:KreissTheorem}
\eeq
where $e$ is Euler's number. 
In the next lemma we exploit the Kreiss constant to characterize the small-$\mu$ behavior of the powers of $T$.

\begin{lemma}[Bound on the Powers of $T$]
\label{lem:Kreiss}
Let
\beq
\zeta<\frac{1}{16 \widecheck{\Delta} \gamma(A)}, \quad \mu<\min\left\{\frac{2}{\zeta\,(\eta + \nu)},\,\mu^{\star}\right\},
\label{eq:stabcondinLemmaBis}
\eeq
where $\mu^{\star}$ is the positive root of the equation in \eqref{eq:mu0eq2}.
Then we have that:
\beq
\boxed{
\sup_{i \geq 0} \|T^i\|_1\leq K(\mu),
}
\label{eq:KreissKreiss}
\eeq
where $K(\mu)$ is a function of $\mu$, independent of $i$, with:
\beq
K(\mu)=O(1) \textnormal{ as $\mu\rightarrow 0$}.
\label{eq:conditionsonK}
\eeq
\end{lemma}

\begin{IEEEproof}
Let us evaluate a bound on the Kreiss constant associated with matrix $T$. 
Accordingly, we will examine the behavior of the function
\beq
f(z)=(|z|-1)\|\mathcal{R}_T(z)\|_1,
\label{eq:fdef}
\eeq
over the range $|z|\geq 1$.
We have seen in the proof of Lemma~\ref{lem:Tstab} --- see the argument following \eqref{eq:matinvlemmapply} --- that under \eqref{eq:stabcondinLemmaBis} it is legitimate to use the representation in \eqref{eq:matinvlemmapply}. 
Applying now the triangle inequality to \eqref{eq:matinvlemmapply} we have:
\begin{align}
&\|\mathcal{R}_T(z)\|_1\leq
\|\mathcal{R}_{T_0}(z)\|_1
+\frac{\|\mathcal{R}_{T_0}(z)\|^2_1\times\|v_{\mu,\zeta}\mathds{1}_N^{\top}\|_1}
{|1 - \mathds{1}_N^{\top}\,\mathcal{R}_{T_0}(z)\,v_{\mu,\zeta}|}\nonumber\\
&\leq
\|\mathcal{R}_{T_0}(z)\|_1
+O(\mu^2)\,\frac{\|\mathcal{R}_{T_0}(z)\|^2_1}
{1 - g_T(1)},
\label{eq:KreissResolvineq}
\end{align}
where the $O(\mu^2)$ term comes from the behavior of the perturbation vector $v_{\mu,\zeta}$ in \eqref{eq:varepsilondef00}, whereas the bound involving the term $g_T(1)$ comes from \eqref{eq:gTzdef} and \eqref{eq:gTz2}. 
Therefore, from \eqref{eq:fdef} and \eqref{eq:KreissResolvineq} we conclude that:
\beq
f(z)\leq 
(|z|-1) \|\mathcal{R}_{T_0}(z)\|_1
+
O(\mu^2)\,\frac{(|z|-1)\|\mathcal{R}_{T_0}(z)\|^2_1}
{1 - g_T(1)}.
\label{eq:fineqapp}
\eeq
Let us examine the behavior of $\|\mathcal{R}_{T_0}(z)\|_1$. 
Exploiting the structure in \eqref{eq:zIminusT} and applying of the triangle inequality, we can write:
\begin{align}
\|\mathcal{R}_{T_0}(z)\|_1
\!\leq\!
\max
\left\{
\frac{1}{|z-\tau|},
\frac{N \tau_{12}\|\mathcal{R}_E(z)\|_1}{|z-\tau|}
\!+\!
\|\mathcal{R}_E(z)\|_1
\right\}.
\nonumber\\
\label{eq:Kreissineq2}
\end{align}
First, we examine the resolvent of matrix $E$. 
Since by assumption Eq. \eqref{eq:zetastab} is verified, the spectral radius of $E$ is strictly less than $1$ in view of Lemma~\ref{lem:Estab}. 
This implies that all eigenvalues of $E$ lie strictly inside in the unit disc, which in turn guarantees the existence of a constant $C_E$ such that~\cite{Johnson-Horn}:
\beq
\sup_{z\in\mathbb{C}: |z|> 1} \|\mathcal{R}_E(z)\|_1=C_E<\infty.
\label{eq:KE1}
\eeq
Moreover, the stability of $E$ implies that all powers of $E$ are bounded, which, in view of the lower bound in \eqref{eq:KreissTheorem}, implies the existence of a finite Kreiss constant:\footnote{We use the symbol $K_E'$ in place of $K_E$ since we are working with the $\ell_1$-norm, while the definition of Kreiss constant uses the $\ell_2$-norm.}
\beq
\sup_{z\in\mathbb{C}: |z|> 1} (|z|-1) \|\mathcal{R}_E(z)\|_1=K_E'<\infty.
\label{eq:KE2}
\eeq
We remark that both constants $C_E$ and $K_E'$ are independent of $\mu$, since so is matrix $E$.

Let us focus on the second term in \eqref{eq:fineqapp}. Using \eqref{eq:Kreissineq2}, we can write:
\begin{align}
&(|z|-1)\|\mathcal{R}_{T_0}(z)\|^2_1\leq(|z|-1)\nonumber\\
&\times\max
\left\{
\frac{1}{|z-\tau|^2},
\left(
\frac{N \tau_{12}\|\mathcal{R}_E(z)\|_1}{|z-\tau|}
+
\|\mathcal{R}_E(z)\|_1
\right)^2
\right\}.\nonumber\\
\label{eq:veryintermediateineq}
\end{align}
Let us examine the first argument of the maximum in \eqref{eq:veryintermediateineq}. 
Now, the known inequality $|z-\tau|\geq | \,|z| - \tau\,|$, turns into $|z-\tau|\geq (|z| - \tau)$ since $|z|\geq 1$ over the considered range and $\tau<1$ in view of \eqref{eq:stabcondinLemma}. 
Therefore we can write: 
\beq
\frac{|z|-1}{|z-\tau|^2}\leq 
\underbrace{\frac{|z|-1}{|z|-\tau}}_{\textnormal{$\leq 1 $ since $\tau<1$}}
\times
\underbrace{\frac{1}{|z|-\tau}}_{
\substack{\leq 1/(1-\tau) \\ \textnormal{since $|z|\geq 1$}}}
\leq \frac{1}{1-\tau}=\frac{1}{\mu\,\zeta\,\nu},
\label{eq:prefinalKreissTinequality1}
\eeq
where the equality comes from the definition of $\tau$ in \eqref{eq:tautau12firstappear}. 

Let us switch to the analysis of the second term in \eqref{eq:veryintermediateineq}, which, by expanding the square, yields:
\begin{align}
&N^2 \tau_{12}^2\,\underbrace{\frac{|z|-1}{|z-\tau|^2}}_{\substack{\leq 1/(\mu\zeta\nu)\\ \textnormal{see \eqref{eq:prefinalKreissTinequality1}}}}\,
\underbrace{\|\mathcal{R}_E(z)\|_1^2}_{\textnormal{$\leq C_E^2$, see \eqref{eq:KE1}}}
+
\underbrace{(|z|-1)\,\|\mathcal{R}_E(z)\|_1^2}_{\textnormal{$\leq C_E K_E'$, see \eqref{eq:KE1} and \eqref{eq:KE2}}}
\nonumber\\
&+
2 N \tau_{12}\,
\underbrace{
\frac{|z|-1}{|z-\tau|}}_{\leq 1}
\, \underbrace{\|\mathcal{R}_E(z)\|_1^2}_{\leq C_E^2}\nonumber\\
&\leq \frac{N^2 \tau_{12}^2 C_E^2}{\mu\,\zeta\,\nu} + C_E K_E' + 2 N \tau_{12}C_E^2=O(1/\mu).
\label{eq:prefinalKreissTinequality2}
\end{align}
Using \eqref{eq:prefinalKreissTinequality1} and \eqref{eq:prefinalKreissTinequality2} in \eqref{eq:veryintermediateineq}, we conclude that:
\beq
\sup_{z\in\mathbb{C}:|z|>1} (|z|-1) \,\|\mathcal{R}_{T_0}(z)\|_1^2 = O(1/\mu).
\label{eq:KreissTpiece2}
\eeq
Reasoning along the same lines we can show that:
\beq
\sup_{z\in\mathbb{C}:|z|>1} (|z|-1) \,\|\mathcal{R}_{T_0}(z)\|_1 = O(1).
\label{eq:KreissTpiece1}
\eeq
Applying now \eqref{eq:KreissTpiece2} and \eqref{eq:KreissTpiece1} in \eqref{eq:fineqapp}, we get:
\beq
f(z)=O(1)+\frac{O(\mu)}{1-g_T(1)},
\label{eq:prefinalKreiss}
\eeq
which implies, in view of \eqref{eq:Kreissdef}, the existence of a function $K(\mu)$ such that, under assumption \eqref{eq:stabcondinLemmaBis}, we are allowed to write:
\beq
\sup_{i\geq 0} \|T^i\|_1
\leq
K(\mu)=O(1)+\frac{O(\mu)}{1-g_T(1)}.
\label{eq:prefinalKreiss}
\eeq
From the properties of $g_T(1)$ examined in Lemma~\ref{lem:Tstab}, we know that $g_T(1)<1$ in the range of $\mu$ permitted by \eqref{eq:stabcondinLemmaBis}, and $g_T(1)\rightarrow 0$ as $\mu\rightarrow 0$, which implies the claim of the lemma \eqref{eq:conditionsonK}.
\end{IEEEproof}

\section{Proof of Theorem~\ref{th:MSEstab}}
\label{app:proofofTheorem2}
\begin{IEEEproof}
Developing the recursion in \eqref{eq:systemRecursion} we have:
\beq
\mathscr{P}[\widehat{\bm{q}}_i]
\preceq
T^i \mathscr{P}[\widehat{\bm{q}}_0] + (I_N - T)^{-1}x,
\label{eq:developRec1}
\eeq
and by application of the triangle inequality:
\begin{align}
\E\|\widehat{\bm{q}}_i\|^2\!=\!
\|\mathscr{P}[\widehat{\bm{q}}_i]\|_1
\leq
\|T\|_1^i \,\|\mathscr{P}[\widehat{\bm{q}}_0]\|_1 \!+\! 
\|(I_N-T)^{-1}\|_1\,x,\nonumber\\
\label{eq:staticbound00}
\end{align}
where the first equality comes from the definition of average energy operator in \eqref{eq:avenopdef}.
In view of Lemma~\ref{lem:Tstab}, under the assumptions of the theorem the matrix $T$ has spectral radius strictly less than $1$, which, in view of \eqref{eq:staticbound00}, implies:
\beq
\limsup_{i\rightarrow\infty} \E \|\widehat{\bm{q}}_i\|^2\leq \|(I_N-T)^{-1}\|_1\,x <\infty.
\label{eq:limsupqhat}
\eeq
On the other hand, from the network coordinate transformation in \eqref{eq:hatq}, we have $\widetilde{\bm{q}}_i=\mathcal{V}^{-1} \widehat{\bm{q}}_i$, which, in view of \eqref{eq:limsupqhat}, implies:
\beq
\limsup_{i\rightarrow\infty} \E \|\widetilde{\bm{q}}_{k,i}\|^2<\infty,
\label{eq:limsupqtilde}
\eeq
where we used the fact that the squared norm of the {\em extended} vector $\widetilde{\bm{q}}_{i}$ is the sum of the norms of the $N$ individual vectors $\|\widetilde{\bm{q}}_{k,i}\|^2$.
The claim in \eqref{eq:merestability} follows from the fact that $\widetilde{\bm{w}}_{k,i}$ is a convex combination of $\{\widetilde{\bm{q}}_{\ell,i}\}_{\ell\in\mathcal{N}_k}$ --- see \eqref{centeredACTC}.

We move on to prove \eqref{eq:orderofstability}, for which we need to examine the small-$\mu$ behavior of $(I_N - T)^{-1}$ in \eqref{eq:developRec1}. To this end, let us specialize \eqref{eq:matinvlemmapply} to the case $z=1$:
\begin{align}
(I_N-T)^{-1}&=\mathcal{R}_T(1)=
(I_N-T_0)^{-1}
\nonumber\\&+ 
\frac{(I_N-T_0)^{-1}\,v_{\mu,\zeta}\,\mathds{1}_N^{\top}\,(I_N-T_0)^{-1}}
{1 - \mathds{1}_N^{\top}\,(I_N-T_0)^{-1}\,v_{\mu,\zeta}}.
\label{eq:matinvlemmapplyapp}
\end{align}
On the other hand, specializing \eqref{eq:zIminusT} to the case $z=1$, it is immediate to see that:
\beq
    (I_N-T_0)^{-1} \preceq
    \begin{bmatrix}
       O(1/\mu) & O(1/\mu)\,\mathds{1}_{N-1}^{\top}  
       \\
       \\
       0 & O(1)\,\mathds{1}_{N-1}\,\mathds{1}_{N-1}^{\top}
    \end{bmatrix}.
\label{eq:IT0invapp1}
\eeq
Therefore we can write:
\begin{align}
&(I_N-T_0)^{-1}\,v_{\mu,\zeta}
\preceq
\begin{bmatrix}
O(\mu) 
\\
\\
O(\mu^2)\,\mathds{1}_{N-1}
\end{bmatrix},
\label{eq:IT0invapp21}
\\
\nonumber\\
&\mathds{1}^{\top}\,(I_N-T_0)^{-1}
\preceq
\begin{bmatrix}
O(1/\mu)& O(1/\mu)\mathds{1}_{N-1}^{\top}
\end{bmatrix},
\label{eq:IT0invapp22}
\end{align}
which further implies:
\begin{align}
(I_N-T_0)^{-1}\,&v_{\mu,\zeta}\,\mathds{1}_N^{\top}\,(I_N - T_0)^{-1}
\preceq
\nonumber\\
&\begin{bmatrix}
O(1)&O(1)\,\mathds{1}_{N-1}^{\top}
\\
\\
O(\mu)\,\mathds{1}_{N-1}&O(\mu)\,\mathds{1}_{N-1}\,\mathds{1}_{N-1}^{\top}
\end{bmatrix}.
\label{eq:IT0invapp3}
\end{align}
Combining now \eqref{eq:IT0invapp1} and \eqref{eq:IT0invapp3} we obtain:
\beq
    (I_N-T)^{-1} = 
    \begin{bmatrix}
       O(1/\mu) & O(1/\mu)\,\mathds{1}_{N-1}^{\top}  
       \\
       \\
       O(\mu)\,\mathds{1}_{N-1} & O(1)\,\mathds{1}_{N-1}\,\mathds{1}_{N-1}^{\top}
    \end{bmatrix}.
\label{eq:IT0invapp4}
\eeq
On the other hand, examining the entries of vector $x$ in \eqref{eq:Tandx} with the help of Table~\ref{tab:TransferMaTable} we readily see that:
\beq
x \preceq O(\mu^2)\,\mathds{1}_N.
\label{eq:xorderapp}
\eeq
Combining \eqref{eq:developRec1} with \eqref{eq:xorderapp} we get:
\beq
\|(I_N-T)^{-1}\|_1\,x=O(\mu),
\label{eq:norm1IminusTOmu}
\eeq
which completes the proof.
\end{IEEEproof}

\section{Proof of Theorem~\ref{th:transient}}
\label{app:Theorem3}
In the following, we make repeated use of the following known equality, holding for any two nonzero scalars with $a\neq b$:  
\begin{align}
\sum_{j=0}^{i-1} a^j b^{i-1-j}&=b^{i-1}\sum_{j=0}^{i-1} (a/b)^j=b^{i-1} \frac{1-(a/b)^i}{1-(a/b)}\nonumber\\
&=\frac{b^i - a^i}{b-a}.
\label{eq:fundgeomseries}
\end{align}

\begin{IEEEproof}
In view of assumption \eqref{eq:zetastaboundtheorem3}, we can use Lemma~\ref{lem:Kreiss} in \eqref{eq:staticbound00} to conclude that, for all $i\geq 1$, and for sufficiently small $\mu$:
\beq
\|\mathscr{P}[\widehat{\bm{q}}_i]\|_1
\leq
K(\mu)\|\mathscr{P}[\widehat{\bm{q}}_0]\|_1 + 
\|(I_N-T)^{-1}\|_1\,x=O(1),
\label{eq:staticbound}
\eeq
where $K(\mu)$ is $O(1)$ in view of \eqref{eq:conditionsonK} while the order of magnitude of $\|(I_N-T)^{-1}\,x\|_1$ is $O(\mu)$ in view of \eqref{eq:norm1IminusTOmu}.

Let us develop the recursion in \eqref{eq:systemRecursion} by separating the role of the unperturbed matrix $T_0$ and the rank-one perturbation $v_{\mu,\zeta}\,\mathds{1}_N^{\top}$ in \eqref{eq:Trepresfirstappear}, yielding:
\begin{align}
\mathscr{P}[\widehat{\bm{q}}_{i}]
&\preceq 
T_0 \, \mathscr{P}[\widehat{\bm{q}}_{i-1}] +
v_{\mu,\zeta}\,\mathds{1}_N^{\top} \mathscr{P}[\widehat{\bm{q}}_{i-1}]
+ x=\nonumber\\
&\preceq
T_0 \, \mathscr{P}[\widehat{\bm{q}}_{i-1}] +
O(\mu^2)\,\mathds{1}_N,
\label{eq:Pqrecursapp1}
\end{align}
where the second term on the RHS is $O(\mu^2)$ because so are $v_{\mu,\zeta}$ and $x$, while 
\beq
\mathds{1}_N^{\top} \mathscr{P}[\widehat{\bm{q}}_{i-1}]=\|\mathscr{P}[\widehat{\bm{q}}_{i-1}]\|_1,
\eeq
which is $O(1)$ in view of \eqref{eq:staticbound}. 
Developing the recursion in \eqref{eq:Pqrecursapp1} we have:
\begin{align}
\mathscr{P}[\widehat{\bm{q}}_{i}]
&\preceq
T_0^i \mathscr{P}[\widehat{\bm{q}}_0] + O(\mu^2)\,(I_N-T_0)^{-1}\,\mathds{1}_N\nonumber\\
&\preceq T_0^i \mathscr{P}[\widehat{\bm{q}}_0] + 
\begin{bmatrix}
O(\mu)
\\
\\
O(\mu^2)\,\mathds{1}_{N-1}
\end{bmatrix},
\label{eq:developRec2}
\end{align}
where the estimate of the last term comes from \eqref{eq:IT0invapp1}. 

Let us now evaluate the $i$-th power of $T_0$. 
Since $T_0$ is block upper-triangular, its $i$-th power admits the representation~\cite{ChenSayedTIT2015part1}:
\begin{align}
T_0^i = 
\begin{bmatrix}
\tau^{i} & \tau_{12}\mathds{1}_{N-1}^{\top}\, (\tau I_{N-1} - E)^{-1}(\tau^{i} I_{N-1} - E^i) \\
\\
0 & E^i
\end{bmatrix},\nonumber\\
\label{eq:T0Powers}
\end{align}
where we note that in the small-$\mu$ regime the matrix $(\tau I - E)^{-1}$ is certainly well-defined and has nonnegative entries, since as $\mu\rightarrow 0$ we have $\tau=1-\mu\,\zeta\,\nu > \rho(E)$ (and since $E$ has nonnegative entries).
Therefore, in the small-$\mu$ regime we can write:
\beq
\tau_{12}\mathds{1}_{N-1}^{\top} (\tau I - E)^{-1}(\tau^{i}I - E^i)\preceq
\tau_{12}\mathds{1}_{N-1}^{\top} (\tau I - E)^{-1}\,\tau^i .
\label{eq:term12bound}
\eeq
Considering the evolution of the first component $\E\|\bar{\bm{q}}_i\|^2$ of $\mathscr{P}[\widecheck{\bm{q}}_i]$ as dictated by \eqref{eq:developRec2}, from \eqref{eq:T0Powers} and \eqref{eq:term12bound} we get:\footnote{
The quantity $\mathds{1}_{N-1}(\tau I - E)^{-1}$ is $O(1)$ as $\mu\rightarrow 0$. This can be seen, e.g., by noticing that $(\tau I - E)^{-1}\rightarrow \mathcal{R}_E(1)$ as $\mu\rightarrow 0$, and we have already shown that $\mathcal{R}_E(1)$ is bounded.
}
\beq
\E\|\bar{\bm{q}}_i\|^2\leq O(1)\, \tau^i + O(\mu).
\label{eq:prefinalboundbar}
\eeq
Likewise, exploiting \eqref{eq:developRec2}, \eqref{eq:T0Powers}, and \eqref{eq:term12bound} to get the evolution relative to the network-error component we can write:
\beq
\mathscr{P}[\widecheck{\bm{q}}_i]
\preceq
E^i \mathscr{P}[\widecheck{\bm{q}}_0] + O(\mu^2)\mathds{1}_{N-1},
\eeq
which allows us to write~\cite{Johnson-Horn}:
\beq
\E\|\widecheck{\bm{q}}_i\|^2\leq O(1)\,\rho_{\rm{net}}^i + O(\mu^2),
\label{eq:prefinalboundcheck}
\eeq
where by definition $\rho_{\rm{net}}=\rho(E)+\epsilon<1$.

We now exploit the exponential bounds in \eqref{eq:prefinalboundbar} and \eqref{eq:prefinalboundcheck} to get a new recursion that would allow us to obtain refined estimates of both the transient and the steady-state errors. 
To this end, we revisit the norm on the LHS of \eqref{eq:lemma3baralmost}, and manage to obtain a better bound by computing explicitly the norm and then applying the Cauchy-Schwarz inequality, yielding:  
\begin{align}
&\|(I_M - \mu\zeta\bm{G}_{11,i-1})\bar{\bm{q}}_{i-1} - \mu\zeta\bm{G}_{12,i-1}\widecheck{\bm{q}}_{i-1}\|^2
\nonumber\\
&=
\|(I_M - \mu\zeta\bm{G}_{11,i-1})\bar{\bm{q}}_{i-1}\|^2
+
\mu^2 \zeta^2\|\bm{G}_{12,i-1}\widecheck{\bm{q}}_{i-1}\|^2\nonumber\\
&-
2\mu\zeta\big[(I_M - \mu \zeta \bm{G}_{11,i-1})\bar{\bm{q}}_{i-1}\big]^{\top} \bm{G}_{12,i-1}\widecheck{\bm{q}}_{i-1}
\nonumber\\
&\leq
(1 - \mu\zeta\nu)^2\,\|\bar{\bm{q}}_{i-1}\|^2
+
\mu^2 \zeta^2\,\sigma^2_{12} \,\|\widecheck{\bm{q}}_{i-1}\|^2\nonumber\\
&+
2\mu\zeta
\|(I_M - \mu\zeta \bm{G}_{11,i-1})\bar{\bm{q}}_{i-1}\|
\times\|\bm{G}_{12,i-1}\widecheck{\bm{q}}_{i-1}\|\nonumber\\
&\leq
(1-\mu\zeta\nu)^2\,\|\bar{\bm{q}}_{i-1}\|^2
+
\mu^2 \zeta^2\,\sigma^2_{12} \,\|\widecheck{\bm{q}}_{i-1}\|^2\nonumber\\
&+
2\mu\zeta (1-\mu\zeta\nu) \sigma_{12}
\|\bar{\bm{q}}_{i-1}\|\times 
\|\widecheck{\bm{q}}_{i-1}\|.
\end{align}
Taking expectations and applying the Cauchy-Schwartz inequality for random variables, we obtain:
\begin{align}
&\E\|(I_M-\mu\zeta\bm{G}_{11,i-1})\bar{\bm{q}}_{i-1}-\mu\zeta\bm{G}_{12,i-1}\widecheck{\bm{q}}_{i-1}\|^2
\nonumber\\
&\leq
(1-\mu\zeta\nu)^2\,\E\|\bar{\bm{q}}_{i-1}\|^2
+
\mu^2 \zeta^2\,\sigma^2_{12} \,\E\|\widecheck{\bm{q}}_{i-1}\|^2\nonumber\\
&+
2\mu\zeta (1-\mu\zeta\nu) \sigma_{12}
\sqrt{\E\|\bar{\bm{q}}_{i-1}\|^2
\times\E\|\widecheck{\bm{q}}_{i-1}\|^2}
\nonumber\\
&
\leq
(1-\mu\zeta\nu)^2\,\E\|\bar{\bm{q}}_{i-1}\|^2
+
\mu^2 \zeta^2\,\sigma^2_{12} \,\E\|\widecheck{\bm{q}}_{i-1}\|^2\nonumber\\
&+
O(\mu)\,
\sqrt{
\Big(O(1)\,\tau^i+O(\mu)\Big)
\Big(O(1)\,\rho_{\rm{net}}^{i} + O(\mu^2)\Big)}
\nonumber\\
&
\leq
(1-\mu\zeta\nu)^2\,\E\|\bar{\bm{q}}_{i-1}\|^2
+
\mu^2 \zeta^2\,\sigma^2_{12} \,\E\|\widecheck{\bm{q}}_{i-1}\|^2\nonumber\\
&+
O(\mu)\,
\left(O(1)\,\tau^{i/2}+O(\sqrt{\mu})\right)
\left(O(1)\,\rho_{\rm{net}}^{i/2} + O(\mu)\right)
\nonumber\\
&
=
(1-\mu\zeta\nu)^2\,\E\|\bar{\bm{q}}_{i-1}\|^2
+
\mu^2 \zeta^2\,\sigma^2_{12} \,\E\|\widecheck{\bm{q}}_{i-1}\|^2\nonumber\\
&+
\underbrace{
O(\mu^2)\,\tau^{i/2} 
+
O(\mu)\,\rho_{\rm{net}}^{i/2} 
+ O(\mu^{5/2})}_{\chi_i},
\label{eq:ineqafterCSineq2}
\end{align}
where we applied \eqref{eq:prefinalboundbar} and \eqref{eq:prefinalboundcheck}, along with the inequality $\sqrt{a + b} \leq \sqrt{a} + \sqrt{b}$ for $a, b \geq 0$. 

Using \eqref{eq:ineqafterCSineq2}, we can rearrange the first row of the transfer matrix $T_q$ and of the vector $x_q$ appearing in Lemma~\ref{lem:qstaterecextended}, obtaining a new matrix $T_q'$ and a new vector $x_q'$ defined by:
\begin{align}
[T_q']_{11}=\tau^2,~~[T_q']_{12}=\mu^2 \zeta^2\,\sigma^2_{12}\,\mathds{1}_{N-1}^{\top},~~
\bar{x}_q'=\begin{bmatrix}
\bar{x}_q + \chi_i
\\
\\
\widecheck{x}_q
\end{bmatrix},\nonumber\\
\label{eq:Tandxprime}
\end{align}
which allows us to replace the matrix $T$ appearing in Theorem~\ref{Quantized state recursion} with a matrix $T'$ of the following form:
\beq
T'=
\underbrace{
\begin{bmatrix}
\tau^2&\tau_{12}'\,\mathds{1}_{N-1}^{\top}
\\
\\
0& E
\end{bmatrix}}_{T_0'}
+
O(\mu^2)\,\mathds{1}_N\,\mathds{1}_N^{\top},
\label{eq:Tprimerepres}
\eeq
where we defined:
\beq
\tau_{12}'=16\,\zeta^2\,\bar{\Delta},
\label{eq:tau12prime}
\eeq
with the $\mu^2$-term appearing in $[T_q']_{12}$ being conveniently embodied in the overall $O(\mu^2)$ rank-one perturbation.
Likewise, we construct a new driving vector $x'=x+\chi_i$ by replacing $x_q$ in \eqref{eq:Tandx} with $x_q'$ in \eqref{eq:Tandxprime}.
Replacing now $T$ and $x$ with $T'$ and $x'$ in the recursion \eqref{eq:systemRecursion}, we get:
\begin{align}
&\mathscr{P}[\widehat{\bm{q}}_i]\preceq
T_0' \,\mathscr{P}[\widehat{\bm{q}}_{i-1}] 
+O(\mu^2)\,\mathds{1}_{N}\,
\underbrace{\mathds{1}_{N}^{\top}\,\mathscr{P}[\widehat{\bm{q}}_{i-1}]}_{\E\|\bar{\bm{q}}_i\|^2+\E\|\widecheck{\bm{q}}_i\|^2} + x'
\nonumber\\
&\preceq
T_0' \,\mathscr{P}[\widehat{\bm{q}}_{i-1}] 
+
\underbrace{
O(\mu^2)\left(
O(1)\,\tau^i + O(1)\,\rho_{\rm{net}}^{i} + O(\mu)
\right)}_{\chi_i'}
\,\mathds{1}_{N}\nonumber\\
&+ 
\begin{bmatrix}
\bar{x} + \chi_i
\\
\\
\widecheck{x}
\end{bmatrix}
\label{eq:tildeRecurs2},
\end{align}
where in the last step we used \eqref{eq:prefinalboundbar} and \eqref{eq:prefinalboundcheck}.
Developing the recursion in \eqref{eq:tildeRecurs2} we get:
\begin{align}
\mathscr{P}[\widehat{\bm{q}}_i]&\preceq
(T_0')^{i} \,\mathscr{P}[\widehat{\bm{q}}_0] 
+
\sum_{j=0}^{i-1} \chi_{i-j}' \,(T_0')^{j} \mathds{1}_N
\nonumber\\
&+ 
\sum_{j=0}^{i-1} (T_0')^{j}\,
\begin{bmatrix}
\chi_{i-j}
\\
\\
0
\end{bmatrix}
+
(I_N - T_0')^{-1}
\begin{bmatrix}
\bar{x}
\\
\\
\widecheck{x}
\end{bmatrix}.
\label{eq:tildeRecurs3}
\end{align}
Applying now \eqref{eq:T0Powers} (with $\tau^2$ in place of $\tau$, and $\tau_{12}'$ in place of $\tau_{12}$), and reasoning as done to obtain the bound in \eqref{eq:term12bound}, we conclude that the first row of matrix $(T_0')^i$ can be upper bounded as:
\beq
(T_0')^i
\preceq
\begin{bmatrix}
O(1)\,\tau^{2 i}\,\mathds{1}_N^{\top}
\\
\\
\textnormal{not needed}
\end{bmatrix}.
\label{eq:T0primefirstrow}
\eeq
Moreover, we can evaluate $(I_N - T_0')^{-1}$ through \eqref{eq:zIminusT} applied with $\tau^2$ in place of $\tau$, and $\tau_{12}'$ in place of $\tau_{12}$, obtaining:
\begin{align}
(I_N - T_0')^{-1}=
\begin{bmatrix}
\displaystyle{\frac{1}{1-\tau^2}}&\displaystyle{\frac{\tau_{12}' \,\mathds{1}^{\top}_{N-1}\,(I_{N-1} - E)^{-1}}{1-\tau^2}}\\
\\
0&(I_{N-1} - E)^{-1}
\end{bmatrix}.\nonumber\\
\label{eq:zIminusT0prime}
\end{align}
Using \eqref{eq:T0primefirstrow} and \eqref{eq:zIminusT0prime} \eqref{eq:tildeRecurs3}, we can obtain an inequality recursion on the first entry of $\mathscr{P}[\widehat{\bm{q}}_i]$:
\begin{align}
\E\|\bar{\bm{q}}_i\|^2 &\leq
O(1)\,\tau^{2 i} \, \E\|\widehat{\bm{q}}_0\|^2\, 
+ O(1)\,\sum_{j=0}^{i-1}\tau^{2 j}\,\chi_{i-j}
\nonumber\\
&+ \frac{\bar{x}}{1-\tau^2} 
+\displaystyle{\frac{\tau_{12}' \,\mathds{1}^{\top}_{N-1}\,(I_{N-1} - E)^{-1}}{1-\tau^2}}\,\widecheck{x},
\label{eq:tildeRecurs4}
\end{align}
where we have ignored the term $\chi_i'$, since comparing this term against the term $\chi_i$ in \eqref{eq:ineqafterCSineq2}, we see that $\chi_i'$ is dominated by $\chi_i$ as $\mu\rightarrow 0$ and, hence, can be formally embodied into $\chi_i$ through the Big-O notation.
Applying the definition of $\chi_i$ in \eqref{eq:ineqafterCSineq2} we can write:
\begin{align}
&\sum_{j=0}^{i-1}\tau^{2 j}\,\chi_{i-j}=
O(\mu^2)\,
\sum_{j=0}^{i-1}\tau^{2 j}\tau^{(i-j)/2} \nonumber\\
&+
O(\mu)\,
\sum_{j=0}^{i-1}\tau^{2 j}
\rho_{\rm{net}}^{(i-j)/2} 
+ O(\mu^{5/2})\,\sum_{j=0}^{i-1}\tau^{2 j}\nonumber\\
&\leq
O(\mu^2)\frac{\tau^{i/2}}{\tau^{1/2}-\tau^2}
+
O(\mu)\,\frac{\tau^{2 i}}{\tau^2 - \rho_{\rm{net}}^{1/2}}
+
\frac{O(\mu^{5/2})}{1-\tau^2},
\label{eq:tildeRecurs5}
\end{align}
where in the last inequality we exploited the geometric summation in \eqref{eq:fundgeomseries}. 

Now we examine the small-$\mu$ behavior of the three terms appearing on the RHS of \eqref{eq:tildeRecurs5}. 
The first term is $O(\mu)\,\tau^{i/2}$, since we have that:
\begin{align}
\lim_{\mu\rightarrow 0} \frac{\mu}{\tau^{1/2}-\tau^2}=
\lim_{\mu\rightarrow 0} \frac{\mu}{(1-\mu\zeta\nu)^{1/2} - (1-\mu\zeta\nu)^2}=\frac{2}{3 \zeta \nu}.\nonumber\\
\end{align}
The second term on the RHS of \eqref{eq:tildeRecurs5} is $O(\mu)\,\tau^{2 i}$, since: 
\beq
\lim_{\mu\rightarrow 0} \tau^2 - \rho_{\rm{net}}^{1/2}=1-\rho_{\rm{net}}^{1/2}>0.
\eeq
Finally, the third term is $O(\mu^{3/2})$ since:
\beq
\frac{1}{1-\tau^2}=\frac{1}{\mu\,\zeta\,\nu}\,\frac{1}{2-\mu\,\zeta\,\nu}=\frac{1}{\mu\,\zeta\,\nu}\,
\left(\frac{1}{2} + O(\mu)\right).
\label{eq:oneminustausquare}
\eeq
As a result, we can use \eqref{eq:tildeRecurs5} in \eqref{eq:tildeRecurs4} and substitute the estimated orders of the aforementioned three terms to obtain:
\begin{align}
\E\|\bar{\bm{q}}_i\|^2 &\leq
O(1)\,\tau^{2i}\,+O(\mu)\,\tau^{i/2} + O(\mu^{3/2})
\nonumber\\
&+ \frac{\bar{x}}{1-\tau^2} 
+\displaystyle{\frac{\tau_{12}' \,\mathds{1}^{\top}_{N-1}\,(I_{N-1} - E)^{-1}}{1-\tau^2}}\,\widecheck{x}.
\label{eq:tildeRecurs6}
\end{align}
It remains to examine the small-$\mu$ behavior of the last two terms in \eqref{eq:tildeRecurs6}. 
First, we observe that the quantities $\bar{\Delta}$ and $\widecheck{\Delta}$ defined in Table~\ref{tab:Notation} and characterizing vector $x$ in \eqref{eq:Tandx} are proportional to the maximum compression factor in \eqref{eq:omegamax}, namely,
\beq
\bar{\Delta}\propto \Omega, \qquad\qquad \widecheck{\Delta}\propto \Omega,
\label{eq:DeltaPropToOmega}
\eeq
Then, exploiting Table~\ref{tab:TransferMaTable}, the definition of $x$ in \eqref{eq:Tandx}, and \eqref{eq:DeltaPropToOmega}, we obtain:
\begin{align}
\bar{x}&=\zeta^2 \bar{x}_s + \zeta^2\bar{\Delta} \Big(\bar{x}_s + 
\mathds{1}_{N-1}^{\top}(\widecheck{x}_s+\widecheck{x}_{\delta})\Big)\nonumber\\
&=
\mu^2\,\zeta^2 \,
\left(\sum_{k=1}^N \pi_k \alpha_k^2 \sigma_k^2 + \kappa_1\,\Omega\right),
\label{eq:barxconvenientexpression}
\end{align}
where $\kappa_1$ is a suitable constant (i.e., independent of $\mu$).
Using \eqref{eq:oneminustausquare}, we can therefore represent the first fraction on the RHS of \eqref{eq:tildeRecurs6} as:
\beq
\frac{\bar{x}}{1-\tau^2}=
\mu\,\zeta\,\left(
\frac{\sum_{k=1}^N \pi_k \alpha_k^2 \sigma_k^2}{2\nu}\,  + \kappa_1\,\Omega 
\right)
+O(\mu^2).
\eeq
Let us focus on the last term in \eqref{eq:tildeRecurs6}. 
Resorting again to \eqref{eq:DeltaPropToOmega}, from \eqref{eq:tau12prime} and \eqref{eq:Tandx} we get:
\beq
\tau_{12}'=\kappa_2\, \zeta^2 \, \Omega,~~~ 
\widecheck{x}=\left(\kappa_3 + \kappa_4 \, \Omega\right)\mu^2,
\label{eq:tau12checkxOmega}
\eeq  
for some constants $\kappa_2$, $\kappa_3$, and $\kappa_4$. 
Using now \eqref{eq:oneminustausquare} and \eqref{eq:tau12checkxOmega}, the last term in \eqref{eq:tildeRecurs6} can be represented as $\mu\,\zeta\,\kappa_5 \Omega (1+\Omega) + O(\mu^2)$, for a suitable constant $\kappa_5$. 
Joining this representation with \eqref{eq:barxconvenientexpression}, we can finally represent \eqref{eq:tildeRecurs6} as:
\begin{align}
\E\|\bar{\bm{q}}_i\|^2 &\leq
O(1)\,\tau^{2i}\,+O(\mu)\,\tau^{i/2} + O(\mu^{3/2})
\nonumber\\
&+ 
\mu\,\zeta \,\left(
\frac{\sum_{k=1}^N \pi_k \alpha_k^2 \sigma_k^2}{2\nu} + c_q \,\Omega\,(1+\Omega)
\right),
\label{eq:tildeRecurs7}
\end{align}
where $c_q$ is a constant independent of $\mu$.

It remains to characterize the behavior of the mean-square-deviation at an individual agent $k$. 
To this end, we evaluate the individual entry of the extended vector $\widetilde{\bm{q}}_i$ though \eqref{eq:individualvsnetwork}, which allows us to write:
\begin{align}
&\E\|\bm{q}_{k,i}\|^2 = 
\E\|\bar{\bm{q}}_{i}\|^2+\E\|\mathcal{T}_k\widecheck{\bm{q}}_{i}\|^2+2 \E[\bar{\bm{q}}_{i}^{\top} \mathcal{T}_k \widecheck{\bm{q}}_{i}] \nonumber\\
& \leq 
\E\|\bar{\bm{q}}_{i}\|^2+\E\|\mathcal{T}_k\widecheck{\bm{q}}_{i}\|^2+2 \big|\E[\bar{\bm{q}}_{i}^{\top} \mathcal{T}_k \widecheck{\bm{q}}_{i}]\big| \nonumber \\
&\leq \E\|\bar{\bm{q}}_{i}\|^2
+
\|\mathcal{T}_k\|^2\,\E\|\widecheck{\bm{q}}_{i}\|^2
\nonumber\\
&+ 2 \|\mathcal{T}_k\| 
\sqrt{\E\|\bar{\bm{q}}_{i}\|^2 \, \E\|\widecheck{\bm{q}}_{i}\|^2}, 
\label{eq:finalStepForRate}
\end{align}
where we resorted again to the Cauchy-Schwartz inequality for random variables. 
Now, from \eqref{eq:tildeRecurs7} we can write $\E\|\bar{\bm{q}}_{i}\|^2\leq O(1)\,\tau^{2 i} + O(\mu)$, which, used along with \eqref{eq:prefinalboundcheck}, yields:
\begin{align}
&\sqrt{\E\|\bar{\bm{q}}_{i}\|^2 \, \E\|\widecheck{\bm{q}}_{i}\|^2}\nonumber\\
&\leq
\sqrt{\Big(O(1)\tau^{2i} + O(\mu)\Big) \, \Big(O(1)\rho_{\rm{net}}^i + O(\mu^2)\Big) }
\nonumber\\
&\leq
\underbrace{O(1)\tau^{i} \rho_{\rm{net}}^{i/2}
+ O(\mu^{1/2})\rho_{\rm{net}}^{i/2}}_{O(1)\rho_{\rm{net}}^{i/2}}
+ \underbrace{O(\mu)\tau^i}_{\leq O(\mu)\tau^{i/2}}
+O(\mu^{3/2})
\label{eq:developCauchySchwartz}
\end{align}
Further noticing that from \eqref{eq:prefinalboundcheck} we can write $\E\|\widecheck{\bm{q}}_{i}\|^2 \leq O(1)\,\rho_{\rm{net}}^{i/2} + O(\mu^2)$, the claim of the theorem follows by using \eqref{eq:finalStepForRate} and \eqref{eq:developCauchySchwartz} in \eqref{eq:tildeRecurs7}.
\end{IEEEproof}



\begin{thebibliography}{10}
\providecommand{\url}[1]{#1}
\csname url@samestyle\endcsname
\providecommand{\newblock}{\relax}
\providecommand{\bibinfo}[2]{#2}
\providecommand{\BIBentrySTDinterwordspacing}{\spaceskip=0pt\relax}
\providecommand{\BIBentryALTinterwordstretchfactor}{4}
\providecommand{\BIBentryALTinterwordspacing}{\spaceskip=\fontdimen2\font plus
\BIBentryALTinterwordstretchfactor\fontdimen3\font minus
  \fontdimen4\font\relax}
\providecommand{\BIBforeignlanguage}[2]{{%
\expandafter\ifx\csname l@#1\endcsname\relax
\typeout{** WARNING: IEEEtran.bst: No hyphenation pattern has been}%
\typeout{** loaded for the language `#1'. Using the pattern for}%
\typeout{** the default language instead.}%
\else
\language=\csname l@#1\endcsname
\fi
#2}}
\providecommand{\BIBdecl}{\relax}
\BIBdecl

\bibitem{ProcIEEEspecialissue2020}
U.~A. Khan, W.~U. Bajwa, A.~Nedi\'{c}, M.~G. Rabbat, and A.~H. Sayed, {\em Editors}, ``Optimization for Data-Driven Learning and Control,'' \emph{Proceedings of the IEEE}, vol.~108, no.~11, pp. 1863--1868, Nov. 2020

\bibitem{TsitsiklisBertsekasAthansTAC1986}
J.~N. Tsitsiklis, D.~P. Bertsekas, and M.~Athans, ``Distributed asynchronous deterministic and stochastic gradient optimization algorithms,'' \emph{IEEE Trans. Autom. Control}, vol.~31, no.~9, pp. 803--812, Sep. 1986.

\bibitem{NedicBertsekasSIAM2001}
A.~Nedi\'{c} and D.~P. Bertsekas, ``Incremental subgradient methods for nondifferentiable optimization,'' \emph{SIAM J. Optim.}, vol.~12, no.~1, pp. 109--138, 2001.

\bibitem{NedicOzdaglar2010}
A.~Nedi\'{c} and A.~Ozdaglar, ``Cooperative distributed multi-agent optimization,'' in \emph{Convex Optimization in Signal Processing and Communications}, Y.~Eldar and D.~Palomar Eds.\hskip 1em plus 0.5em minus 0.4em\relax Cambridge University Press, 2010, pp. 340--386.

\bibitem{BoydFoundTrends}
S.~Boyd, N.~Parikh, E.~Chu, B.~Peleato, and J.~Eckstein, ``Distributed optimization and statistical learning via the alternating direction method of multipliers,'' \emph{Found. Trends Mach. Learn.}, vol.~3, no. 1, pp. 1--122, 2010.

\bibitem{NedicJSTSP2013}
S.~Lee and A.~Nedi\'{c}, ``Distributed random projection algorithm for convex optimization,'' \emph{IEEE J. Sel. Topics Signal Process.}, vol.~7, no.~2, pp. 221--229, Apr. 2013.

\bibitem{KhanTAC2016}
C.~Xi and U.~A. Khan, ``Distributed subgradient projection algorithm over directed graphs,'' \emph{IEEE Trans. Autom. Control}, vol.~62, no.~8, pp. 3986--3992, Oct. 2016.

\bibitem{KhanTAC2018}
C.~Xi, V.~S. Mai, R.~Xin, E.~Abed, and U.~A. Khan, ``Linear convergence in optimization over directed graphs with row-stochastic matrices,'' \emph{IEEE Trans. Autom. Control}, vol.~63, no.~10, pp. 3558--3565, Oct. 2018.

\bibitem{RabbatRibeiroCoopGraphSP2018}
M.~G. Rabbat and A.~Ribeiro, ``Multiagent distributed optimization,'' in \emph{Cooperative and Graph Signal Processing}, P.~Djuric and C.~Richard, Eds.\hskip 1em plus 0.5em minus 0.4em\relax Elsevier, 2018, pp. 147--167.

\bibitem{BajwaTSIPN}
M.~Nokleby and W.~U. Bajwa, ``Stochastic optimization from distributed streaming data in rate-limited networks,'' \emph{IEEE Trans. Signal Inf. Process. Netw.}, vol.~5, no.~1, pp. 152--167, Mar. 2019.

\bibitem{SayedTuChenZhaoTowficSPmag2013}
A.~H. Sayed, S.~Y. Tu, J.~Chen, X.~Zhao, and Z.~J. Towfic, ``Diffusion strategies for adaptation and learning over networks,'' \emph{IEEE Signal Process. Mag.}, vol.~30, no.~3, pp. 155--171, May 2013.

\bibitem{SayedProcIEEE2014}
A.~H. Sayed, ``Adaptive networks,'' \emph{Proceedings of the IEEE}, vol.~102, no.~4, pp. 460--497, Apr. 2014.

\bibitem{ChenSayedTIT2015part1}
J.~Chen and A.~H. Sayed, ``On the learning behavior of adaptive networks --- part I: Transient analysis,'' \emph{IEEE Trans. Inf. Theory}, vol.~61, no.~6, pp. 3487--3517, Jun. 2015.

\bibitem{ChenSayedTIT2015part2}
J.~Chen and A.~H. Sayed, ``On the learning behavior of adaptive networks --- part II: Performance analysis,'' \emph{IEEE Trans. Inf. Theory}, vol.~61, no.~6, pp. 3518--3548, Jun. 2015.

\bibitem{DeGroot}
M.~H. DeGroot, ``Reaching a consensus,'' \emph{J. Amer. Statist. Assoc.},
  vol.~69, no. 345, pp. 118--121, 1974.

\bibitem{XiaoBoydSCL2004}
L.~Xiao and S.~Boyd, ``Fast linear iterations for distributed averaging,'' \emph{Systems and Control Letters}, vol.~53, no.~1, pp. 65--78, Sep. 2004.

\bibitem{BoydGhoshPrabhakarShahTIT2006}
S.~Boyd, A.~Ghosh, B.~Prabhakar, and D.~Shah, ``Randomized gossip algorithms,'' \emph{IEEE Trans. Inf. Theory}, vol.~52, no.~6, pp. 2508--2530, Jun. 2006.

\bibitem{DimakisKarMouraRabbatScaglioneProcIEEE2010}
A.~G. Dimakis, S.~Kar, J.~M.~F. Moura, M.~G. Rabbat, and A.~Scaglione, ``Gossip algorithms for distributed signal processing,'' \emph{Proceedings of the IEEE}, vol.~98, no.~11, pp. 1847--1864, Nov. 2010.

\bibitem{Sayed}
A.~H. Sayed, ``Adaptation, Learning, and Optimization over Networks,'' \emph{Found. Trends Mach. Learn.}, vol.~7, no. 4-5, pp. 311--801, 2014.

\bibitem{MattaSayedCoopGraphSP2018}
V.~Matta and A.~H. Sayed, ``Estimation and detection over adaptive networks,'' in \emph{Cooperative and Graph Signal Processing}, P.~Djuric and C.~Richard, Eds.\hskip 1em plus 0.5em minus 0.4em\relax Elsevier, 2018, pp. 69--106.

\bibitem{LamReibmanTC1993}
Wai-Man Lam and A.~R. Reibman, ``Design of quantizers for decentralized estimation systems,'' \emph{IEEE Trans. Commun.}, vol.~41, no.~11, pp. 1602--1605, Nov. 1993.

\bibitem{GubnerTIT1993}
J. A. Gubner, ``Distributed estimation and quantization,'' \emph{IEEE Trans. Inf. Theory}, vol.~39, no.~4, pp. 1456--1459, Jul. 1993. 

\bibitem{AsymptoticDesignQuantizers2007}
S.~Marano, V.~Matta, and P.~Willett, ``Asymptotic design of quantizers for decentralized MMSE estimation,'' \emph{IEEE Trans. Signal Process.}, vol.~55, no.~11, pp. 5485--5496, Nov. 2007.

\bibitem{TongARE}
P.~Venkitasubramaniam, L.~Tong, and A.~Swami, ``Quantization for maximin ARE in distributed estimation,'' \emph{IEEE Trans. Signal Process.}, vol.~55, no.~7, pp. 3596--3605, Jul. 2007.

\bibitem{LongoLookabaughGrayTIT1990}
M.~Longo, T.~D. Lookabaugh, and R.~M. Gray, ``Quantization for decentralized hypothesis testing under communication constraints,'' \emph{IEEE Trans. Inf. Theory}, vol.~36, no.~2, pp. 241--255, Mar. 1990.

\bibitem{Tsitsiklis1993}
J.~N. Tsitsiklis,, ``Decentralized detection,'' \emph{Advances in Statistical Signal Processing}, vol.~2, pp. 297--344, 1993.

\bibitem{ViswanathanVarshneyProcIEEE1997}
R.~Viswanathan and P.~K. Varshney, ``Distributed detection with multiple sensors Part I. Fundamentals,'' \emph{Proceedings of the IEEE}, vol.~85, no.~1, pp. 54--63, Jan. 1997.

\bibitem{BlumKassamPoorProcIEEE1997}
R.~S. Blum, S.~A. Kassam, and H.~V. Poor, ``Distributed detection with multiple sensors Part II. Advanced topics,'' \emph{Proceedings of the IEEE}, vol.~85, no.~1, pp. 64--79, Jan. 1997.

\bibitem{ChamberlandVeeravalliTSP2003}
J.~-F. Chamberland and V.~V. Veeravalli, ``Decentralized detection in sensor networks,'' \emph{IEEE Trans. Signal Process.}, vol.~51, no.~2, pp. 407--416, Feb. 2003.

\bibitem{SaligramaTSP2006}
V.~Saligrama, M.~Alanyali, and O.~Savas, ``Distributed detection in sensor networks with packet losses and finite capacity links," \emph{IEEE Trans. Signal Process.}, vol.~54, no.~11, pp. 4118--4132, Nov.~2006.

\bibitem{HanAmariTIT1998}
Te Sun Han and S.~Amari, ``Statistical inference under multiterminal data compression,'' \emph{IEEE Trans. Inf. Theory}, vol.~44, no.~ 6, pp. 2300--2324, Oct. 1998.

\bibitem{CEOproblem}
T.~Berger, Zhen Zhang, and H.~Viswanathan, ``The CEO problem [multiterminal source coding],'' \emph{IEEE Trans. Inf. Theory}, vol.~ 42, no.~3, pp. 887--902, May 1996.

\bibitem{MergenNawareTongTSP2007}
G.~Mergen, V.~Naware, and L.~Tong, ``Asymptotic detection performance of type-based multiple access over multiaccess fading channels,'' \emph{IEEE Trans. Signal Process.}, vol.~55, no.~3, pp. 1081--1092, Mar. 2007.

\bibitem{LBMA}
S.~Marano, V.~Matta, L.~Tong, and P.~Willett, ``A likelihood-based multiple access for estimation in sensor networks,'' \emph{IEEE Trans. Signal Process.}, vol.~55, no.~11, pp. 5155--5166, Nov. 2007.

\bibitem{RagoWillettBarShalomTAES1996}
C.~Rago, P.~Willett, and Y.~Bar-Shalom, ``Censoring sensors: a low-communication-rate scheme for distributed detection,'' \emph{IEEE Trans. Aerosp. Electron. Syst.}, vol.~32, no.~2, pp. 554--568, Apr. 1996.

\bibitem{Widrow1956}
B.~Widrow, ``A study of rough amplitude quantization by means of Nyquist sampling theory," \emph{IRE Trans. on Circuit Theory}, vol.~ 3, no.~4, pp. 266--276, Dec. 1956.

\bibitem{GrayStockhamTIT1993}
R.~M. Gray and T.~G. Stockham, ``Dithered quantizers,'' \emph{IEEE Trans. Inf. Theory}, vol.~39, no.~3, pp. 805--812, May 1993.

\bibitem{LuoTIT2005}
Z.-Q. Luo, ``Universal decentralized estimation in a bandwidth constrained sensor network,'' \emph{IEEE Trans. Inf. Theory}, vol.~ 51, no.~6, pp. 2210--2219, Jun. 2005.

\bibitem{PreddKulkarniPoorSPM2006}
J.~B. Predd, S.~B. Kulkarni, and H.~V. Poor, ``Distributed learning in wireless sensor networks,'' \emph{IEEE Signal Process. Mag.}, vol.~23, no.~4, pp. 56--69, Jul. 2006.

\bibitem{MaranoMattaWillettTSP2013}
S.~Marano, V.~Matta, and P.~Willett, ``Nearest-Neighbor distributed learning by ordered transmissions,'' \emph{IEEE Trans. Signal Process.}, vol.~61, no.~21, pp. 5217--5230, Nov. 2013.

\bibitem{AlistarhNIPS2017}
D.~Alistarh, D.~Grubic, J.~Z. Li, R.~Tomioka, and M.~Vojnovic, ``QSGD: communication-efficient SGD via gradient quantization and encoding'' in \emph{Proc. Neural Information Processing Systems (NIPS)}, Long Beach, CA, USA, Dec. 2017, pp. 1707--1718.

\bibitem{GershoGrayBook}
A.~Gersho and R.~M. Gray, \emph{Vector Quantization and Signal Compression}.\hskip 1em plus 0.5em minus 0.4em\relax Springer Science+Business Media, New York, 2001.

\bibitem{SeideFuDroppoLiYuINTERSPEECH2014}
F.~Seide, H.~Fu, J.~Droppo, G.~Li, and D.~Yu, ``1-Bit Stochastic Gradient Descent
and its Application to Data-Parallel Distributed Training of Speech DNNs,'' in \emph{Proc. Conference of the International Speech Communication Association}, Singapore, Sep. 2014, pp. 1058--1062.

\bibitem{WuHuangHuangZhangICMLR2018}
J.~Wu, W.~Huang, J.~Huang, and T.~Zhang, ``Error Compensated Quantized SGD and its Applications to Large-scale
Distributed Optimization,'' in \emph{Proc. International Conference on Machine Learning (ICML)}, Stockholm, Sweden, Jul. 2018, pp. 5235--5333.

\bibitem{StichCordonnierJaggiNIPS2018}
S.~U. Stich, J-B. Cordonnier, and M.~Jaggi, ``Sparsified SGD with Memory,'' in \emph{Proc. Neural Information Processing Systems (NIPS)}, Montréal, Canada, Dec. 2018, pp. 4447--4458.

\bibitem{KarimireddyRebjockStichJaggiICML2019}
S.~P. Karimireddy, Q.~Rebjock, S.~U. Stich, and M.~Jaggi, ``Error Feedback Fixes SignSGD and other Gradient
Compression Schemes,'' in \emph{Proc. International Conference on Machine Learning (ICML)}, Long Beach, CA, USA, Jun. 2019, pp. 3252--3261.

\bibitem{YuanYingZhaoSayedSP2019part1}
K.~Yuan, B.~Ying, X.~Zhao, and A.~H. Sayed, ``Exact diffusion for distributed optimization and learning -- Part I: Algorithm development,'' in \emph{IEEE Trans. Signal Process.}, vol.~67, no.~3, pp. 708--723, Feb. 2019.

\bibitem{YuanYingZhaoSayedSP2019part2}
K.~Yuan, B.~Ying, X.~Zhao, and A.~H. Sayed, ``Exact diffusion for distributed optimization and learning -- Part II: Convergence analysis,'' in \emph{IEEE Trans. Signal Process.}, vol.~67, no.~3, pp. 724--739, Feb. 2019.

\bibitem{LinKostinaHassibiISIT2021}
C.-Y. Lin, V.~Kostina, and B.~Hassibi, ``Differentially quantized gradient descent,'' in \emph{Proc. IEEE International Symposium on Information Theory (ISIT)}, Melbourne, Victoria, Australia, Jul. 2021, pp. 1200--1205.

\bibitem{NedicOlshevskyRabbatIEEEPROC2018}
A.~Nedi\'c, A.~Olshevsky, and M.~G. Rabbat, ``Network topology and communication-computation tradeoffs in
decentralized optimization,'' \emph{Proceedings of the IEEE}, vol.~106, no.~5, pp. 953--976, Apr. 2018.

\bibitem{ZhaoTuSayedSP2012}
X.~Zhao, S.-Y. Tu, and A.~H. Sayed, ``Diffusion adaptation over networks under imperfect information exchange and non-stationary data,'' in \emph{IEEE Trans. Signal Process.}, vol.~60, no.~7, Apr. 2012, pp. 3460--3475.

\bibitem{NedicOlshevskyOzdaglarTsitsiklisCDC2008}
A.~Nedi\'c, A.~Olshevsky, A.~Ozdaglar, and J.~Tsitsiklis, ``Distributed subgradient methods and quantization effects,'' in \emph{Proc. IEEE Conference on Decision and Control (CDC)}, Cancun, Mexico, Dec. 2008, pp. 4177--4184.

\bibitem{DoanMaguluriRombergIEEEAUTCONT12020}
T.~Doan, S.~T. Maguluri, and J.~Romberg, ``Convergence rates of distributed gradient methods under random
quantization: A stochastic approximation approach,'' \emph{IEEE Trans. Autom. Control}, vol.~66, no.~10, pp. 4469--4484, Oct. 2021.

\bibitem{ReisidazehMokhtariHassaniPedarsaniIEEESP2019}
A.~Reisidazeh, A.~Mokhtari, H.~Hassani, and R.~Pedarsani, ``An exact quantized decentralized gradient descent algorithm,'' \emph{IEEE Trans. Signal Process.}, vol.~67, no.~19, pp. 4934--4947, Aug. 2019.

\bibitem{KoloskovaStichJaggiICML2019}
A.~Koloskova, S.~U. Stich, and M.~Jaggi, ``Decentralized stochastic optimization and gossip algorithms with compressed communication,'' in \emph{Proc. International Conference on Machine Learning (ICML)}, Long Beach, CA, USA, Jun. 2019, pp. 3478--3487.

\bibitem{KoloskovaLinStichJaggiICLR2020}
A.~Koloskova, T.~Lin, S.~U. Stich, and M.~Jaggi, ``Decentralized deep learning with arbitrary communication compression,'' in \emph{Proc. International Conference on Learning Representations (ICLR)}, Addis Ababa, Ethiopia, Apr./May 2020, pp. 1--22.

\bibitem{BitsForFree}
D.~Kovalev, A.~Koloskova, M.~Jaggi, P.~Richt\'arik, and S.~U. Stich, ``A linearly convergent algorithm for decentralized optimization: sending less bits for free!,'' in \emph{Proc. International Conference on Artificial Intelligence and Statistics (AISTATS)}, San Diego, CA, USA, Apr. 2021, pp. 4087--4095.

\bibitem{ChenSayedPareto} 
J.~Chen and A.~H. Sayed, ``Distributed Pareto optimization via diffusion strategies,'' \emph{IEEE J. Sel. Topics Signal Process.}, vol.~7, no. 2, pp. 205--220, Apr. 2013.

\bibitem{Johnson-Horn}
R.~A. Horn and C.~R. Johnson, \emph{Matrix Analysis}.\hskip 1em plus 0.5em minus
  0.4em\relax Cambridge University Press, New York, 1985.

\bibitem{GolubWilkinsonSIAM1976}
G.~H. Golub and J.~H. Wilkinson, ``Ill-Conditioned Eigensystems and the Computation of the Jordan Canonical Form,'' \emph{SIAM Review}, vol.~18, no.~4, pp. 578--619, 1976.

\bibitem{Bronson}
R. Bronson, \emph{Matrix Methods. An Introduction}.\hskip 1em plus 0.5em minus
  0.4em\relax Gulf Professional Publishing, Houston, 1991.

\bibitem{bib:diagonalrankone}
J.~Moro, J.~V. Burke, and M.~L. Overton, ``On the Lidskii-Vishik-Lyusternik perturbation theory for eigenvalues of matrices with arbitrary Jordan structure,'' \emph{SIAM Journal of Matrix Analysis and Applications}, vol.~18, no.~4, pp. 793--817, Oct. 1997.

\bibitem{SayedPrimalDual}
Z.~J. Towfic and A.~H. Sayed, ``Stability and performance limits of adaptive primal-dual networks,'' \emph{IEEE Trans. Signal Process.}, vol.~63, no.~11, pp. 2888--2903, Jun. 2015.

\bibitem{VlaskiSayedTSP2021part1}
S.~Vlaski and A.~H. Sayed, ``Distributed learning in non-convex environments --- part I: Agreement at a linear rate,'' \emph{IEEE Trans. Signal Process.}, vol.~69, pp. 1242--1256, Jan. 2021.

\bibitem{VlaskiSayedTSP2021part2}
S.~Vlaski and A.~H. Sayed, ``Distributed learning in non-convex environments --- part II: Polynomial escape from saddle-points,'' \emph{IEEE Trans. Signal Process.}, vol.~69, pp. 1257--1270, Jan. 2021.

\bibitem{VlaskiSayedTAC2021}
S.~Vlaski and A.~H. Sayed, ``Second-order guarantees of stochastic gradient descent in non-convex optimization,'' \emph{IEEE Trans. Autom. Control}, 2022, {\em to appear}, available online as arXiv:1908.07023v1 [math.OC].

\bibitem{Trefethen-Embree}
L.~N. Trefethen and M. Embree, \emph{Spectra and Pseudospectra. The Behavior of Nonnormal Matrices and Operators}.\hskip 1em plus 0.5em minus
  0.4em\relax Princeton University Press, New Jersey, 2005.

\end{thebibliography}
\end{document}